\RequirePackage{fix-cm}
\documentclass{article}
\usepackage{PRIMEarxiv}

\usepackage[utf8]{inputenc}
\usepackage[T1]{fontenc}
\usepackage{microtype}
\usepackage{graphicx}
\usepackage{booktabs}
\usepackage{array}
\usepackage{multirow}
\usepackage{subcaption}
\usepackage{wrapfig}
\usepackage{xcolor}
\usepackage{colortbl}
\usepackage{amsmath}
\usepackage{amssymb}
\usepackage[authoryear]{natbib}
\bibpunct{[}{]}{,}{a}{,}{,}
\usepackage[hidelinks]{hyperref}
\usepackage{url}

\usepackage{amsmath,amsfonts,bm}

\def\eqref#1{equation~\ref{#1}}

\def\1{\bm{1}}

\DeclareMathAlphabet{\mathsfit}{\encodingdefault}{\sfdefault}{m}{sl}
\SetMathAlphabet{\mathsfit}{bold}{\encodingdefault}{\sfdefault}{bx}{n}

\newtheorem{proposition}{Proposition}

\title{A First-Step Bias Restores Lost Diversity\\ in Diffusion Model Distillation}

\author{
Wenwu Tang\textsuperscript{$\diamond$},
Dong Wang\textsuperscript{$\diamond$}, 
Yun Cheng\textsuperscript{$\star$}, 
Olga Saukh\textsuperscript{$\diamond$,§} \\
\textsuperscript{$\diamond$}Graz University of Technology, 
\textsuperscript{§}Complexity Science Hub, Austria \\
\textsuperscript{$\star$}Swiss Data Science Center, Switzerland \\
\texttt{\{wenwu.tang, dong.wang, saukh\}@tugraz.at}, 
\texttt{yun.cheng@sdsc.ethz.ch} \\
}

\begin{document}

\maketitle

\begin{abstract}
Distribution-matching distillation enables high-quality image generation in only a few denoising steps. However, distilled student models often map different initial noise samples to similar outputs under the same prompt. To locate where diversity is lost, we compare the teacher and student along the sampling trajectory. The largest measured reduction in the response to different initial noises occurs during the first, high-noise step. When we replace the first step with the teacher's corresponding trajectory and leave the final three student steps unchanged, most of the output diversity returns. This suggests that the first step is a key point for restoring diversity. Motivated by
this finding, we introduce \emph{ReSeed}, a post-distillation method that
adds a seed-response bias adapter only to the first
student step and aligns its velocity with the teacher's first-interval secant
velocity. All student weights and later denoising steps remain unchanged. We evaluate ReSeed on three text-to-image models (SD3-Medium, SD3.5-Medium, and FLUX.1-dev) and one text-to-video model (Wan2.1-T2V-1.3B). For a distilled 4-step FLUX.1-dev model, \emph{ReSeed} requires only $0.00196\%$ additional trainable parameters relative to the base model and is trained on 8,000 prompt--noise pairs, while keeping the distilled student fully frozen and requiring no target images, perceptual losses, or adversarial losses. It improves DINO diversity by $48.3\%$ while maintaining competitive visual quality. These results show that substantial diversity can be recovered through a lightweight, data-efficient tuning of only the first denoising step, without finetuning the full student model. Additional results are available on our \href{https://twwinde.github.io/reseed-project-page/}{\textcolor{blue}{Project Page}}.

\end{abstract}

\begin{figure}[!ht]
\centering
\setlength{\abovecaptionskip}{4pt}
\includegraphics[width=\textwidth]{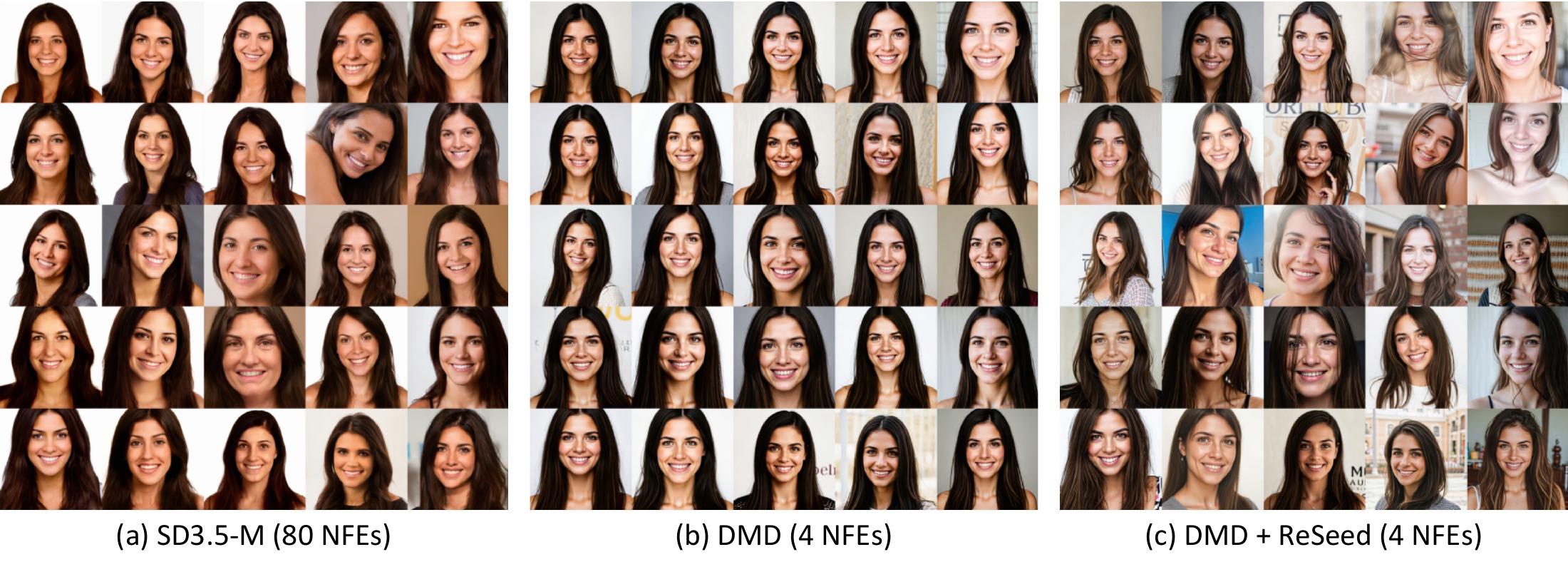}
\caption{\textbf{ReSeed recovers visible same-prompt diversity.}
All panels use the prompt ``A smiling woman with long dark brown hair, hazel
eyes, and a warm expression'' and the same 25 fixed initial noises, matched
cell by cell. (a) The 40-step SD3.5-Medium teacher at CFG $4.5$. (b) The frozen
4-step DMD student, and (c) The same frozen 4-step DMD student with ReSeed.}
\label{fig:sd35m-teaser}
\end{figure}

\section{Introduction}

Diffusion and flow models generate high-quality images and videos, but often
require many network evaluations \citep{ho2020ddpm,esser2024sd3}.
Distillation reduces this cost by training a student to generate samples in
only a few denoising steps \citep{salimans2022progressive,
song2023consistency,yin2024dmd}. However, students trained with distribution
matching distillation (DMD) often produce similar outputs from different
initial noises under the same prompt. This loss of diversity reduces the
range of results that users can obtain from a few-step model.

Previous works try to address this problem by aligning the student's response
to initial noise \citep{huang2026gad}, assigning different training objectives
to early and late denoising steps \citep{wu2026dpdmd}, or using the base model
for the first sampling step \citep{gandikota2025distilling}. These studies
highlight the importance of preserving variation during early generation. Further related work is discussed in ~\ref{app:related-work}.
Building on these findings, we ask a practical question:
\emph{can we recover diversity in an already distilled student by changing
only its first denoising step, while keeping its original weights and later
steps fixed?}

We first examine this question through controlled interventions on
SD3-Medium. We replace the first student step with the teacher's trajectory
over the same time interval, then run the final three student steps without
any changes. This yields a higher DINOv3 diversity than that of the 28-step teacher.
The result shows that the
unchanged later steps can preserve much of the diversity recovered by the
teacher's first interval. It supports our central hypothesis: a substantial
part of the measured diversity loss can be restored at the first step,
without retraining the rest of the student.

Motivated by this finding, we introduce \emph{ReSeed}, a post-distillation
method that trains small bias adapters only for the first student step
(Figure~\ref{fig:dcb-schematic}). We add learned per-channel biases to
transformer block outputs and keep all original student weights frozen.
Training matches the student's first-step velocity to the teacher's average
velocity over the same interval, using the same prompt and initial noise.
Matching these velocities also matches the states reached at the first
student boundary. At inference, the biases are enabled only during the
first step; later steps use the unchanged student, with no teacher calls
or additional denoising steps.
The biases are shared across prompts and noise seeds, so they do not
provide new seed-specific information. Instead, the frozen nonlinear
network turns the shared biases into input-dependent corrections.
Our analysis links diversity recovery to better alignment with the
teacher's seed-dependent responses, rather than simply increasing
response magnitudes. This explains why we study a small first-step
correction instead of changing the entire sampling process.

We evaluate ReSeed on SD3-Medium, SD3.5-Medium, and SenseFlow-FLUX,
a 4-step student of FLUX.1-dev
\citep{blackforestlabs2024flux,ge2026senseflow}, and extend the evaluation
to Wan2.1 text-to-video generation. For 4-step FLUX.1-dev student, ReSeed trains
only $233{,}472$ parameters ($0.00196\%$ of the denoiser parameters) using $8{,}000$ prompt--noise pairs.
It requires no target images, perceptual losses, or adversarial losses.
DINOv3 diversity increases by $48.3\%$, while FID decreases from
$40.22$ to $34.65$. Experiments on the other image models and the video
model also show diversity gains, with trade-offs in some quality
metrics. These results show that substantial diversity can be recovered
with very few trainable parameters after distillation.

Our main contributions are:
\begingroup
\setlength{\leftmargini}{1.5em}
\begin{itemize}
\setlength{\itemsep}{1pt}
\setlength{\topsep}{2pt}

\item \textbf{Most diversity loss occurs in the first denoising step.}
We show that most of the diversity loss in few-step distilled models occurs in the first denoising step, while the frozen suffix retains the ability to preserve diversity.

\item \textbf{A small first-step adapter can recover lost diversity.}
We introduce \emph{ReSeed}, a first-step residual tuning that restores lost diversity while keeping the pretrained distilled student frozen, requiring less than $0.005\%$ additional trainable parameters and approximately 8,000 prompt--noise pairs.

\item \textbf{Existing distilled models can be repaired without repeating distillation.}
Our results demonstrate that diversity in an already distilled model can be substantially improved through a lightweight post-hoc tuning, without repeating distillation or fine-tuning the whole original student. This provides a practical way to enhance the diversity of existing few-step generative models after training.

\end{itemize}
\endgroup

\section{Preliminaries}
\label{sec:prelim}

\paragraph{Flow matching.} Diffusion models \citep{ho2020ddpm} and flow
matching \citep{lipman2023flow,liu2023rectified} can both be read as a learned
velocity field that defines a probability-flow ODE \citep{song2021sde}: a
diffusion model derives the velocity from its predicted score, flow matching
regresses it directly. For rectified-flow models, a clean latent $x_0$ and
a Gaussian noise $\epsilon$ are joined by the straight path
$x_t=(1-t)\,x_0+t\,\epsilon$ with flow time
$t\in[0,1]$, whose derivative is the constant velocity
$\mathrm{d}x_t/\mathrm{d}t=\epsilon-x_0$. A network
$v_\theta(x_t,t,c)$, with trainable parameters $\theta$ and prompt
condition $c$, is trained to predict this velocity,
\begin{equation}
\mathcal{L}_{\mathrm{FM}}(\theta)=
\mathbb{E}_{t,x_0,\epsilon}
\bigl\lVert v_\theta(x_t,t,c)-(\epsilon-x_0)\bigr\rVert_2^2 ,
\label{eq:fm}
\end{equation}
where $\mathcal{L}_{\mathrm{FM}}$ is the flow-matching objective,
$\mathbb{E}_{t,x_0,\epsilon}$ averages over the sampled flow time,
training latent, and Gaussian noise, and $\lVert\cdot\rVert_2$ is the Euclidean
norm over latent elements. The predicted velocity gives the clean estimate
$\hat{x}_0=x_t-t\,v_\theta(x_t,t,c)$. Sampling starts
from pure noise $z=x_1$ and integrates
$\mathrm{d}x_t/\mathrm{d}t=v_\theta$ from $t=1$ to $0$. We
call the exact solution of this ODE
$\Phi_{\mathrm{}}:z\mapsto x_0$. With
Euler steps on a decreasing grid $1=t_0>t_1>\dots>t_K=0$,
\begin{equation}
x_{t_{k+1}}=x_{t_k}+(t_{k+1}-t_k)\,
v_\theta(x_{t_k},t_k,c),
\qquad k=0,\dots,K-1 .
\label{eq:euler}
\end{equation}
Here $K$ is the number of denoising steps, $k$ indexes a step, and
$x_{t_k}$ is the latent state at flow time $t_k$; the subscript $k$ on $t_k$
indexes denoising boundaries, and the signed increment $t_{k+1}-t_k$ is negative
because sampling moves from
noise to data. The sampler is deterministic: for a fixed prompt, everything that
differs between two images comes from the initial noise $z$.

\paragraph{Distribution matching distillation.} DMD
\citep{yin2024dmd,yin2024dmd2} trains a few-step student to match the teacher's
output distribution. Let $x_\theta=F_\theta(z,c)$ be the student output and let
$p_{\mathrm{stu},\theta}$ and $p_{\mathrm{tea}}$ denote the student and teacher
output distributions. The ideal distribution-matching objective is the reverse
Kullback--Leibler divergence
\begin{equation}
\mathcal{L}_{\mathrm{DMD}}(\theta)
\mathrel{:=}
D_{\mathrm{KL}}\!\left(p_{\mathrm{stu},\theta}\,\|\,p_{\mathrm{tea}}\right).
\label{eq:dmd-objective}
\end{equation}
Here $\theta$ denotes the student parameters, $z$ is the initial noise, $c$ is
the prompt, and $D_{\mathrm{KL}}$ is the KL divergence. This divergence is not
evaluated directly. In practice, DMD updates the student using estimated scores
of re-noised samples \citep{yin2024dmd,yin2024dmd2}. Under the flow
parameterization used here, we form $x_t=(1-t)x_\theta+t\epsilon$, where
$\epsilon\sim\mathcal{N}(0,I)$ is independent Gaussian noise. The score-based
update is
\begin{equation}
\nabla_{\theta}\ell_{\mathrm{DMD}}(\theta) =
\mathbb{E}_{z,t,\epsilon}\!\left[
w\,J_\theta(x_t)^\top
\bigl(\hat{s}_{\mathrm{stu}}(x_t,t,c)-\hat{s}_{\mathrm{tea}}(x_t,t,c)\bigr)
\right],
\label{eq:dmd}
\end{equation}
Here $\nabla_{\theta}\ell_{\mathrm{DMD}}(\theta)$ is the score-based update vector,
$J_\theta(x_t)=\partial x_t/\partial\theta$ is the Jacobian of the re-noised
sample, and $w$ is a scalar weighting factor. The expectation averages over
initial noise $z$, re-noising time $t$, and re-noising noise $\epsilon$.
The frozen teacher provides $\hat{s}_{\mathrm{tea}}$, while an auxiliary model
fitted online to student samples provides $\hat{s}_{\mathrm{stu}}$.
These estimate the gradients of the log densities of the re-noised teacher
and student outputs, respectively.
Reverse KL is
mode seeking: it penalises student samples the teacher finds unlikely, not
teacher modes the student never produces.

\paragraph{Diversity as the response to the initial noise.} For a fixed prompt
$c$, the deterministic sampler maps initial noise $z$ to the output
$F_\theta(z,c)$. We call the change in output caused by changing the initial
noise the \emph{seed response}. We write the complete sampler as
$F_\theta(z,c)=H_\theta(R_\theta(z,c),c)$, where $R_\theta$ maps the initial
noise to the first-step state and $H_\theta$ denotes the remaining steps.
We use this decomposition to test whether the observed diversity loss can be
corrected at the first step while keeping the remaining steps unchanged.

\section{First-step correction restores diversity}
\label{sec:diagnosis}
\label{sec:mechanism}

A first-step repair can restore output diversity only if the unchanged later
steps still preserve differences between seeds. We test this condition on the
4-step SD3-Medium student, then ask whether better first-step seed responses
accompany more diverse final images.

\subsection{First-step replacement restores diversity}

\begin{figure}[t]
\centering
\includegraphics[width=0.834\textwidth]{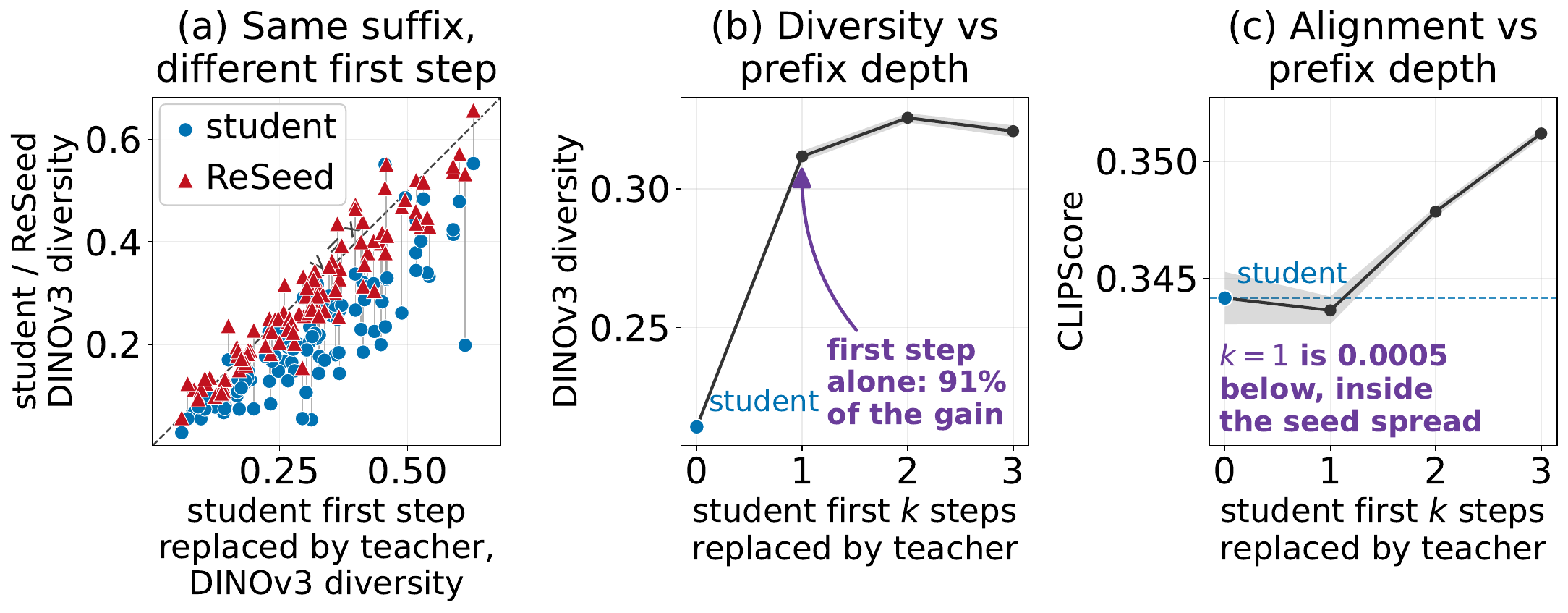}
\caption{First-step replacement on SD3-Medium. (a) With student steps 2--4
fixed, replacing step 1 by the teacher raises diversity for $98/100$ prompts;
ReSeed moves them toward the teacher-prefix result. (b,c) Replacing the first
$k$ steps shows that $k=1$ gives $91\%$ of the diversity gain. CLIPScore
changes little until later steps are replaced. Bands show the standard
deviation over three seeds.}
\label{fig:first-interval-swap}
\end{figure}

We replace the first student step with corresponding teacher steps covering
the same time interval. The remaining three steps use the unchanged student.
Both samplers share the same prompts and initial noises, so the replacement
changes only how they reach the first student boundary.
This replacement increases
DINOv3 diversity from $0.2142$ to $0.3117$. Replacing additional student intervals brings smaller further gains
(Figure~\ref{fig:first-interval-swap}).
The  replacement requires extra teacher evaluations and serves as a
diagnostic test, rather than an equal-cost alternative to the student.

These results show that the unchanged final three steps can produce diverse
images when given a different first-boundary state. They identify the first
step as an effective repair point in this setting, without requiring changes
to the remaining steps.
These findings motivate matching the student's first-step state to the
teacher's. For an Euler step, this is equivalent to matching the student's
velocity to the teacher's average velocity over the same interval, which
forms the training objective in Section~\ref{sec:method}.

\subsection{Better seed-response alignment accompanies higher diversity}
\label{sec:gap}

For a fixed prompt, a seed response is the difference between two first-step
endpoints from different initial noises. On 5{,}000 COCO prompts with ten seeds
each, we compare this response with the teacher's for the frozen SD3-Medium
student and seven ReSeed checkpoints. Mean response error falls by a factor of
$2.7$ as final-image DINOv3 diversity rises from $0.213$ to $0.290$
(Figure~\ref{fig:prop1-response-diversity}). Across checkpoints, lower error
accompanies higher diversity for $90\%$ of prompts. This association supports
correcting the first endpoint while keeping later steps unchanged.

\section{ReSeed: a seed-response bias for the first step}
\label{sec:method}

\begin{figure}[!t]
\centering
\includegraphics[width=0.80\textwidth]{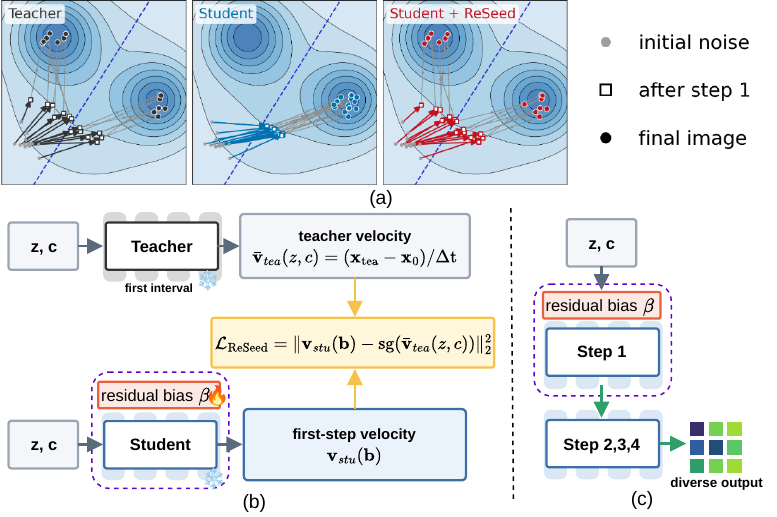}
\caption{\textbf{Overview of ReSeed}. (a) First-step contraction can merge
seed trajectories; the teacher and ReSeed retain alternative paths. (b)
Training fits first-step block-output biases to the teacher's interval-average
velocity with frozen base weights. (c) The bias is off in steps 2--4.}
\label{fig:dcb-schematic}
\end{figure}

ReSeed adds a constant bias adapter to each transformer block, active only in
the first student step (Figure~\ref{fig:dcb-schematic}). The four steps share
backbone parameters, so changing an original bias would affect them all.
Instead, we freeze the backbone and disable the adapters in steps 2--4. The
adapters contain no seed or prompt information; their input-dependent effect
arises through the frozen network.

\paragraph{Teacher-secant target.} Let
$x_{t_0}=z$ be the initial Gaussian latent and let
$x^{\mathrm{tea}}_{t_1}(z,c)$ be the endpoint reached by the teacher
from the same latent and prompt after the teacher steps corresponding to the
student's first denoising interval. The target
is the teacher's average velocity across this interval,
\begin{equation}
\bar v_{\mathrm{tea}}(z,c)=
\frac{x^{\mathrm{tea}}_{t_1}(z,c)-x_{t_0}}{t_1-t_0},
\end{equation}
where $z$ is the sampled initial noise, $c$ is the prompt,
$t_0$ and $t_1$ are the two flow-time boundaries of the first denoise interval, and the bar
denotes an average over this interval. This is a displacement target, not the
teacher's instantaneous velocity at the last step. The training loss and the sampler
are
\begin{equation}
\mathcal{L}_{\mathrm{ReSeed}}=
\mathbb{E}_{c,z}\!\left[
\frac{1}{d}\bigl\lVert v^{\beta}_{\mathrm{stu}}(x_{t_0},t_0,c)
-\operatorname{sg}\!\left(\bar v_{\mathrm{tea}}(z,c)\right)\bigr\rVert_2^2
\right],
\qquad
x_{t_1}=x_{t_0}+(t_1-t_0)
v^{\beta}_{\mathrm{stu}}(x_{t_0},t_0,c),
\end{equation}
where
$\mathbb{E}_{c,z}$ averages over prompt--noise pairs, $d$ is the number of
latent elements, $\beta$ is the collection of trainable bias adapters,
$v^{\beta}_{\mathrm{stu}}$ is the frozen student evaluated with those biases,
$\operatorname{sg}$ stops gradients through the teacher target. The second expression
is the Euler update that produces the adapted first endpoint $x_{t_1}$. For intervals $k\in\{1,2,3\}$, the adapter is disabled and the
original frozen student advances from the modified first endpoint.
The training loss is defined for individual prompt--noise pairs, whereas
diversity concerns differences between samples under the same prompt. The
following proposition connects these two views: the same loss also bounds
the error in pairwise seed responses at the first endpoint.

\begin{proposition}[Pointwise regression bounds pairwise seed-response error]
\label{prop:training-pairwise-bound}
For each prompt $c$, let $z,z'$ be independent noises from the same initial-noise
distribution. Let $F_\beta(z,c)$ denote the adapted first endpoint and
$T(z,c)$ the teacher endpoint. Define
$\Delta F_\beta=F_\beta(z,c)-F_\beta(z',c)$,
and $\Delta T=T(z,c)-T(z',c)$. Let $d$ be the number of
latent elements. The pointwise regression objective used during training
upper-bounds the expected same-prompt pairwise response error per latent
element at the fitted endpoint, up to the factor $2(t_1-t_0)^2$:
\begin{equation}
\frac{1}{d}\mathbb{E}_{c,z,z'}\!\left\|\Delta F_\beta-\Delta T\right\|_2^2
\leq 2(t_1-t_0)^2\mathcal{L}_{\mathrm{ReSeed}}.
\label{eq:pairwise-bound}
\end{equation}
\end{proposition}

\begin{figure}[h]
\centering
\includegraphics[width=\textwidth]{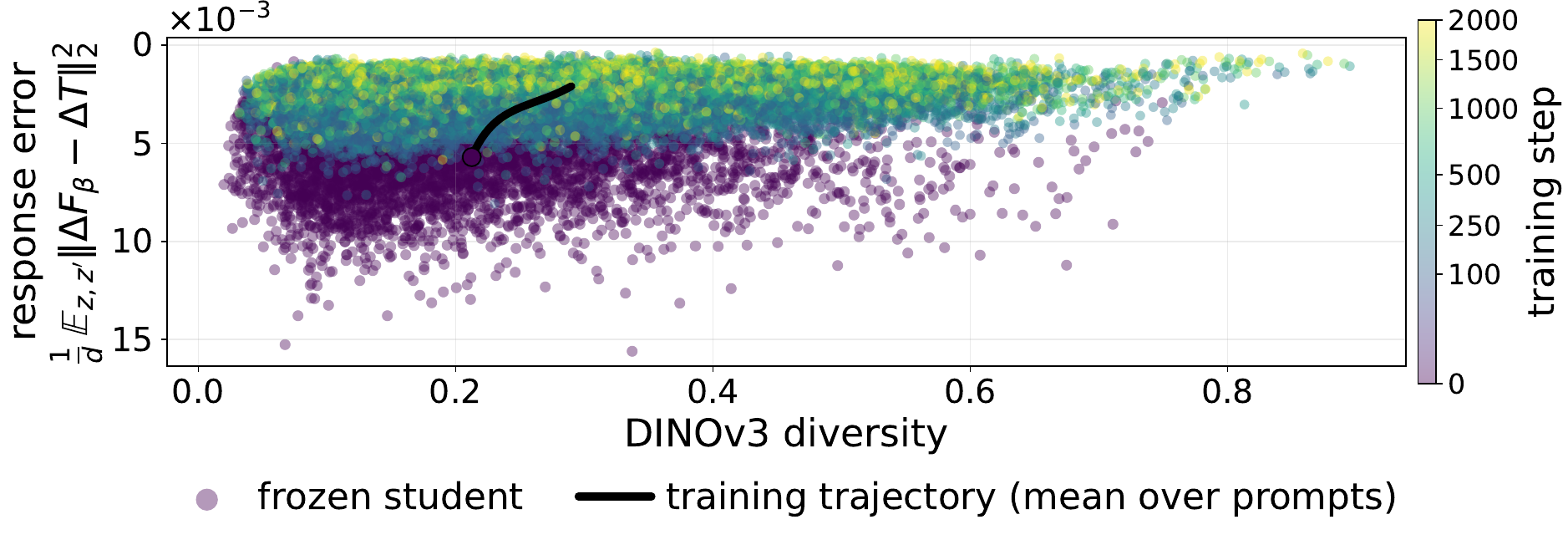}
\caption{Seed-response error versus DINOv3 diversity on COCO5K
($5{,}000$ prompts $\times$ 10 seeds). The error axis is reversed; black
curve is checkpoint means.}
\label{fig:prop1-response-diversity}
\end{figure}

Appendix~\ref{app:pairwise-response} gives the full proof.
This bound controls first-endpoint response error, not final-image
diversity; whether the recovered response can improve the generated image diversity is tested
empirically in Section~\ref{sec:diagnosis}.
Figure~\ref{fig:prop1-response-diversity}
shows that lower pairwise endpoint response error is associated with higher
final-image DINOv3 diversity.

\paragraph{Bias adapter parameterisation.} For block $b$ in the model, we add constant
per-channel vectors
$\beta_b\in\mathbb{R}^{N}$ to the block output of both streams, each $\beta_b$ is
an $N$-dimensional channel vector. The bias is broadcast over the batch and
token dimensions and initialized to zero, so an
untrained adapter reproduces the original student model exactly. For SD3-Medium, $N_{\mathrm{bias}}=72{,}192$
parameters or $0.00356\%$ of the $2{,}028{,}328{,}000$ student parameters.
Unlike BitFit \citep{benzaken2022bitfit}, ReSeed does not unfreeze existing model
biases; it adds new zero-initialised residual vectors whose activity is gated by
the denoising interval.

\section{Experiments and Results}
\label{sec:results}

\subsection{Experimental setup}
\label{sec:setup}

\paragraph{Models and datasets.} We evaluate 4-step DMD students of
SD3-Medium and SD3.5-Medium \citep{esser2024sd3,yin2024dmd}, the 4-step
SenseFlow student of FLUX.1-dev
\citep{blackforestlabs2024flux,ge2026senseflow}, and the 4-step
Wan2.1-T2V-1.3B DMD2 student \citep{wan2025,yin2024dmd2} released with
Data-Forcing Distillation (DFD) \citep{chen2026dataforcing}. Image adapters use
JourneyDB prompts \citep{sun2023journeydb}; the video adapter uses a pool of
4{,}000 ViPE-Wild-1M captions \citep{huang2025vipe}. Training requires only
prompts and initial noise.

\paragraph{Training.} We train only first-step bias adapters, keeping the
teachers and student backbones frozen. We use three training seeds for the
main image results and one for video. Reported checkpoints, training budgets, and
sampling settings are given in Appendix~\ref{app:details}.

\paragraph{Evaluation.} We evaluate image quality on 5{,}000 images generated
from COCO-2017 validation captions \citep{lin2014coco}, using CLIP score
\citep{radford2021clip}, PickScore \citep{kirstain2023pickscore}, HPSv2.1
\citep{wu2023hpsv2}, the LAION aesthetic score \citep{schuhmann2022laion},
and FID \citep{heusel2017fid}. Image diversity uses a separate set of 100 prompts
with $R=10$ independent initial noises per prompt. For a feature extractor $\phi$,
we define same-prompt diversity as
\begin{equation}
D_{\phi}(c)
=1-\frac{2}{R(R-1)}
\sum_{1\leq i<j\leq R}
\cos\!\left(
\phi\!\left(x_{\theta}^{(i)}(c)\right),
\phi\!\left(x_{\theta}^{(j)}(c)\right)
\right).
\label{eq:feature-diversity}
\end{equation}
Here $x_{\theta}^{(i)}(c)$ is the image generated by model $\theta$ for prompt
$c$ from noise sample $i$; $i,j$ index distinct samples, $\phi$ maps images
to feature vectors, and $\cos$ denotes cosine similarity.
We average $D_{\phi}(c)$ over prompts using DINOv3
\citep{simeoni2025dinov3} and CLIP features, and also report mean pairwise
LPIPS-Alex distance \citep{zhang2018lpips}.

For video, we use a category-balanced subset of 320 VBench prompts
\citep{huang2024vbench}. Each method generates eight videos per prompt for
diversity and text alignment; five of these are used for VBench quality.
We report seven quality dimensions, CLIP text alignment, and video-level
and per-frame diversity using CLIP and DINOv2 features
\citep{oquab2024dinov2}. Within each comparison, all methods
share the same prompts, initial noises, and scorers.
Appendix~\ref{app:video-protocol} gives the video protocol.

\subsection{Text-to-Image Generation}
\label{sec:text-to-image}
\label{sec:main-results}

\par\medskip
\noindent
\begin{minipage}[t]{0.46\textwidth}
\vspace{0pt}
\fontsize{10.5}{12}\selectfont
\textbf{Comparison baselines.} Our primary baseline is the original
4-step DMD student. We also compare with GAD, which matches teacher and
student responses to initial-noise perturbations \citep{huang2026gad}, and
DP-DMD, which uses a teacher-derived prediction target for the first step and
the DMD loss for later steps \citep{wu2026dpdmd}. Both methods change the
distillation process. In contrast, ReSeed is applied after distillation: it
freezes the student and trains only first-step bias adapters. We also apply
ReSeed to a GAD student to test whether the two methods can work together.
All 4-step students
comparisons are sampled at CFG $1$.
\end{minipage}\hfill
\begin{minipage}[t]{0.52\textwidth}
\vspace{0pt}
\centering
\includegraphics[width=0.95\linewidth]{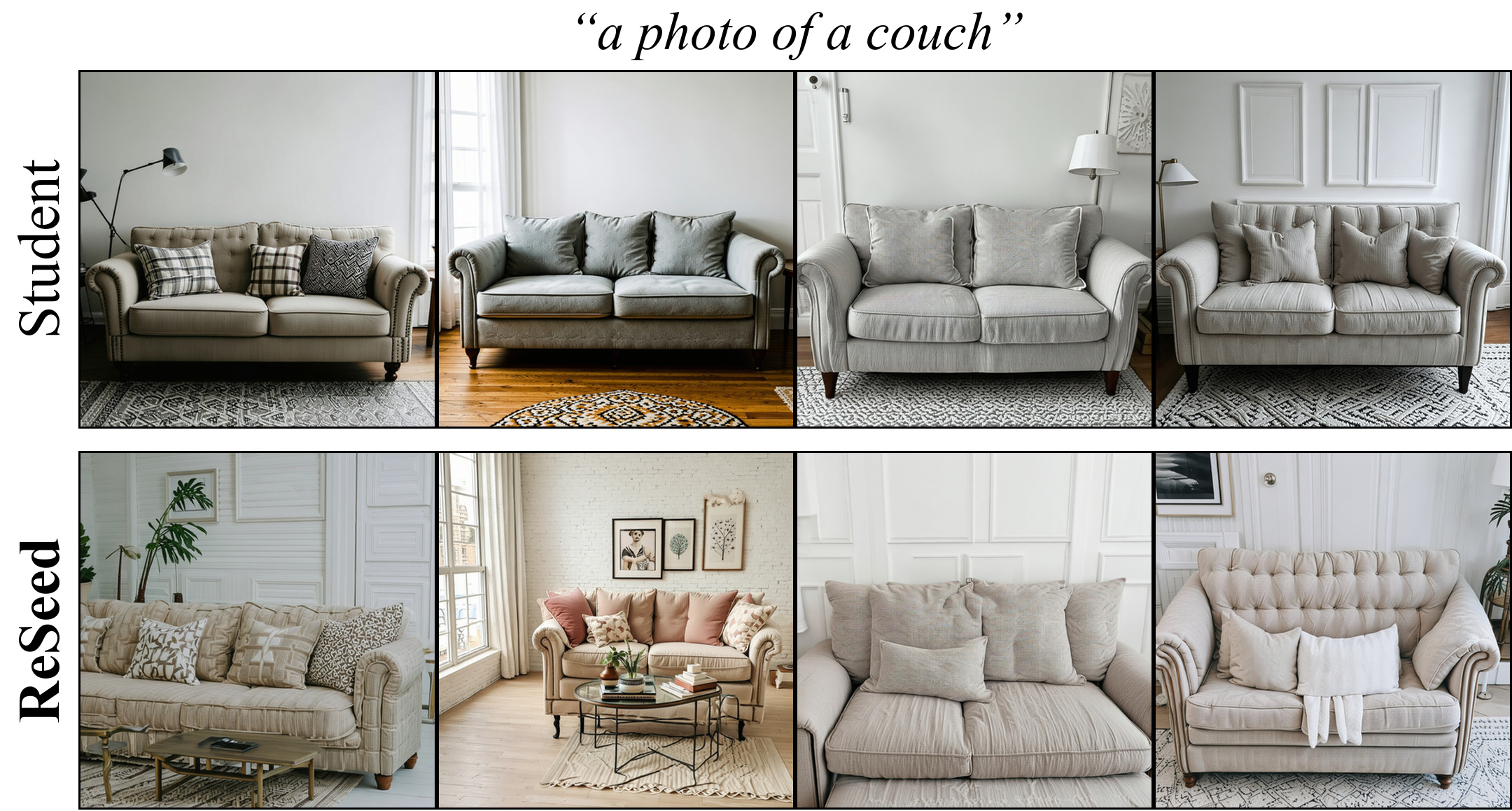}
\captionsetup{font=small,skip=4pt}
\captionof{figure}{Matched-noise SD3.5-Medium samples. Top: the frozen DMD student.
Bottom: the same student with ReSeed.}
\label{fig:sd35m-couch-matched}
\end{minipage}
\par\medskip

\begin{table}[htbp]
\centering
\scriptsize
\setlength{\tabcolsep}{1.6pt}
\renewcommand{\arraystretch}{1.08}
\caption{Results on SD3-Medium and SD3.5-Medium under a common evaluation protocol.}
\label{tab:main}
\resizebox{\textwidth}{!}{%
\begin{tabular}{@{}l|c|ccc|ccccc@{}}
\toprule
\multirow{2}{*}{\textbf{Method}} & \multirow{2}{*}{\textbf{NFEs}}
& \multicolumn{3}{c|}{\textbf{Diversity}}
& \multicolumn{5}{c}{\textbf{Quality}} \\
& & \textbf{DINO $\uparrow$} & \textbf{CLIP dist. $\uparrow$} & \textbf{LPIPS $\uparrow$}
& \textbf{CLIP $\uparrow$} & \textbf{Pick $\uparrow$} & \textbf{HPS $\uparrow$} & \textbf{AeS $\uparrow$} & \textbf{FID $\downarrow$} \\
\midrule
\rowcolor{gray!16}\multicolumn{10}{@{}l}{\textit{SD3-Medium (teacher CFG $=7$)}} \\
\textcolor{gray}{Base Model} & \textcolor{gray}{56} & \textcolor{gray}{0.269} & \textcolor{gray}{0.148} & \textcolor{gray}{0.696} & \textcolor{gray}{0.3534} & \textcolor{gray}{22.516} & \textcolor{gray}{0.2895} & \textcolor{gray}{5.33} & \textcolor{gray}{32.15} \\
DMD & 4 & $0.214_{\pm 0.001}$ & $0.134_{\pm 0.0003}$ & $0.655_{\pm 0.006}$ & $\mathbf{0.3442}_{\pm 0.0011}$ & $\mathbf{22.654}_{\pm 0.015}$ & $\mathbf{0.3094}_{\pm 0.0007}$ & $\underline{5.61}_{\pm 0.01}$ & $37.38_{\pm 0.40}$ \\
DP-DMD & 4 & $\underline{0.282}_{\pm 0.007}$ & $\underline{0.167}_{\pm 0.002}$ & $0.666_{\pm 0.016}$ & $0.3276_{\pm 0.0004}$ & $21.983_{\pm 0.096}$ & $0.2805_{\pm 0.0078}$ & $5.20_{\pm 0.09}$ & $43.29_{\pm 2.12}$ \\
GAD & 4 & $0.234_{\pm 0.002}$ & $0.132_{\pm 0.002}$ & $\underline{0.668}_{\pm 0.017}$ & $0.3398_{\pm 0.0020}$ & $\underline{22.587}_{\pm 0.057}$ & $0.3046_{\pm 0.0003}$ & $5.53_{\pm 0.01}$ & $\underline{35.82}_{\pm 0.39}$ \\
\begin{tabular}[t]{@{}l@{}}\textbf{DMD}\\\textbf{+ ReSeed (ours)}\end{tabular} & 4 & $\mathbf{0.294}_{\pm 0.003}$ & $\mathbf{0.168}_{\pm 0.001}$ & $\mathbf{0.721}_{\pm 0.002}$ & $\underline{0.3404}_{\pm 0.0007}$ & $22.488_{\pm 0.012}$ & $\underline{0.3047}_{\pm 0.0003}$ & $\mathbf{5.62}_{\pm 0.01}$ & $\mathbf{33.48}_{\pm 0.03}$ \\
\rowcolor{gray!16}\multicolumn{10}{@{}l}{\textit{SD3.5-Medium (teacher CFG $=4.5$)}} \\
\textcolor{gray}{Base Model} & \textcolor{gray}{80} & \textcolor{gray}{0.283} & \textcolor{gray}{0.154} & \textcolor{gray}{0.697} & \textcolor{gray}{0.3589} & \textcolor{gray}{22.479} & \textcolor{gray}{0.2851} & \textcolor{gray}{5.30} & \textcolor{gray}{28.18} \\
DMD & 4 & $0.236_{\pm 0.003}$ & $0.132_{\pm 0.004}$ & $0.668_{\pm 0.014}$ & $\underline{0.3505}_{\pm 0.0004}$ & $\underline{22.734}_{\pm 0.019}$ & $\mathbf{0.3040}_{\pm 0.0023}$ & $\underline{5.523}_{\pm 0.020}$ & $32.51_{\pm 0.19}$ \\
DP-DMD & 4 & $\underline{0.325}_{\pm 0.005}$ & $\underline{0.167}_{\pm 0.002}$ & $\underline{0.717}_{\pm 0.011}$ & $0.3455_{\pm 0.0012}$ & $22.444_{\pm 0.018}$ & $0.2915_{\pm 0.0007}$ & $5.51_{\pm 0.02}$ & $\mathbf{29.12}_{\pm 0.62}$ \\
GAD & 4 & $0.234_{\pm 0.003}$ & $0.126_{\pm 0.002}$ & $0.642_{\pm 0.009}$ & $\mathbf{0.3523}_{\pm 0.0008}$ & $\mathbf{22.771}_{\pm 0.021}$ & $\underline{0.2990}_{\pm 0.0020}$ & $5.518_{\pm 0.036}$ & $31.18_{\pm 0.19}$ \\
\begin{tabular}[t]{@{}l@{}}\textbf{DMD}\\\textbf{+ ReSeed (ours)}\end{tabular} & 4 & $\mathbf{0.328}_{\pm 0.006}$ & $\mathbf{0.169}_{\pm 0.003}$ & $\mathbf{0.723}_{\pm 0.007}$ & $0.3446_{\pm 0.0010}$ & $22.398_{\pm 0.008}$ & $0.2956_{\pm 0.0014}$ & $\mathbf{5.533}_{\pm 0.024}$ & $\underline{29.30}_{\pm 0.81}$ \\
\bottomrule
\end{tabular}
}
\end{table}

\paragraph{Main comparison.} Against the DMD students, ReSeed raises DINOv3
diversity from $0.2142$ to $0.2936$ ($+37.1\%$) on SD3-Medium and from
$0.2356$ to $0.3282$ ($+39.3\%$) on SD3.5-Medium; CLIP-feature diversity
and LPIPS also rise (Table~\ref{tab:main}). FID improves from $37.38$ to
$33.48$ and from $32.51$ to $29.30$, respectively. PickScore and HPS fall
by $0.7\%$ and $1.5\%$ on SD3-Medium, and by $1.5\%$ and $2.8\%$ on
SD3.5-Medium, while aesthetic scores change little. Thus, diversity and FID
improve, but not every quality measure does. Section~\ref{sec:ablations}
examines the effect of bias strength. Figure~\ref{fig:sd35m-couch-matched}
shows a matched-noise SD3.5-Medium example: the student produces similar
couches, while ReSeed adds variation in their appearance and surroundings.

\paragraph{Comparison with diversity-aware distillation.} On SD3-Medium,
DMD + ReSeed has higher DINOv3 diversity and lower FID than GAD and DP-DMD
(Table~\ref{tab:main}). On SD3.5-Medium it is close to DP-DMD on these
measures, while GAD retains higher CLIP score and PickScore. ReSeed also
improves an already trained GAD student: in a matched seed-42 SD3-Medium
comparison, DINOv3 diversity rises from $0.2324$ to $0.3029$ and FID falls
from $36.05$ to $34.36$. CLIP score, PickScore, HPS, and aesthetic score
decrease. ReSeed can therefore complement GAD, but retains a
quality--diversity trade-off.

\paragraph{Cross-architecture results on FLUX.1-dev.}
\label{sec:senseflow}

\begin{figure}[!htb]
\centering
\includegraphics[width=0.25\textwidth]{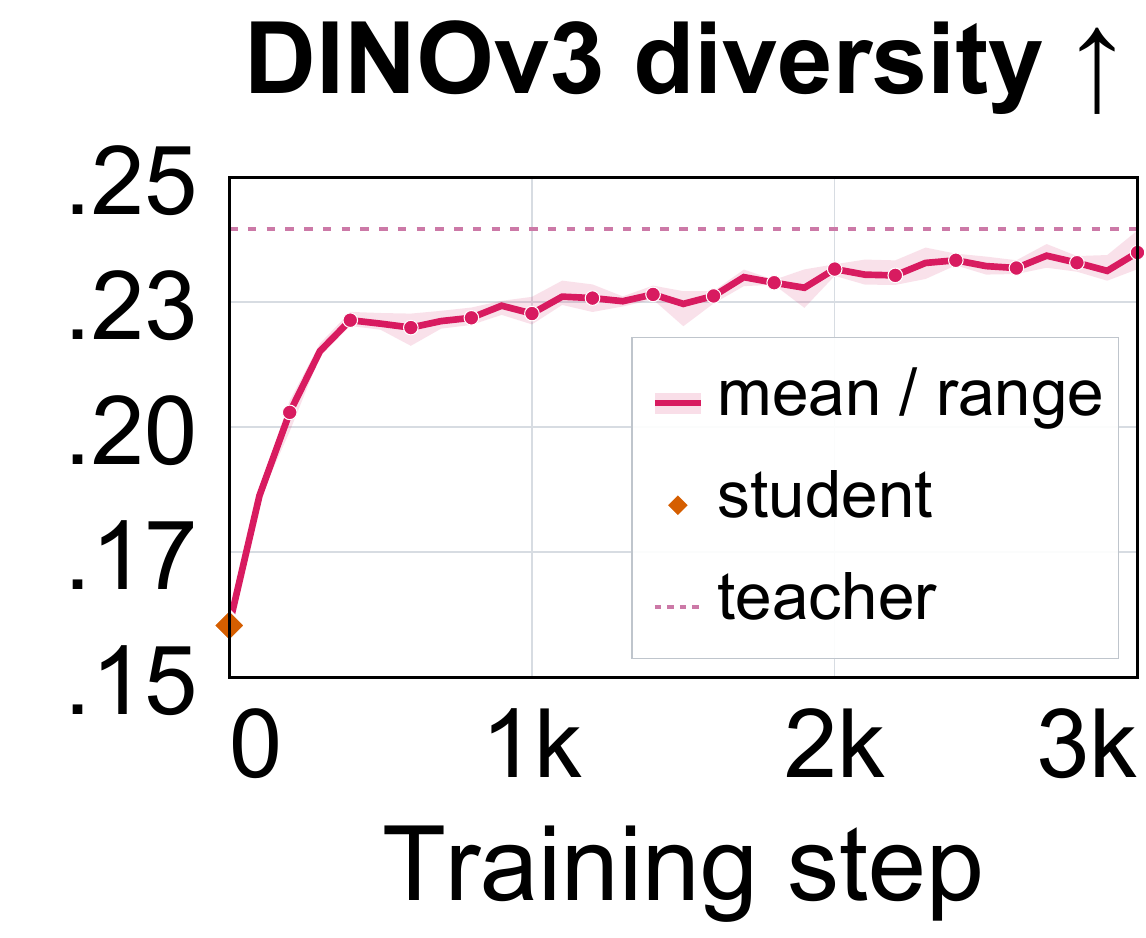}%
\includegraphics[width=0.25\textwidth]{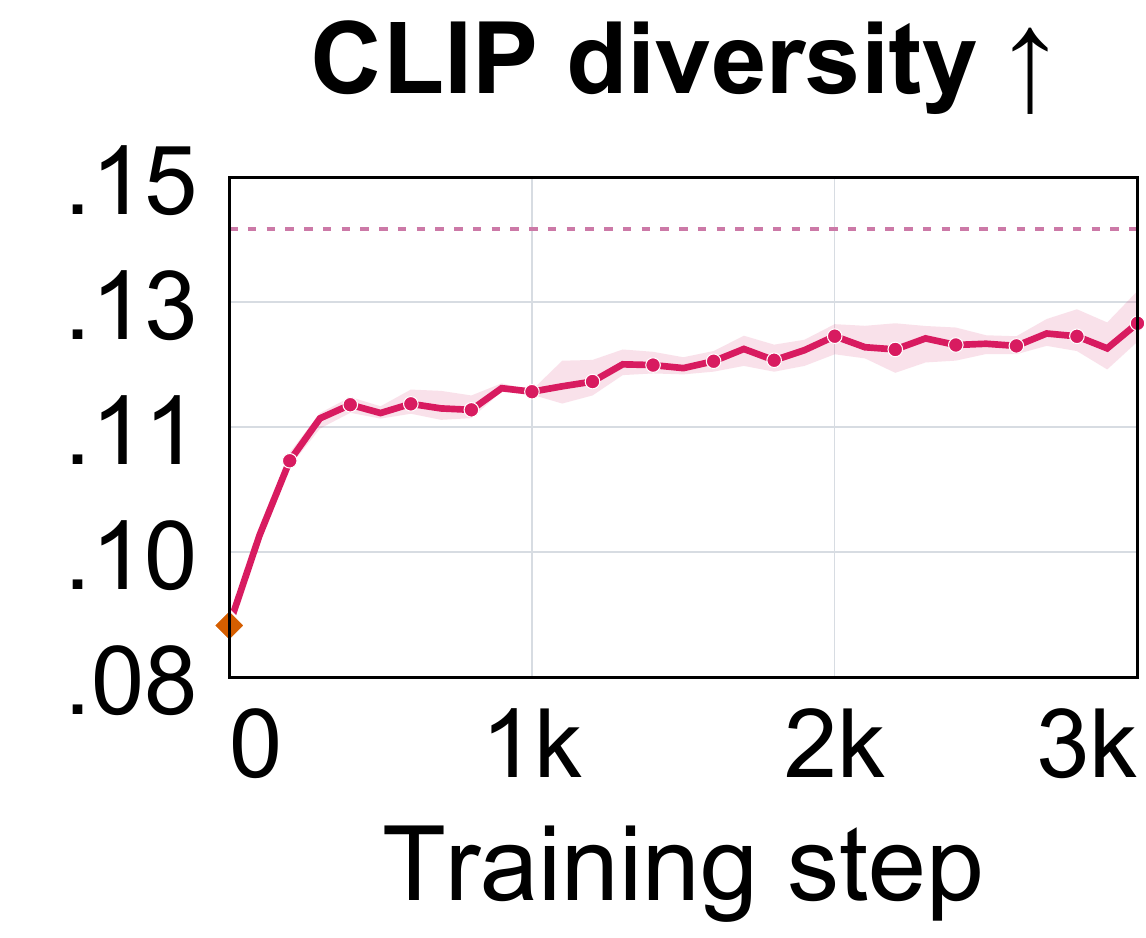}%
\includegraphics[width=0.25\textwidth]{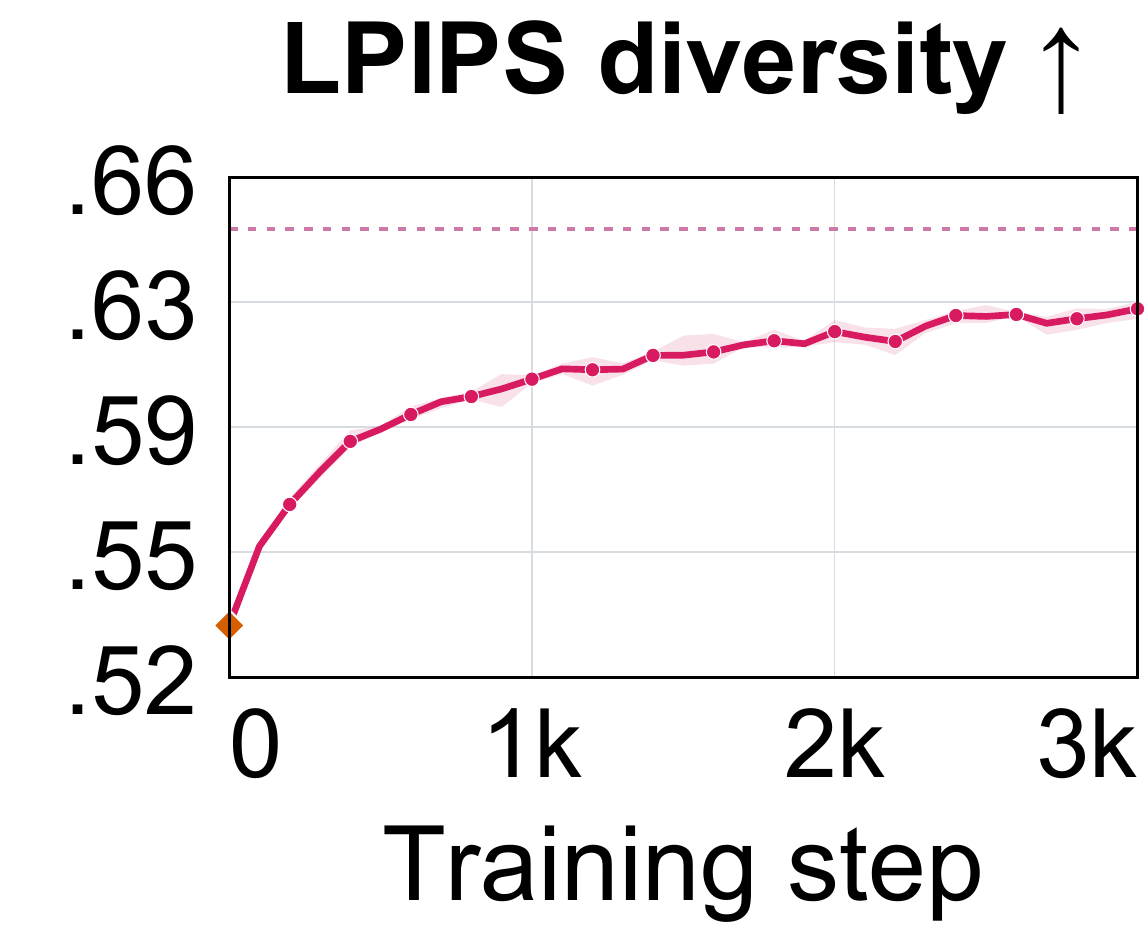}%
\includegraphics[width=0.25\textwidth]{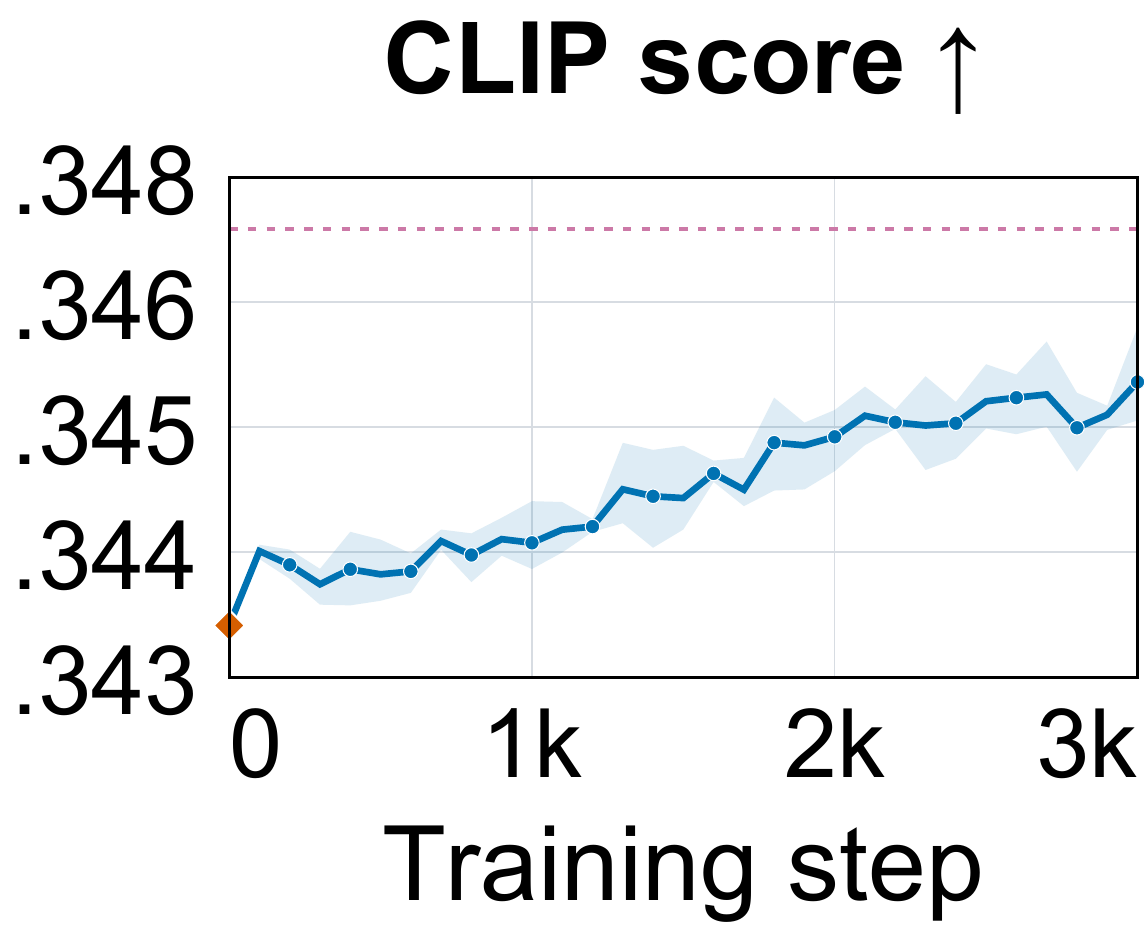}\\[0.7ex]
\includegraphics[width=0.25\textwidth]{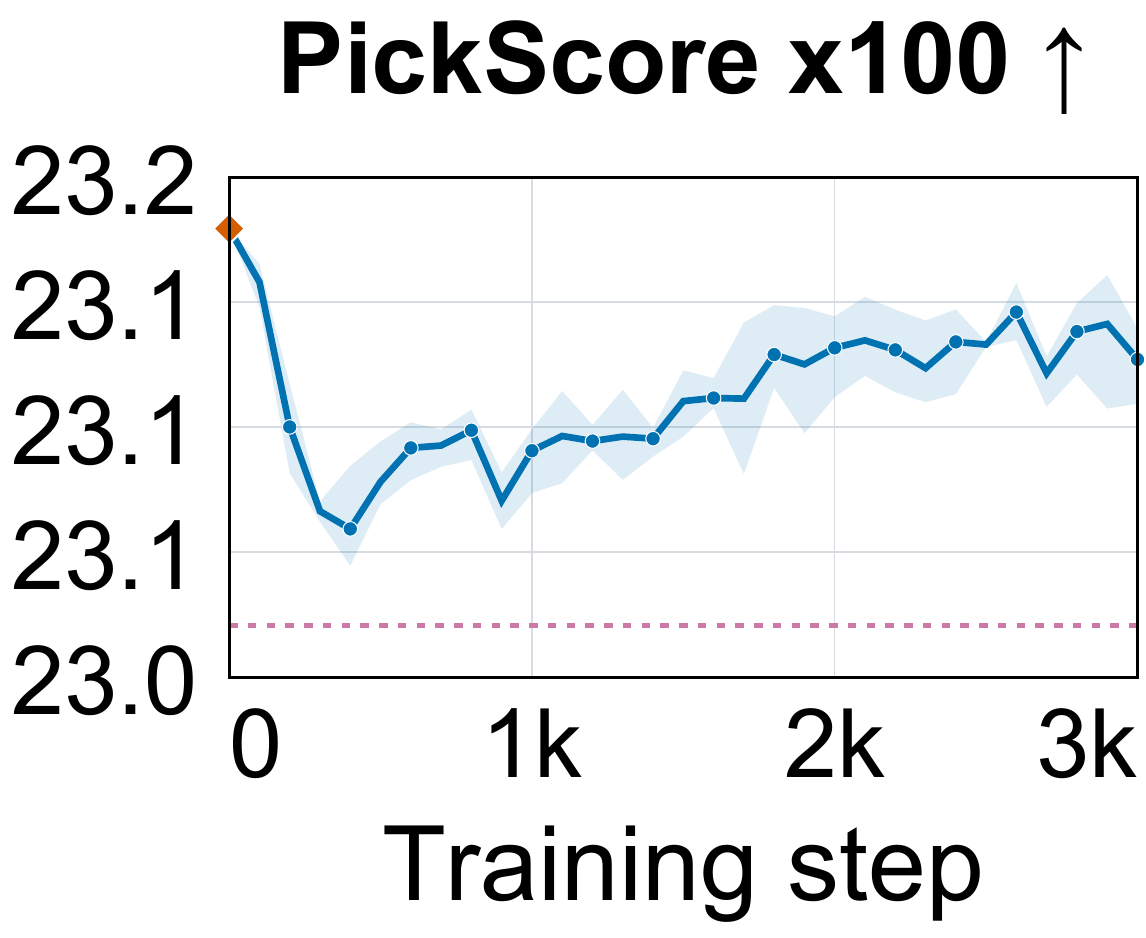}%
\includegraphics[width=0.25\textwidth]{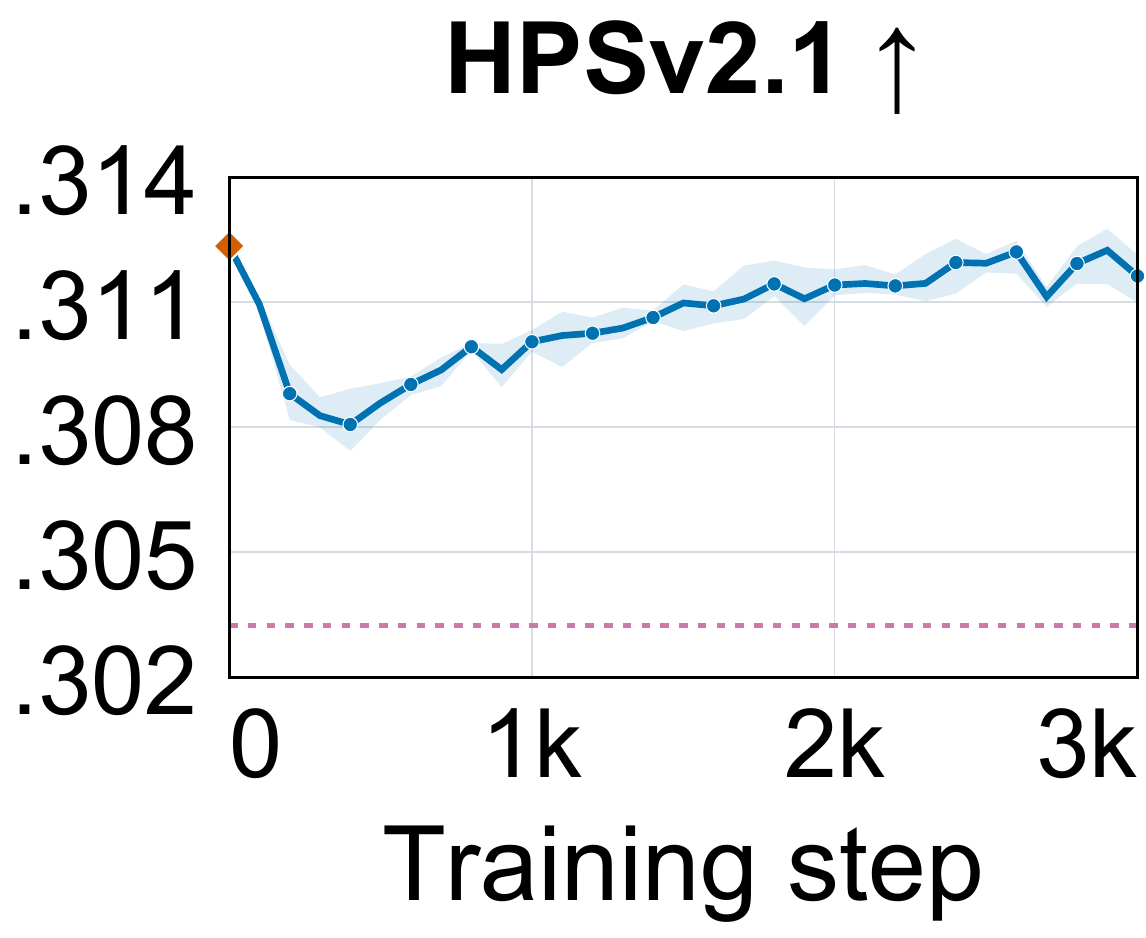}%
\includegraphics[width=0.25\textwidth]{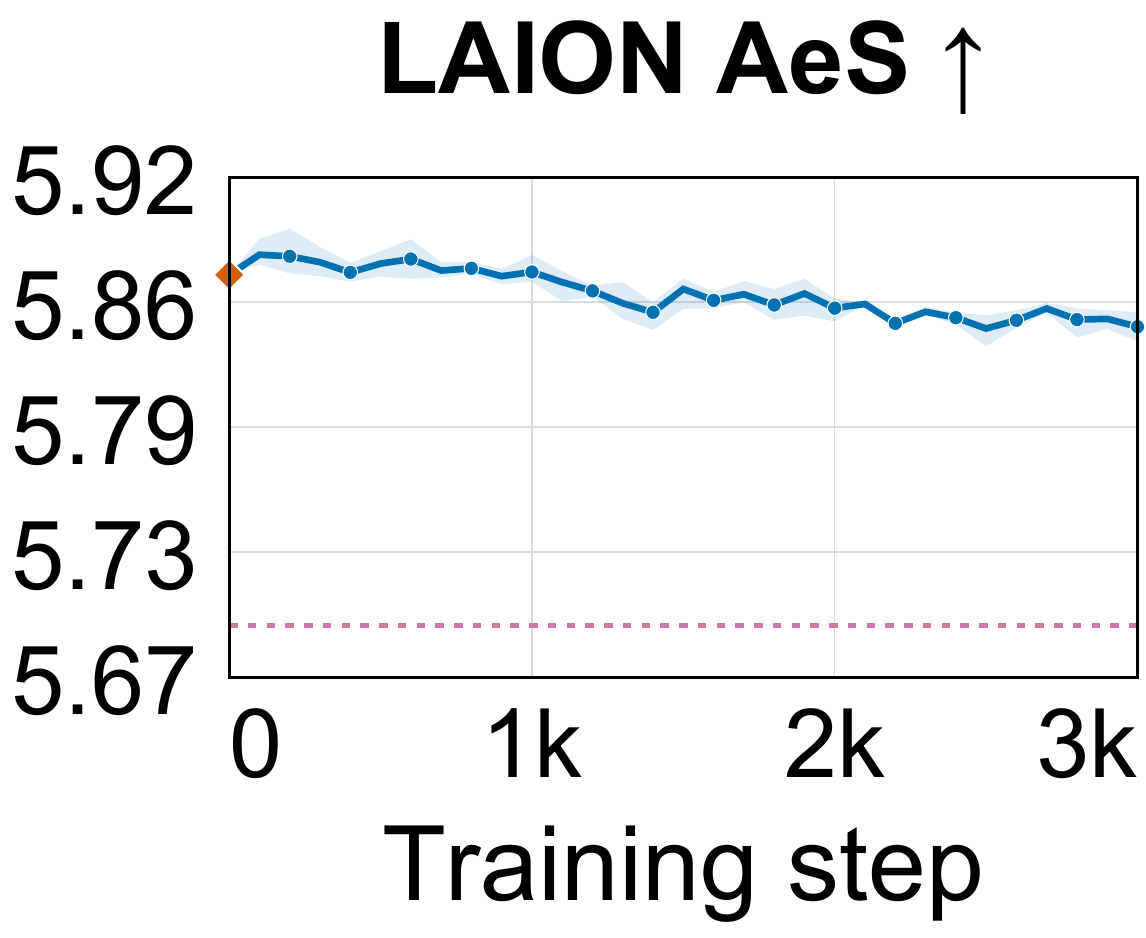}%
\includegraphics[width=0.25\textwidth]{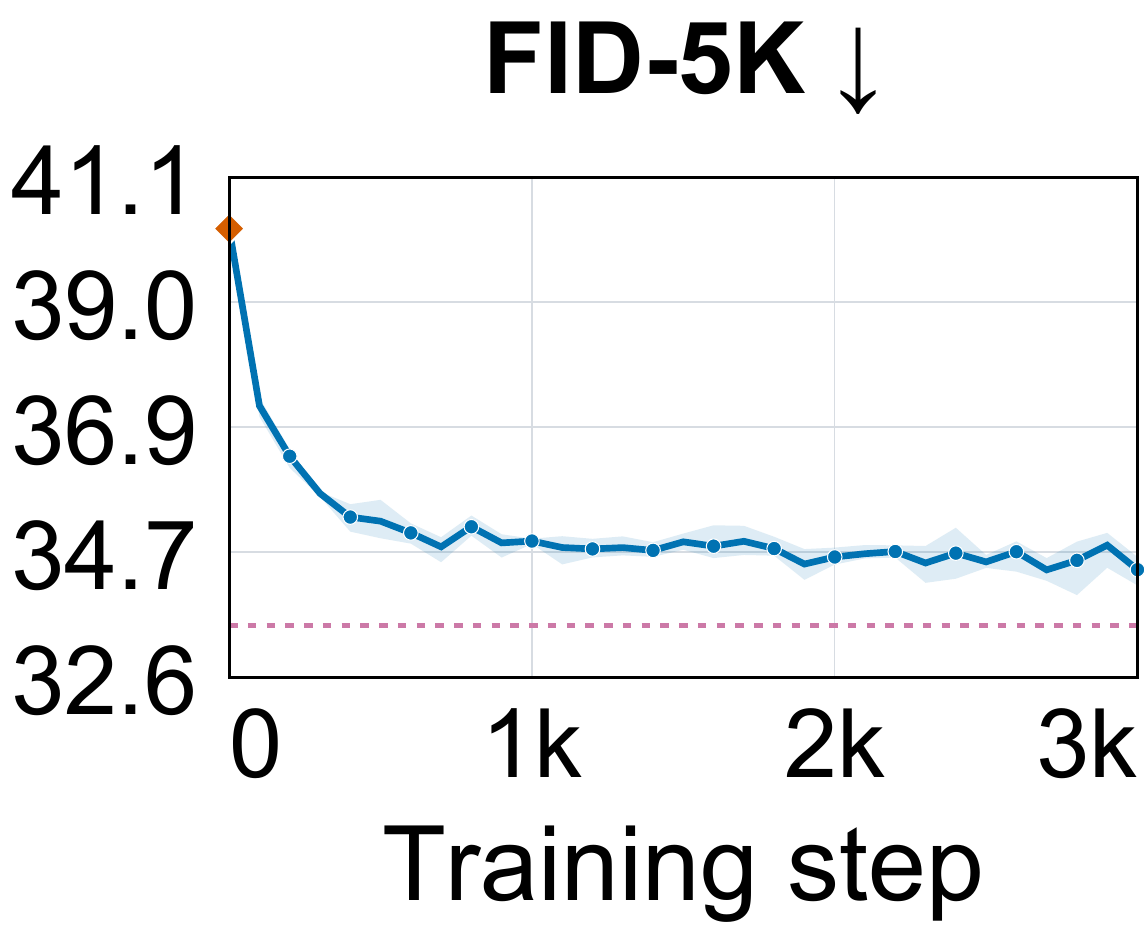}
\caption{4-step FLUX.1-dev student during bias adapter training.}
\label{fig:senseflowtrajectory}
\end{figure}

At 2{,}000 updates, ReSeed increases DINOv3 diversity of the frozen SenseFlow
student \citep{ge2026senseflow} from $0.1575$ to $0.2336$
($+48.3\%$), recovering $89.8\%$ of the measured gap to the FLUX.1-dev
teacher \citep{blackforestlabs2024flux}. CLIP-feature diversity and LPIPS
increase by $40.3\%$ and $16.1\%$, respectively. FID improves from $40.22$
to $34.65$. Unlike the SD3 results, CLIP score also improves, from $0.3431$
to $0.3450$, while PickScore, HPS, and the aesthetic score decrease slightly
(Table~\ref{tab:senseflow}).
The training curves in Figure~\ref{fig:senseflowtrajectory} show that most
DINOv3 diversity recovery occurs within the first 300--400 updates, with
smaller gains afterwards. Figure~\ref{fig:senseflow-perprompt}
in Appendix~\ref{app:sd3-training-trajectories} shows that the diversity gains occur across most
evaluated prompts.

\begin{table}[htbp]
\centering
\fontsize{9.5}{11}\selectfont
\setlength{\tabcolsep}{0.8pt}
\renewcommand{\arraystretch}{1.12}
\caption{FLUX quality--diversity comparison.}
\label{tab:senseflow}
\begin{tabular}{@{}l|c|ccc|ccccc@{}}
\toprule
\multirow{2}{*}{\textbf{Model}} & \multirow{2}{*}{\textbf{NFEs}}
& \multicolumn{3}{c|}{\textbf{Diversity}}
& \multicolumn{5}{c}{\textbf{Quality}} \\
& & \textbf{DINO $\uparrow$} & \textbf{CLIP dist. $\uparrow$} & \textbf{LPIPS $\uparrow$}
& \textbf{CLIP score $\uparrow$} & \textbf{Pick $\uparrow$} & \textbf{HPS $\uparrow$} & \textbf{AeS $\uparrow$} & \textbf{FID $\downarrow$} \\
\midrule
\textcolor{gray}{FLUX.1-dev} & \textcolor{gray}{28} & \textcolor{gray}{0.242} & \textcolor{gray}{0.140} & \textcolor{gray}{0.649} & \textcolor{gray}{0.3472} & \textcolor{gray}{23.050} & \textcolor{gray}{0.3033} & \textcolor{gray}{5.70} & \textcolor{gray}{33.49} \\
SenseFlow & 4 & $0.158$ & $0.090$ & $0.533$ & $0.3431$ & $\mathbf{23.152}$ & $\mathbf{0.3128}$ & $\mathbf{5.87}$ & $40.22$ \\
\shortstack[l]{\textbf{SenseFlow}\\\textbf{+ ReSeed (ours)}} & 4
& \shortstack{$\mathbf{0.234}$\\{\scriptsize $\mathbf{\pm 0.001}$}}
& \shortstack{$\mathbf{0.126}$\\{\scriptsize $\mathbf{\pm 0.002}$}}
& \shortstack{$\mathbf{0.618}$\\{\scriptsize $\mathbf{\pm 0.003}$}}
& \shortstack{$\mathbf{0.3450}$\\{\scriptsize $\mathbf{\pm 0.0003}$}}
& \shortstack{$23.121$\\{\scriptsize $\pm 0.011$}}
& \shortstack{$0.3118$\\{\scriptsize $\pm 0.0003$}}
& \shortstack{$5.85$\\{\scriptsize $\pm 0.01$}}
& \shortstack{$\mathbf{34.65}$\\{\scriptsize $\mathbf{\pm 0.15}$}} \\
\bottomrule
\end{tabular}
\end{table}

\subsection{Extension to Text-to-Video Generation}
\label{sec:text-to-video}

We also test ReSeed on a distilled
video generator. Table~\ref{tab:video-main} compares the Wan2.1 teacher, the
released 4-step DMD2 and DFD checkpoints, and
the same frozen DMD2 student with ReSeed. We evaluate them on the 320-prompt VBench subset described in
Section~\ref{sec:setup}

ReSeed raises video-level DINO diversity from
$0.3374$ to $0.4290$, a relative gain of $27.1\%$. The paired increase is
$+0.0916$, with a $95\%$ bootstrap interval of $[+0.0804,+0.1024]$, and
273 of 320 prompts improve. This recovers $56.2\%$ of the measured
teacher--student DINO gap, although the teacher remains higher at $0.5005$.
Mean DINO diversity increases in all eight categories
(Appendix~\ref{app:video-results}). Video-level CLIP diversity rises from
$0.1267$ to $0.1437$, and per-frame DINO diversity from $0.3748$ to $0.4716$.
The released DFD checkpoint has similar DINO diversity to DMD2 ($0.3358$
versus $0.3374$), but higher projected-CLIP diversity ($0.1351$).
Figure~\ref{fig:wan-filmstrips} in Appendix~\ref{app:grids} shows
matched-noise frame strips for four prompts.
Subject and background consistency stay close
to DMD2, while the flickering, smoothness, and aesthetic scores increase.
However, imaging quality falls from $0.698$ to $0.675$, and dynamic degree
from $0.531$ to $0.329$. The latter is closer to the teacher's $0.371$, but
closeness alone does not show that motion quality improves. CLIP text
alignment falls from $0.2419$ to $0.2365$ (about $2.3\%$). Thus, the larger
feature-space diversity comes with measurable costs and does not by itself
establish better coverage of prompt-consistent content. The bootstrap
intervals describe variation across prompts, not across training runs.
The earlier checkpoint study is retained in Appendix~\ref{app:video-results};
inference cost is reported in Section~\ref{sec:overhead}.

\begin{table*}[h]
\centering
\footnotesize
\setlength{\tabcolsep}{1.8pt}
\renewcommand{\arraystretch}{1.16}
\caption{Wan2.1-T2V-1.3B on 320 VBench prompts. Bold marks the best 4-step result}
\label{tab:video-main}
\resizebox{\textwidth}{!}{%
\begin{tabular}{@{}lc|ccccccc|c|cccc@{}}
\toprule
\multirow{2}{*}{\textbf{Method}} & \multirow{2}{*}{\textbf{Steps}}
& \multicolumn{7}{c|}{\textbf{Video quality (VBench)}}
& \multirow{2}{*}{\shortstack{\textbf{Text}\\\textbf{align.} $\uparrow$}}
& \multicolumn{4}{c}{\textbf{Video diversity}} \\
\cmidrule(lr){3-9}\cmidrule(lr){11-14}
& & \textbf{Subject $\uparrow$} & \textbf{Backg. $\uparrow$}
& \textbf{Flicker $\uparrow$} & \textbf{Smooth. $\uparrow$}
& \textbf{Dynamic $\uparrow$} & \textbf{AeS $\uparrow$}
& \textbf{Imaging $\uparrow$} &
& \shortstack{\textbf{CLIP}\\\textbf{mean} $\uparrow$}
& \shortstack{\textbf{CLIP}\\\textbf{frame} $\uparrow$}
& \shortstack{\textbf{DINO}\\\textbf{mean} $\uparrow$}
& \shortstack{\textbf{DINO}\\\textbf{frame} $\uparrow$} \\
\midrule
\textcolor{gray}{Teacher} & \textcolor{gray}{50}
& \textcolor{gray}{0.970} & \textcolor{gray}{0.961}
& \textcolor{gray}{0.983} & \textcolor{gray}{0.991}
& \textcolor{gray}{0.371} & \textcolor{gray}{0.590}
& \textcolor{gray}{0.703} & \textcolor{gray}{0.2425}
& \textcolor{gray}{0.1634} & \textcolor{gray}{0.1899}
& \textcolor{gray}{0.5005} & \textcolor{gray}{0.5352} \\
\midrule
DMD2 & 4
& \textbf{0.970} & \textbf{0.962} & 0.983 & 0.991
& \textbf{0.531} & 0.615 & 0.698 & \textbf{0.2419}
& 0.1267 & 0.1530 & 0.3374 & 0.3748 \\
DFD & 4
& 0.967 & 0.955 & 0.984 & 0.991
& 0.529 & 0.611 & \textbf{0.711} & 0.2398
& 0.1351 & 0.1659 & 0.3358 & 0.3775 \\
\shortstack[l]{\textbf{DMD2}\\\textbf{+ ReSeed (ours)}} & 4
& 0.967 & 0.961 & \textbf{0.986} & \textbf{0.992}
& 0.329 & \textbf{0.631} & 0.675 & 0.2365
& \textbf{0.1437} & \textbf{0.1733} & \textbf{0.4290} & \textbf{0.4716} \\
\bottomrule
\end{tabular}%
}
\end{table*}

\begin{table}[t]
\centering
\scriptsize
\setlength{\tabcolsep}{3pt}
\caption{ReSeed training and inference overhead.}
\label{tab:cost}
\resizebox{\textwidth}{!}{%
\begin{tabular}{@{}lrrccrccc@{}}
\toprule
 & \multicolumn{2}{c}{\textbf{Adapter}} & \multicolumn{2}{c}{\textbf{DINO diversity}} & \textbf{Peak act.} & \multicolumn{3}{c}{\textbf{Latency (ms)}} \\
\cmidrule(lr){2-3}\cmidrule(lr){4-5}\cmidrule(lr){7-9}
\textbf{Student model} & \textbf{Params} & \textbf{of denoiser} & \textbf{Before $\to$ after} & \textbf{Gain} & \textbf{change} & \textbf{Step 1} & \textbf{Per sample} & \textbf{Overhead} \\
\midrule
SD3-Medium & 72{,}192 & $0.00356\%$ & $0.2142\to0.2936$ & $+37.1\%$ & $+0.00$ MiB & $37.10\!\to\!37.68$ & $148.4\!\to\!149.0$ & $+0.410\%$ \\
SD3.5-Medium & 72{,}192 & $0.00322\%$ & $0.2356\to0.3282$ & $+39.3\%$ & $+0.00$ MiB & $55.69\!\to\!56.25$ & $222.8\!\to\!223.2$ & $+0.194\%$ \\
FLUX.1-dev & 233{,}472 & $0.00196\%$ & $0.1575\to0.2336$ & $+48.3\%$ & $+1.00$ MiB & $253.4\!\to\!255.9$ & $1013.6\!\to\!1016.2$ & $+0.250\%$ \\
Wan2.1-T2V-1.3B & 46{,}080 & $0.00325\%$ & $0.3374\to0.4290$ & $+27.1\%$ & $+0.00$ MiB & $992.7\!\to\!996.1$ & $3978.9\!\to\!3984.3$ & $+0.137\%$ \\
\bottomrule
\end{tabular}
}
\end{table}

\subsection{Training and inference overhead}
\label{sec:overhead}

ReSeed trains only $0.002$--$0.004\%$ of the student denoiser's parameters
(46{,}080--233{,}472; Table~\ref{tab:cost}), excluding text encoders and
autoencoders. The teacher is used only over the first student interval, and
the loss is computed on the first-step velocity, so training needs no later
denoising steps, no decoding, no target images, and no discriminator or
perceptual network. The reported image checkpoints use 1{,}900--2{,}100
updates of four prompt--noise pairs, and Wan2.1 uses 300; most of the gain
appears within the first few hundred updates
(Appendix~\ref{app:overhead-details}). At inference, the bias is added only at
step 1, so the number of function evaluations is unchanged. Per-sample latency
rises by $0.14$--$0.41\%$ and peak activation memory by at most $1$\,MiB.

\subsection{Ablation Studies}
\label{sec:ablations}

\paragraph{Bias strength controls the inference-time trade-off.} We scale
the learned bias by $\alpha$, where $\alpha=0$ gives the original student
and $\alpha=1$ uses the trained bias. Increasing $\alpha$ from $0.5$ to
$1.5$ raises diversity and lowers FID, but reduces CLIP score, PickScore,
and HPS. This allows the quality--diversity balance to be adjusted without
retraining (Table~\ref{tab:doseci}).

\paragraph{Teacher guidance affects the repair target.} We change teacher
guidance from $1$ to $10$ while keeping student inference at CFG $1$.
Higher teacher guidance reduces DINOv3 diversity from $0.3132$ to $0.2873$
and generally increases alignment and preference scores. Our default follows
the student's distillation guidance rather than maximizing diversity.
Full results and training curves are provided in Appendix~\ref{app:ablations}
(Table~\ref{tab:teachercfg} and Figure~\ref{fig:teacher-cfg-ablation}).

\section{Related work}
\label{app:related-work}

\paragraph{Few-step distillation.} Prior approaches progressively halve sampler
steps \citep{salimans2022progressive}, learn consistency maps to trajectory
endpoints \citep{song2023consistency,luo2023lcm}, match output distributions
\citep{yin2024dmd,yin2024dmd2}, match trajectory distributions
\citep{luo2025tdm}, or add adversarial objectives \citep{sauer2023add}.
We study diversity loss in already distilled students.

\paragraph{Diversity-aware diffusion model distillation.} Reduced sample diversity is a known cost of these
methods. Score-identity
distillation studies guidance-dependent diversity and introduces zero- and
anti-guidance objectives \citep{zhou2025fewstepsid}; GAD matches initial-noise
Jacobian--vector products during training \citep{huang2026gad}. At the sampler
level, \citet{gandikota2025distilling} trace part of the loss to premature
structural commitment and recover diversity by using the multi-step base model
for the first denoising step. At the training-schedule level, DP-DMD gives that
step a teacher-derived target-prediction objective while retaining DMD later
\citep{wu2026dpdmd}, and 1.x-Distill separates coarse distribution matching from
detail refinement \citep{li2026onexdistill}. Data-Forcing Distillation instead
performs a short post-training correction using teacher scores on real video
samples \citep{chen2026dataforcing}. CADS improves diversity at inference by
adding noise to the conditioning signal and gradually reducing it during
sampling \citep{sadat2024cads}. ReSeed differs in both question and mechanism:
it diagnoses an already distilled, frozen image generator and changes only the first step of the frozen student without target images, adversarial training, or a
teacher call at inference. We use finite differences only for diagnosis. Vendi
is an alternative set-level diversity summary \citep{friedman2023vendi}.

\paragraph{Parameter-efficient adaptation.} LoRA adds low-rank updates to frozen
weights \citep{hu2022lora} and BitFit tunes biases \citep{benzaken2022bitfit}.
TimeStep Master assigns LoRA experts to different timestep intervals and mixes
them with time-dependent gating \citep{zhuang2025timestepmaster}. Our residual
bias is substantially simpler: it is active only during the first solver interval,
adds no expert router, and serves as both intervention and diagnostic.

\section{Discussion \& Conclusion}

ReSeed repairs the first denoising step of a distilled student while leaving
the later steps unchanged. It adds small residual biases to the frozen
denoiser and trains them to match the teacher's average velocity over the
first high-noise interval. At inference, the biases act only in step 1, so
steps 2--4 use the original student. Training needs no target images or
videos. On SD3-Medium, SD3.5-Medium,
and SenseFlow-FLUX, the adapters add only $0.002$--$0.004\%$ denoiser
parameters and raise DINOv3 diversity by $37.1\%$, $39.3\%$, and $48.3\%$,
respectively, while lowering FID. On Wan2.1-T2V-1.3B DMD2, 300 updates
raise video-level DINOv2 diversity by $27.1\%$, recovering $56.2\%$ of the
measured teacher--student gap. ReSeed thus provides a low-cost
post-distillation correction without retraining the student.

\section{Limitations}
ReSeed training requires multi-step teacher trajectories. Our experiments
use flow-matching transformers; we have not tested non-flow-matching diffusion
models or U-Net backbones, so it remains unclear whether the same first-step
behavior holds for those models.

\section*{Acknowledgments}

This work was supported by the ANT project, funded by the European Union's
Horizon Europe research and innovation programme under Grant Agreement
No.\ 101169439, and by the FFG COMET K1 Centre ``Pro$^{2}$Future II''
(Cognitive and Sustainable Products and Production Systems of the Future;
Contract No.\ 911655). The results presented in this paper were obtained
using the computational resources of the Austrian Scientific
Computing (ASC) infrastructure. Part of this work was carried out during a secondment at AVL List. We thank
Peter Priller for his mentorship and support during this period.

\clearpage

\appendix
\raggedbottom

\section*{Appendix Overview}
\begingroup
\renewcommand{\arraystretch}{1.12}
\newcommand{\appoverviewrow}[1]{%
\ref{#1} & \hyperref[#1]{\nameref*{#1}} & \pageref{#1} \\}
\noindent\begin{tabular*}{\textwidth}{@{}l@{\extracolsep{\fill}}lr@{}}
\toprule
Appendix & Contents & Page \\
\midrule
\appoverviewrow{app:details}
\appoverviewrow{app:full-results}
\appoverviewrow{app:ablations}
\appoverviewrow{app:sd3-training-trajectories}
\appoverviewrow{app:theory}
\appoverviewrow{app:grids}
\bottomrule
\end{tabular*}\par
\endgroup

\section{Implementation and evaluation details}
\label{app:details}

\paragraph{Model and checkpoint sources.}
Table~\ref{tab:checkpoint-sources} lists the sources of the models and checkpoints used in our experiments. The SD3-Medium and SD3.5-Medium DMD, GAD, and
DP-DMD students were trained by us rather than taken from official checkpoint releases.
ReSeed adapters are trained on top of the corresponding frozen students.
For SenseFlow-FLUX, we load the released transformer and retain the text
encoders, autoencoder, and scheduler from FLUX.1-dev, following the release
instructions. The video students are the DMD2 and DFD weights released with
Data-Forcing Distillation.

\begin{table}[!htbp]
\centering
\small
\setlength{\tabcolsep}{4pt}
\renewcommand{\arraystretch}{1.15}
\caption{Model and checkpoint sources. Repository identifiers link to the
public weight releases; locally trained students are listed separately.}
\label{tab:checkpoint-sources}
\begin{tabular}{@{}>{\raggedright\arraybackslash}p{0.23\textwidth}>{\raggedright\arraybackslash}p{0.16\textwidth}>{\raggedright\arraybackslash}p{0.55\textwidth}@{}}
\toprule
\textbf{Model} & \textbf{Source type} & \textbf{Repository or checkpoint} \\
\midrule
SD3-Medium teacher & Public &
\href{https://huggingface.co/stabilityai/stable-diffusion-3-medium-diffusers}{\texttt{stabilityai/stable-diffusion-3-}\newline\texttt{medium-diffusers}} \\
SD3.5-Medium teacher & Public &
\href{https://huggingface.co/stabilityai/stable-diffusion-3.5-medium}{\texttt{stabilityai/}\newline\texttt{stable-diffusion-3.5-medium}} \\
FLUX.1-dev teacher & Public &
\href{https://huggingface.co/black-forest-labs/FLUX.1-dev}{\nolinkurl{black-forest-labs/FLUX.1-dev}} \\
SenseFlow-FLUX student & Public &
\href{https://huggingface.co/domiso/SenseFlow/tree/main/SenseFlow-FLUX}{\nolinkurl{domiso/SenseFlow}}, \nolinkurl{SenseFlow-FLUX/} transformer \citep{ge2026senseflow} \\
Wan2.1-T2V-1.3B teacher & Public &
\href{https://huggingface.co/Wan-AI/Wan2.1-T2V-1.3B-Diffusers}{\texttt{Wan-AI/}\newline\texttt{Wan2.1-T2V-1.3B-Diffusers}} \\
Wan2.1 DMD2 student & Public &
\href{https://huggingface.co/csusupergear/wan_t2v_checkpoints_release}{\texttt{csusupergear/}\newline\texttt{wan\_t2v\_checkpoints\_release}},\newline
\nolinkurl{dmd2_t2v_vipe/0001000} \citep{chen2026dataforcing} \\
Wan2.1 DFD student & Public &
Same release, \nolinkurl{ours_wan_t2v_vipe/0000040} \citep{chen2026dataforcing} \\
SD3-Medium and SD3.5-Medium students & Trained by us &
DMD, GAD, and DP-DMD; training seeds 42, 43, and 44. \\
ReSeed adapters & Trained by us &
First-step bias parameters only; reported training budgets and checkpoint
steps are specified below. \\
\bottomrule
\end{tabular}
\end{table}

\paragraph{Image models and schedule.} SD3-Medium is a rectified-flow
transformer with 24 joint transformer blocks
\citep{esser2024sd3,liu2023rectified,lipman2023flow}. Its 4-step DMD student
is distilled from the 28-step base model with teacher CFG 7
\citep{ho2022cfg,yin2024dmd}. The student uses the shifted flow-matching boundaries
$[1.0,\,0.8984375,\,0.75,\,0.5,\,0]$, obtained from the base boundaries
$[1,\,0.75,\,0.5,\,0.25,\,0]$ with scheduler shift 3. The teacher uses 28 Euler
microsteps, that is seven per student interval. The teacher query grid of the
first interval is
\[
[1.0,\;0.98828125,\;0.9765625,\;0.9609375,\;0.9453125,\;0.9296875,\;0.9140625,\;0.8984375].
\]

The main SD3.5-Medium comparison uses
4-step students at 1024 pixels and a 40-step teacher at CFG 4.5.
The 4-step SenseFlow-FLUX student is paired with the 28-step FLUX.1-dev
teacher. Each model retains its native flow parameterisation and sampler.

\paragraph{SD3-Medium guidance.} The teacher uses guidance scale 7 with the shifted-time
threshold $a=0.96$: below the threshold the guided velocity is
$v^{\mathrm{uncond}}+w(v^{\mathrm{cond}}-v^{\mathrm{uncond}})$, and above it the
conditional velocity is used directly. Here $v^{\mathrm{cond}}$ and
$v^{\mathrm{uncond}}$ are the conditional and unconditional velocity
predictions, $w=7$ is the guidance scale, and $a$ is the noise-level threshold
at which guidance is switched on. The student always runs at guidance
scale 1.

\paragraph{Image-adapter training.} Each student is paired with its corresponding
pretrained teacher and retains its original text encoder and autoencoder.
Teacher targets use the same prompt and initial noise as the student.
Image runs use global batch size 4 and training seeds 42, 43, and 44.
The main results report checkpoints at 1{,}900 updates for SD3-Medium,
2{,}100 for SD3.5-Medium, and 2{,}000 for SenseFlow-FLUX. These correspond
to 7{,}600, 8{,}400, and 8{,}000 prompt--noise pairs per run,
respectively, not necessarily unique prompts. SenseFlow-FLUX training
continues to 3{,}000 updates, with evaluation every 100 updates.

\paragraph{Video sampling.} The Wan2.1-T2V-1.3B teacher uses 50
steps, scheduler shift 5, CFG 5, and the released negative prompt. We use
the released DMD2 checkpoint and DFD
checkpoint. DMD2, DFD, and
DMD2 with ReSeed all sample without CFG at the time boundaries
$t=[0.999,0.937,0.833,0.624,0]$ using the released inference code. For each
generation seed, this code samples one initial noise tensor and reuses it
across prompts. All methods generate 81-frame videos at
$832\times480$, 16 fps, in BF16. Table~\ref{tab:video-main} counts denoising
steps: the teacher uses both conditional and unconditional predictions for
CFG at each step.

\paragraph{Prompt banks and scorers.} Quality uses COCO-2017 validation captions
with 5{,}000 or 1{,}000 fixed prompt and latent pairs. Diversity uses a separate
fixed bank of 100 prompts with 10 independent Gaussian initial noises. The
scorers are a CLIP ViT for the CLIP score and CLIP diversity, PickScore, HPSv2.1,
the LAION aesthetic predictor, LPIPS-Alex and a DINOv3 encoder; FID uses the
standard implementation on matched records. Forward passes are BF16; offline
analysis is float64.
The aesthetic score uses PyIQA's \texttt{laion\_aes} implementation of the
CLIP-based aesthetic predictor.\footnote{\url{https://github.com/christophschuhmann/improved-aesthetic-predictor}}
For image diversity, $R=10$ gives $R(R-1)/2=45$ unordered pairs per prompt.
DINOv3 and CLIP use Equation~\ref{eq:feature-diversity}; LPIPS-Alex uses
the same pairs but averages LPIPS distances rather than cosine distances.
All methods within a comparison use the same prompts, initial noises,
and scoring models.

\paragraph{Training progress and overhead measurements.}
\label{app:overhead-details}
For SD3-Medium, most endpoint-error reduction occurs within 250 updates,
and the alignment statistics plateau after roughly 1{,}000 updates.
The FLUX.1-dev student recovers most DINOv3 diversity within 300--400 updates.
At inference, ReSeed adds one elementwise bias per block only on the first
of four denoising steps. The image models in Table~\ref{tab:cost} are
measured on one H100 at $1024^2$, batch 1, bfloat16, over 100 passes after
20 warm-up passes. The video model uses the same GPU and precision and
reports the median over eight samples of 81 frames at $832\times480$.
Timing reports denoising only. Bias storage is at most $456$\,KiB
($0.002\%$ of student weights). Peak activation memory is unchanged for
SD3-Medium, SD3.5-Medium, and Wan2.1-T2V-1.3B, and increases by $1$\,MiB
for FLUX.1-dev. Parameter percentages use the student denoising transformer
as the denominator, excluding the frozen text encoders and image or video
autoencoder. Image diversity uses DINOv3, whereas video diversity uses
DINOv2-L; absolute values are not directly comparable across modalities.

\subsection{Video training data and evaluation protocol}
\label{app:video-protocol}

\paragraph{Training data and budget.} We use 4{,}000 captions extracted from
the first 40 shards of the \texttt{long} split of the released ViPE dataset
\texttt{csusupergear/vipe\_fliter\_complete}. Only caption text and its
precomputed UMT5 embeddings enter ReSeed training; the associated videos are
not used as targets. Each update samples four prompt--noise pairs on four
H100 GPUs. The 300-update checkpoint therefore uses 1{,}200 prompt--noise pairs.
The full training trajectory contains 2{,}000 updates (8{,}000 prompt--noise pairs),
with checkpoints saved every 100 updates.

\paragraph{Target and optimization.} The teacher starts from the same noise
as the student and takes 13 boundary-aligned Euler microsteps from
$t_0=0.999$ to $t_1=0.937$, with CFG 5 and the released negative prompt.
The microstep grid is uniform in the unshifted time coordinate, then shifted
and aligned to the exact student boundaries. The target is the secant
velocity over this interval, not the teacher's instantaneous prediction or
a decoded video. We fit zero-initialized FP32 bias adapters with AdamW
(learning rate $10^{-3}$, momentum coefficients $0.9$ and $0.999$, zero
weight decay, and gradient-norm clipping at 1). All 30 blocks receive one
1{,}536-channel bias shared over spatial and temporal tokens, for 46{,}080
trainable parameters; the frozen
backbone is evaluated in BF16. Biases are enabled only on the first student
step, with no change to the remaining three steps.
The 13-step Euler training target differs from the 50-step UniPC sampler
used to evaluate the teacher, as described above.

\paragraph{Evaluation prompts and counts.} We take the first 40 unique
prompts from each of the eight VBench category files: animal, architecture,
food, human, lifestyle, plant, scenery, and vehicles. The source is
\texttt{Vchitect/VBench}. A plant prompt
already present in lifestyle is skipped and replaced by the next prompt,
leaving 320 unique prompts. For each
of the teacher, DMD2, DFD, and DMD2 with ReSeed, seeds 0--7 produce
2{,}560 videos.

\section{Full results and statistical analysis}
\label{app:full-results}

\subsection{Additional video results}
\label{app:video-results}

\paragraph{Category-level results.} Table~\ref{tab:video-categories} breaks
down the 320-prompt evaluation. ReSeed increases mean DINO diversity over
DMD2 in every category, but remains below the teacher in every category.
The largest mean increases are in lifestyle and human prompts, and the
smallest is in scenery. These are feature-space distances, not direct counts
of semantic modes. The category intervals are pointwise bootstrap intervals
and are not adjusted for multiple comparisons.

\begin{table}[htbp]
\centering
\small
\setlength{\tabcolsep}{4pt}
\caption{Video-level DINOv2 diversity by VBench category (40 prompts each).
The paired change and the number of improved prompts compare ReSeed with DMD2.}
\label{tab:video-categories}
\begin{tabular}{@{}lrrrrcc@{}}
\toprule
\textbf{Category} & \textbf{Teacher} & \textbf{DMD2} & \textbf{DFD}
& \textbf{+ ReSeed} & \textbf{Change [95\% CI]} & \textbf{Improved} \\
\midrule
Animal       & 0.4235 & 0.2555 & 0.2471 & 0.3313 & $+0.0758\;[+0.053,+0.100]$ & 34/40 \\
Architecture & 0.5948 & 0.4391 & 0.4374 & 0.5278 & $+0.0888\;[+0.056,+0.125]$ & 31/40 \\
Food         & 0.4371 & 0.2797 & 0.2719 & 0.3904 & $+0.1107\;[+0.087,+0.136]$ & 40/40 \\
Human        & 0.4518 & 0.2564 & 0.2635 & 0.3800 & $+0.1236\;[+0.097,+0.152]$ & 37/40 \\
Lifestyle    & 0.4792 & 0.2837 & 0.3117 & 0.4093 & $+0.1255\;[+0.096,+0.155]$ & 38/40 \\
Plant        & 0.5813 & 0.4125 & 0.4146 & 0.4818 & $+0.0693\;[+0.034,+0.104]$ & 31/40 \\
Scenery      & 0.5335 & 0.4128 & 0.3841 & 0.4556 & $+0.0428\;[+0.017,+0.068]$ & 29/40 \\
Vehicles     & 0.5029 & 0.3596 & 0.3558 & 0.4557 & $+0.0961\;[+0.062,+0.133]$ & 33/40 \\
\bottomrule
\end{tabular}%
\end{table}

\paragraph{Paired comparisons and feature choice.} On the full 320-prompt
set, ReSeed's projected-CLIP diversity increase over DMD2 is $+0.0170$
with a $95\%$ interval of $[+0.0119,+0.0221]$; 220 of 320 prompts improve.
Using CLIP's 1{,}024-dimensional pre-projection CLS features instead gives
mean diversity $0.1835$ for DMD2 and $0.2105$ for ReSeed, with a paired
increase of $+0.0270$ and interval $[+0.0211,+0.0333]$. Thus, the direction
of the aggregate gain holds under both feature choices. DFD's DINO
difference from DMD2 is $-0.0016$ with interval $[-0.0090,+0.0059]$,
whereas its projected-CLIP difference is $+0.0084$ with interval
$[+0.0047,+0.0116]$. DFD thus improves projected-CLIP diversity, while its
DINO difference is not distinguishable from zero under this protocol.
ReSeed exceeds DFD in DINO diversity by $0.0932$
with interval $[+0.0830,+0.1035]$, but its text alignment is lower by
$0.0033$ with interval $[-0.0053,-0.0012]$.

\section{Additional ablations}
\label{app:ablations}

\subsection{Teacher guidance}

\begin{figure}[htbp]
\centering
\includegraphics[width=\textwidth]{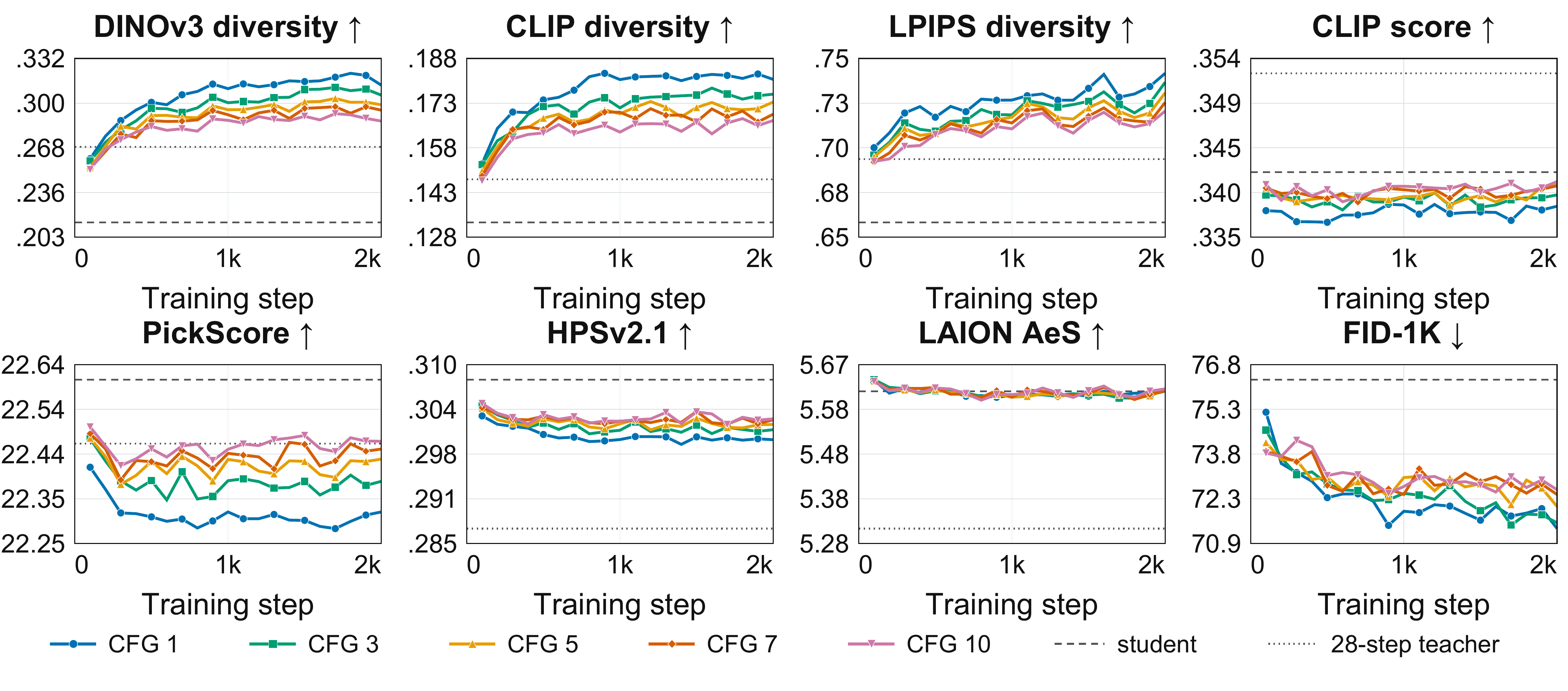}
\caption{Training trajectories for five teacher guidance scales.}
\label{fig:teacher-cfg-ablation}
\end{figure}

\begin{table}[htbp]
\centering
\small
\setlength{\tabcolsep}{5pt}
\caption{Teacher guidance for the first-interval regression target, SD3-Medium,
read at $2{,}000$ updates. Only the guidance used to build the target changes;
the student, its sampler at CFG $1$, the COCO1K prompts, the $100\times10$
diversity bank, the seed and the block range are shared. Bold marks the best
column entry. Diversity falls and alignment generally increases with $w$, and
FID is best at the lowest guidance.}
\label{tab:teachercfg}
\begin{tabular}{@{}c|ccc|ccccc@{}}
\toprule
\multirow{2}{*}{\textbf{Teacher $w$}}
& \multicolumn{3}{c|}{\textbf{Diversity}}
& \multicolumn{5}{c}{\textbf{Quality}} \\
& \textbf{DINO $\uparrow$} & \textbf{CLIP dist. $\uparrow$} & \textbf{LPIPS $\uparrow$}
& \textbf{CLIP score $\uparrow$} & \textbf{Pick $\uparrow$} & \textbf{HPS $\uparrow$}
& \textbf{AeS $\uparrow$} & \textbf{FID-1K $\downarrow$} \\
\midrule
$\mathbf{1}$ & $\mathbf{0.3132}$ & $\mathbf{0.1813}$ & $\mathbf{0.7441}$ & $0.33832$ & $22.318$ & $0.2998$ & $5.617$ & $\mathbf{71.35}$ \\
$3$ & $0.3060$ & $0.1764$ & $0.7391$ & $0.33952$ & $22.384$ & $0.3012$ & $5.616$ & $71.54$ \\
$5$ & $0.2990$ & $0.1737$ & $0.7332$ & $0.34072$ & $22.433$ & $0.3019$ & $5.619$ & $72.08$ \\
$7$ & $0.2951$ & $0.1697$ & $0.7279$ & $0.34049$ & $22.454$ & $0.3026$ & $5.617$ & $72.47$ \\
$10$ & $0.2873$ & $0.1676$ & $0.7230$ & $\mathbf{0.34094}$ & $\mathbf{22.471}$ & $\mathbf{0.3027}$ & $\mathbf{5.621}$ & $72.63$ \\
\bottomrule
\end{tabular}
\end{table}

\subsection{Bias strength and controls}
\label{app:dose}

Table~\ref{tab:doseci} reports COCO5K quality and matched diversity for
inference-time bias scaling. All bias rows use the same seed-42 ReSeed
checkpoint; 4-step student and the teacher are reference models. Channel-shuffled bias
and the blocks 4--23 variant have diversity evaluations only and are not
included in this table.

\begin{table}[htbp]
\centering
\small
\setlength{\tabcolsep}{3.2pt}
\caption{COCO5K quality and diversity.}
\label{tab:doseci}
\begin{tabular}{lcccccccc}
\toprule
Variant & CLIP score & Pick & HPS & AeS & FID $\downarrow$ & DINO $\uparrow$ & CLIPd $\uparrow$ & LPIPS $\uparrow$ \\
\midrule
4-step student (frozen) & 0.34290 & 22.637 & 0.3100 & 5.618 & 37.471 & 0.2142 & 0.1334 & 0.6609 \\
28-step teacher & 0.35343 & 22.516 & 0.2895 & 5.325 & 32.148 & 0.2686 & 0.1478 & 0.6961 \\
$\alpha=-1$ & 0.34346 & 22.597 & 0.3118 & 5.623 & 38.468 & 0.1815 & 0.1230 & 0.5855 \\
$\alpha=0.5$ & 0.34222 & 22.605 & 0.3081 & 5.628 & 35.598 & 0.2545 & 0.1509 & 0.6979 \\
$\alpha=1$ & 0.34051 & 22.471 & 0.3041 & 5.623 & 33.376 & 0.2914 & 0.1698 & 0.7283 \\
$\alpha=1.5$ & 0.33918 & 22.304 & 0.2980 & 5.600 & 31.476 & 0.3190 & 0.1862 & 0.7535 \\
\texttt{random0} & 0.34282 & 22.576 & 0.3081 & 5.623 & 36.646 & 0.2193 & 0.1343 & 0.6569 \\
\bottomrule
\end{tabular}
\end{table}

\section{Training dynamics}
\label{app:sd3-training-trajectories}

\begin{figure}[h]
\centering
\begin{minipage}[t]{0.34\textwidth}
\vspace{0pt}
\centering
\includegraphics[width=\linewidth]{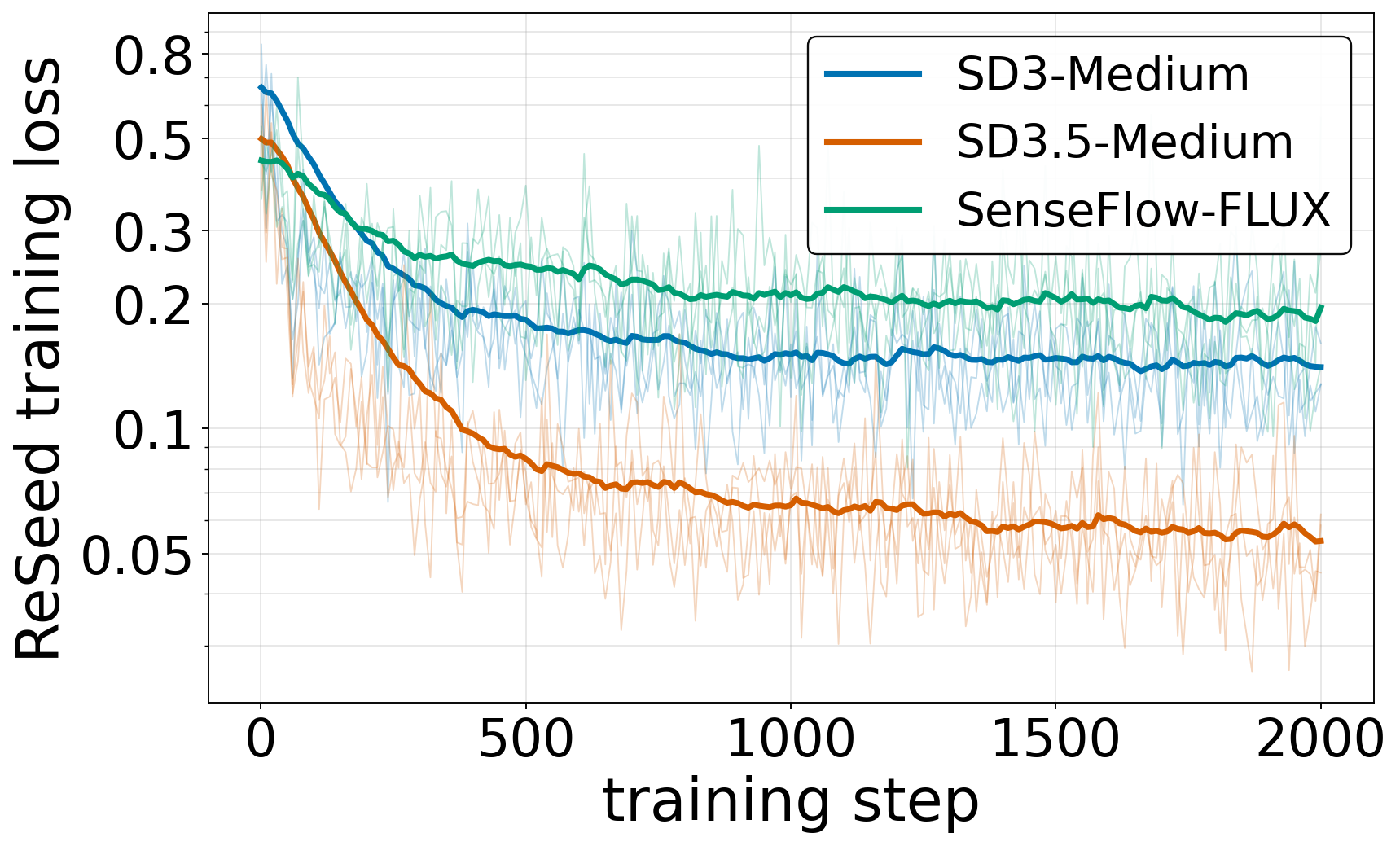}
\end{minipage}\hfill
\begin{minipage}[t]{0.64\textwidth}
\vspace{0pt}
\centering
\includegraphics[width=\linewidth]{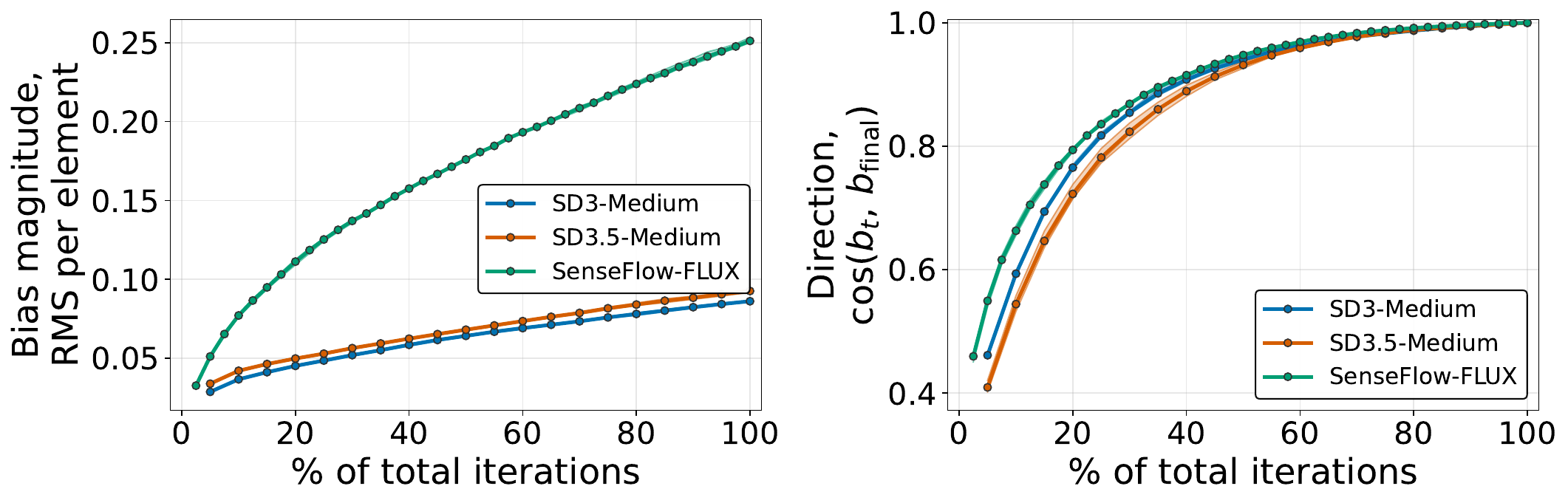}
\end{minipage}
\caption{ReSeed training dynamics. Left: training loss for the three image
models. Middle and right: bias magnitude and cosine with the final bias during training.}
\label{fig:reseed-training-loss}
\label{fig:bias-trajectory}
\end{figure}

The left panel of Figure~\ref{fig:reseed-training-loss} shows the ReSeed training loss for
the three image models. The largest decrease occurs early in training,
followed by a slower decline. These curves describe target fitting, not
final-image quality or diversity, which we evaluate separately below.

\begin{figure}[htbp]
\centering
\includegraphics[width=\textwidth]{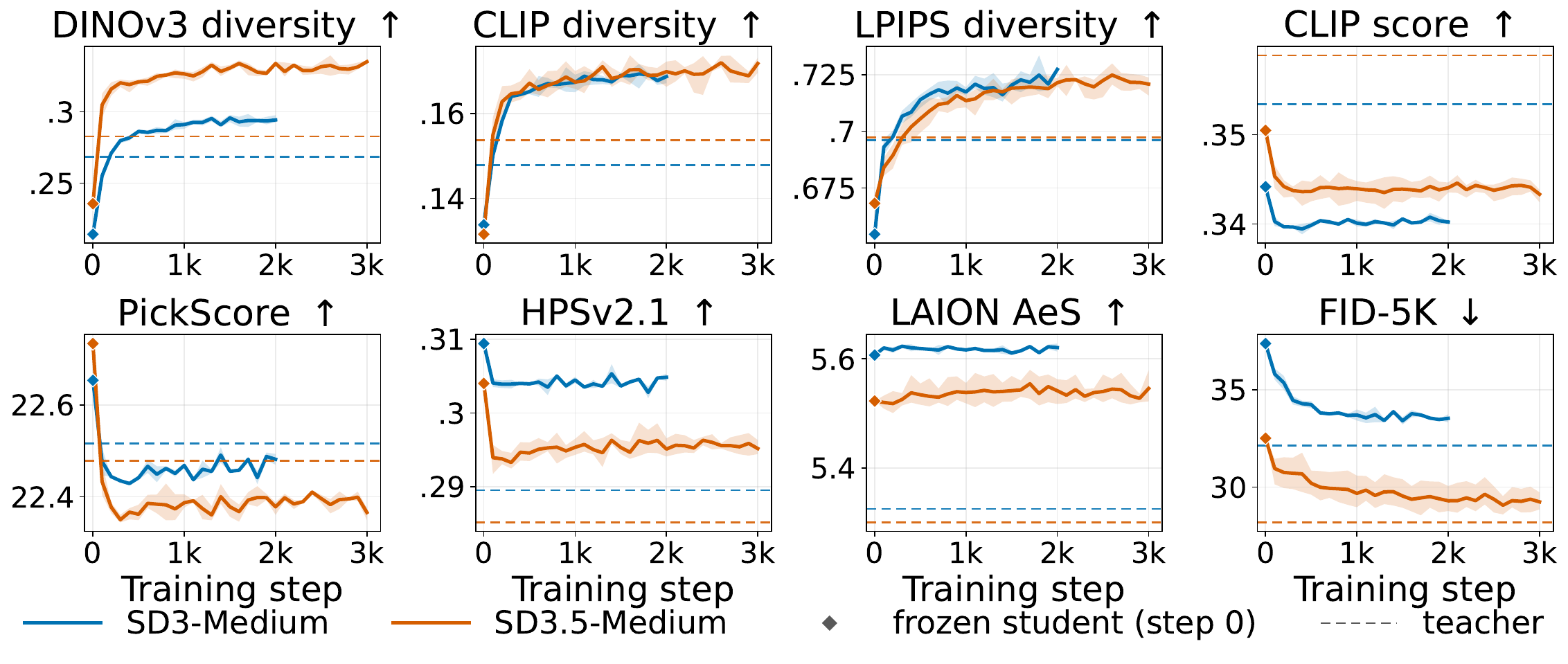}
\caption{SD3-Medium and SD3.5-Medium over ReSeed training. }
\label{fig:sd3-training-trajectories}
\end{figure}

Figure~\ref{fig:sd3-training-trajectories}
shows the training process of SD3-Medium and SD3.5-Medium
Both models were independently
trained with seeds 42, 43, and 44, producing checkpoints every 100 updates. The middle and right panels of
Figure~\ref{fig:bias-trajectory} show the bias norm and its cosine similarity with the final trained bias.

\begin{figure}[htbp]
\centering
\includegraphics[width=\textwidth]{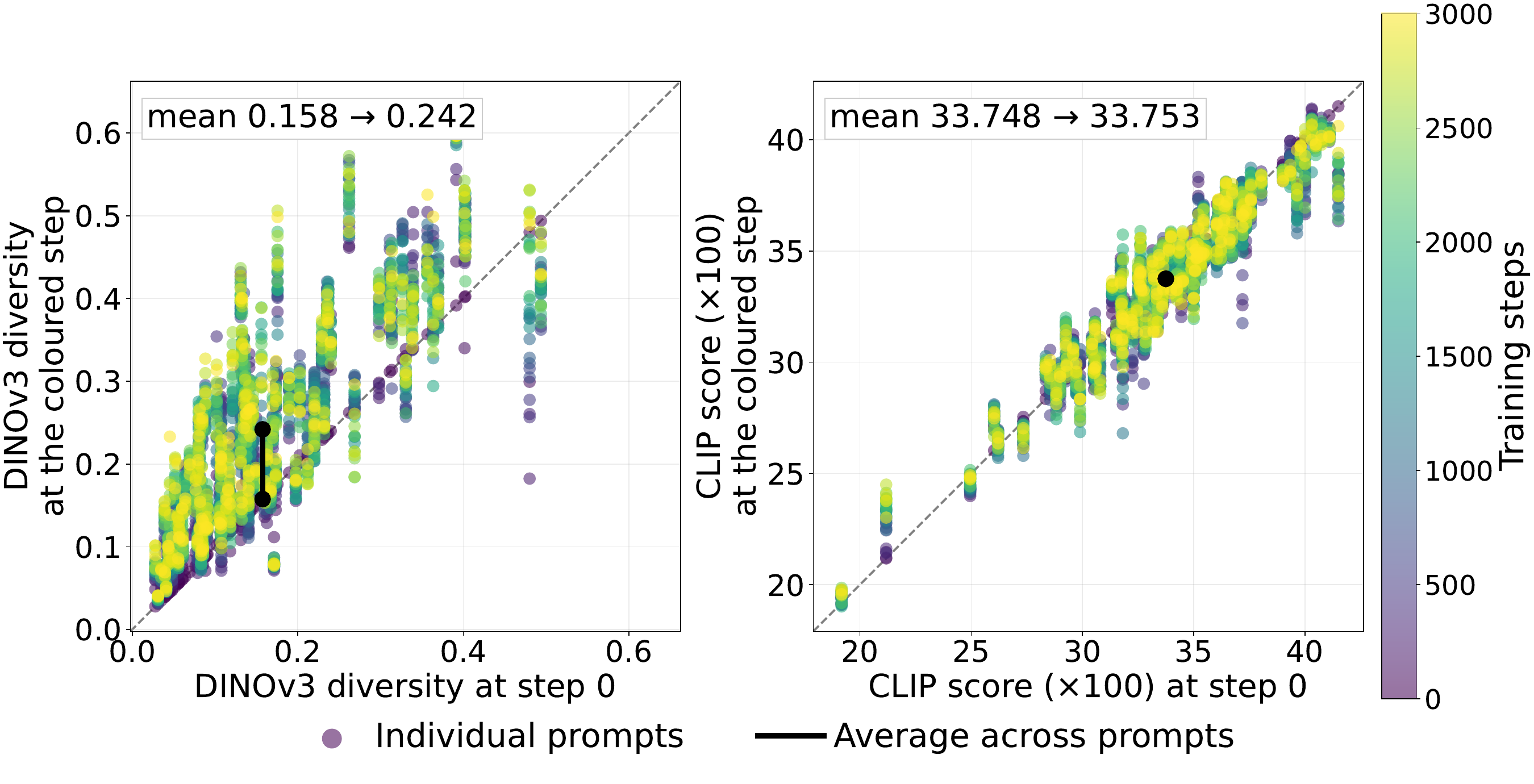}
\caption{Prompt-level trajectories for one FLUX.1-dev adapter on 100 prompts $\times$ 10 initial noises. Each point compares one prompt at
a training checkpoint (vertical axis) with the original student (horizontal
axis); colour denotes the checkpoint and the black line shows the prompt mean.
Left: DINOv3 diversity improves for 95 of 100 prompts, from $0.158$ to $0.242$
on average. Right: CLIP returns to its initial mean
($0.33748\!\rightarrow\!0.33753$).}
\label{fig:senseflow-perprompt}
\end{figure}

Figure~\ref{fig:senseflow-perprompt} shows prompt-level trajectories for the
FLUX.1-dev run with training seed 43, complementing the aggregate curves in
Figure~\ref{fig:senseflowtrajectory}.

\section{Theoretical analysis}
\label{app:theory}

This appendix explains what first-step regression controls and how a shared
bias can change responses to different seeds. We first prove
Proposition~\ref{prop:training-pairwise-bound}, then explain the gap between
the bound and the measured error, including the correction for a finite seed
bank. We next relate these measurements to final-image diversity and finish
with a local analysis of the bias. The mathematical results concern endpoint
responses; the connection to output diversity is tested empirically.

\subsection{From velocity regression to pairwise endpoint error}
\label{app:pairwise-response}

This subsection proves Proposition~\ref{prop:training-pairwise-bound}.
The argument has two steps: velocity regression controls the error of each
first endpoint, and this pointwise error bounds the error between pairs
of endpoints.

\paragraph{Definitions and assumptions.}
Fix a prompt $c$. Let $z,z'$ be independently sampled from the same initial-noise
distribution, and let $\beta$ collect the learned residual biases.
Write $F_\beta(z,c)\in\mathbb{R}^d$ for the adapted student's first endpoint
and $T(z,c)\in\mathbb{R}^d$ for the teacher endpoint reached over the same
interval. Here $d$ is the number of latent elements, and $T$ denotes the
teacher map, not a transpose. Define the endpoint error and its mean over noise:
\[
e(z,c)=F_\beta(z,c)-T(z,c),
\qquad
\bar e(c)=\mathbb{E}_z e(z,c).
\]
Assume $\mathbb{E}_{c,z}\|e(z,c)\|_2^2<\infty$, where
$\|\cdot\|_2$ is the Euclidean norm. Expectations over $c$ use the prompt
distribution in the regression objective. For the same prompt, define
the student and teacher seed responses as
\begin{equation}
\Delta F_\beta=F_\beta(z,c)-F_\beta(z',c),
\qquad
\Delta T=T(z,c)-T(z',c).
\end{equation}
Their mismatch is $\Delta F_\beta-\Delta T=e(z,c)-e(z',c)$.
It measures agreement with the teacher's response, not the size of the
student response alone.

\paragraph{Velocity error determines endpoint error.}
Fix a prompt $c$ and an initial noise $z$. Let $t_0$ and $t_1$ be the
boundaries of the first student interval, and write $\Delta t=t_1-t_0\ne0$.
Let $d$ be the number of scalar elements in a flattened velocity
tensor. We divide squared errors by $d$ to measure error per latent element.
The teacher and student start from the same latent $z$ at $t_0$ and are
compared at the same boundary $t_1$.

The teacher reaches $t_1$ through $K$ denoising steps, with boundaries
$t_0=\tau_0,\tau_1,\ldots,\tau_K=t_1$. Starting from $x^T_0=z$, its
$j$-th update is
\[
x^T_{j+1}=x^T_j+(\tau_{j+1}-\tau_j)
v_{\mathrm{tea}}(x^T_j,\tau_j,c).
\]
Adding these updates gives the teacher endpoint $T(z,c)=x^T_K$:
\[
T(z,c)=z+\sum_{j=0}^{K-1}(\tau_{j+1}-\tau_j)
v_{\mathrm{tea}}(x^T_j,\tau_j,c).
\]
We define the teacher's secant velocity over the entire student interval by
dividing its total displacement by $\Delta t$:
\[
\bar v_{\mathrm{tea}}(z,c)
=\frac{T(z,c)-z}{\Delta t}
=\sum_{j=0}^{K-1}\frac{\tau_{j+1}-\tau_j}{\Delta t}
v_{\mathrm{tea}}(x^T_j,\tau_j,c).
\]
The coefficients in the sum add to one. Thus, $\bar v_{\mathrm{tea}}$
describes the teacher's average velocity across the interval; it is not
the instantaneous velocity at the final microstep.

The adapted student takes one Euler step from the same starting point:
\[
F_\beta(z,c)=z+\Delta t\,v^\beta_{\mathrm{stu}}(z,t_0,c).
\]
Subtracting the teacher endpoint cancels the common starting point $z$.
The endpoint error $e(z,c)=F_\beta(z,c)-T(z,c)$ is therefore
\begin{equation}
\begin{aligned}
e(z,c)
&=\bigl[z+\Delta t\,v^\beta_{\mathrm{stu}}(z,t_0,c)\bigr]
-\bigl[z+\Delta t\,\bar v_{\mathrm{tea}}(z,c)\bigr]\\
&=\Delta t\bigl[v^\beta_{\mathrm{stu}}(z,t_0,c)
-\bar v_{\mathrm{tea}}(z,c)\bigr].
\end{aligned}
\end{equation}
This equality holds for each prompt--noise pair. Squaring both sides and
dividing by $d$ gives
\[
\frac{\|e(z,c)\|_2^2}{d}
=(\Delta t)^2\frac{\bigl\|v^\beta_{\mathrm{stu}}(z,t_0,c)
-\bar v_{\mathrm{tea}}(z,c)\bigr\|_2^2}{d}.
\]
Define the expected velocity loss at a fixed prompt and over the prompt
distribution as
\[
\mathcal{L}_{\mathrm{ReSeed}}(c)
=\mathbb{E}_z\!\left[\frac{\bigl\|v^\beta_{\mathrm{stu}}(z,t_0,c)
-\bar v_{\mathrm{tea}}(z,c)\bigr\|_2^2}{d}\right],
\qquad
\mathcal{L}_{\mathrm{ReSeed}}=\mathbb{E}_c\mathcal{L}_{\mathrm{ReSeed}}(c).
\]
Taking expectations in the preceding equality yields
\[
\frac{1}{d}\mathbb{E}_{c,z}\|e(z,c)\|_2^2
=(t_1-t_0)^2\mathcal{L}_{\mathrm{ReSeed}}.
\]
Thus, expected squared velocity error and expected squared first-endpoint
error differ only by the known factor $(t_1-t_0)^2$. This is an equality
for the expected objective; a finite training average alone does not bound
error on unseen prompts or noises.

\paragraph{Proof of the pairwise bound.}
Fix a prompt $c$ and sample $z$ and $z'$ independently from the same
initial-noise distribution. Recall that
$\Delta F_\beta-\Delta T=e(z,c)-e(z',c)$, and let
$\bar e(c)=\mathbb{E}_z e(z,c)$. Expanding the square and using
independence in one calculation gives
\begin{equation}
\begin{aligned}
\mathbb{E}_{z,z'}\|\Delta F_\beta-\Delta T\|_2^2
&=\mathbb{E}_{z,z'}\|e(z,c)-e(z',c)\|_2^2\\
&=2\mathbb{E}_z\|e(z,c)\|_2^2
-2\bigl\langle\mathbb{E}_z e(z,c),
\mathbb{E}_{z'}e(z',c)\bigr\rangle\\
&=2\bigl(\mathbb{E}_z\|e(z,c)\|_2^2
-\|\bar e(c)\|_2^2\bigr).
\end{aligned}
\label{eq:bound-identity}
\end{equation}
The second line uses independent, identically distributed noises; the last
line is the mean--variance identity. Thus, an endpoint error shared by all
seeds under a prompt cancels from their pairwise response.

Averaging over prompts, dividing by the latent dimension $d$, and using
the velocity-to-endpoint equality above gives
\begin{equation}
\begin{aligned}
\frac{1}{d}\mathbb{E}_{c,z,z'}\|\Delta F_\beta-\Delta T\|_2^2
&=\frac{2}{d}\bigl(
\mathbb{E}_{c,z}\|e(z,c)\|_2^2
-\mathbb{E}_c\|\bar e(c)\|_2^2\bigr)\\
&\leq \frac{2}{d}\mathbb{E}_{c,z}\|e(z,c)\|_2^2\\
&=2(t_1-t_0)^2\mathcal{L}_{\mathrm{ReSeed}}.
\end{aligned}
\label{eq:pairwise-bound-derivation}
\end{equation}
The inequality drops the nonnegative mean-error term. This proves
Proposition~\ref{prop:training-pairwise-bound}.

\subsection{Bound tightness and finite-sample estimates}
\label{app:bound-tightness}

\paragraph{The mean error determines population slack.}
For a fixed prompt with nonzero endpoint error, define
\[
s(c)=\frac{\|\bar e(c)\|_2^2}{\mathbb{E}_z\|e(z,c)\|_2^2}\in[0,1].
\]
This is the fraction of squared error due to the mean error across seeds.
Equation~\ref{eq:bound-identity} shows that the ratio of pairwise response
error to its pointwise upper bound is exactly $1-s(c)$.
If all endpoint errors are equal, $s(c)=1$ and pairwise error is zero.
If the mean error is zero, $s(c)=0$ and the bound is tight.
When endpoint error is zero, both sides vanish and the ratio is undefined.

\paragraph{Finite-sample correction.}
The population identity uses two independent noise samples. In an evaluation
bank of $n$ noises, we instead average over the $n(n-1)/2$ distinct pairs.
For a fixed prompt, let $e_i=e(z_i,c)$, let
$\hat e=n^{-1}\sum_i e_i$ be the sample mean error, and define
\[
\widehat P=\frac{2}{n(n-1)d}\sum_{i<j}\|e_i-e_j\|_2^2,
\qquad
\widehat B=\frac{2}{nd}\sum_i\|e_i\|_2^2.
\]
Here $\widehat P$ is measured pairwise response error and $\widehat B$ is the
empirical counterpart of the regression-based upper bound. The sample identity
$\sum_{i<j}\|e_i-e_j\|_2^2=n\sum_i\|e_i-\hat e\|_2^2$ gives
\begin{equation}
\widehat P
=\frac{n}{n-1}\left(\widehat B-\frac{2}{d}\|\hat e\|_2^2\right)
\leq\frac{n}{n-1}\widehat B.
\label{eq:pairwise-finite-sample}
\end{equation}
Thus, a finite-bank estimate can exceed $\widehat B$ without violating the
population result. With $n=10$, the corrected empirical upper bound is
$10\widehat B/9$. The numerical results below report the uncorrected ratio
$\widehat P/\widehat B$ unless stated otherwise.

\subsection{Empirical experiments and the relation to output diversity}
\label{app:response-diversity-check}

\paragraph{Evaluation design.}
We evaluate both sides of Equation~\ref{eq:pairwise-bound} on 5{,}000 COCO5K
prompts with ten seeds each, using the frozen SD3-Medium student and seven
checkpoints from one ReSeed run (100--1{,}900 updates). This gives
40{,}000 prompt--checkpoint comparisons. Table~\ref{tab:prop1-coco5k}
reports the endpoint statistics and final-image feature diversity. These
are measured in different spaces and answer different questions.

\paragraph{Consistency and tightness of the bound.}
All measured pairwise errors satisfy the finite-sample bound in
Equation~\ref{eq:pairwise-finite-sample} with $n=10$. This confirms that
the two errors were computed consistently. The median ratio
$\widehat P/\widehat B$ rises from $0.68$ to $0.92$ during training,
as the estimated seed-shared share of endpoint error falls from
$0.39$ to $0.17$ (Table~\ref{tab:prop1-coco5k}).
We next examine whether the reduction in pairwise response error
accompanies a change in final-image diversity.

\paragraph{Association with final-image diversity.}
Mean pairwise response error decreases by a factor of $2.7$,
with a decrease on every prompt. Mean DINOv3 diversity rises from $0.213$
to $0.290$ ($+36.5\%$), with gains on $92\%$ of prompts; as the bound's
right-hand side falls, this diversity rises
(Table~\ref{tab:prop1-coco5k}).

Figure~\ref{fig:prop1-response-diversity} shows the relation between pairwise
response error and final-image diversity. Across the frozen student and seven
trained checkpoints, lower mean error consistently accompanies higher mean
diversity (Spearman $\rho=-1$). Comparing checkpoints within each prompt,
halving the response error is associated with a $0.056$ increase in DINOv3
diversity (95\% CI $[0.054, 0.057]$, prompt-clustered bootstrap). The
within-prompt Spearman correlation is $-0.54$ (CI $[-0.55, -0.53]$), and
the association is negative for $90\%$ of prompts. These results show an association, not a causal effect of response
error on final diversity.

\begin{table}[h]
\centering
\small
\caption{Pairwise response error, its bound, and final-image diversity on
COCO5K ($5{,}000$ prompts $\times$ 10 seeds) for one SD3-Medium ReSeed run.
Step 0 is the DMD 4-step student. LHS and RHS are the empirical pairwise error
$\widehat P$ and uncorrected bound $\widehat B$, averaged over prompts and
reported in units of $10^{-3}$. LHS/RHS and the estimated mean-error share $s$ are
medians over prompts.}
\label{tab:prop1-coco5k}
\begin{tabular}{rcccccc}
\toprule
Step & LHS & RHS & LHS/RHS & $s$ & DINOv3 div.\ $\uparrow$ & CLIP div.\ $\uparrow$ \\
\midrule
0     & 5.70 & 8.53 & 0.68 & 0.39 & 0.213 & 0.122 \\
100   & 3.09 & 3.66 & 0.85 & 0.23 & 0.253 & 0.139 \\
200   & 2.74 & 3.08 & 0.90 & 0.19 & 0.267 & 0.145 \\
300   & 2.60 & 2.93 & 0.89 & 0.20 & 0.277 & 0.149 \\
400   & 2.46 & 2.73 & 0.91 & 0.18 & 0.279 & 0.150 \\
700   & 2.28 & 2.49 & 0.92 & 0.17 & 0.281 & 0.152 \\
1{,}000 & 2.20 & 2.40 & 0.92 & 0.17 & 0.287 & 0.154 \\
1{,}900 & 2.10 & 2.29 & 0.92 & 0.17 & 0.290 & 0.155 \\
\bottomrule
\end{tabular}
\end{table}

\begin{figure}[t]
\centering
\includegraphics[width=0.6\textwidth]{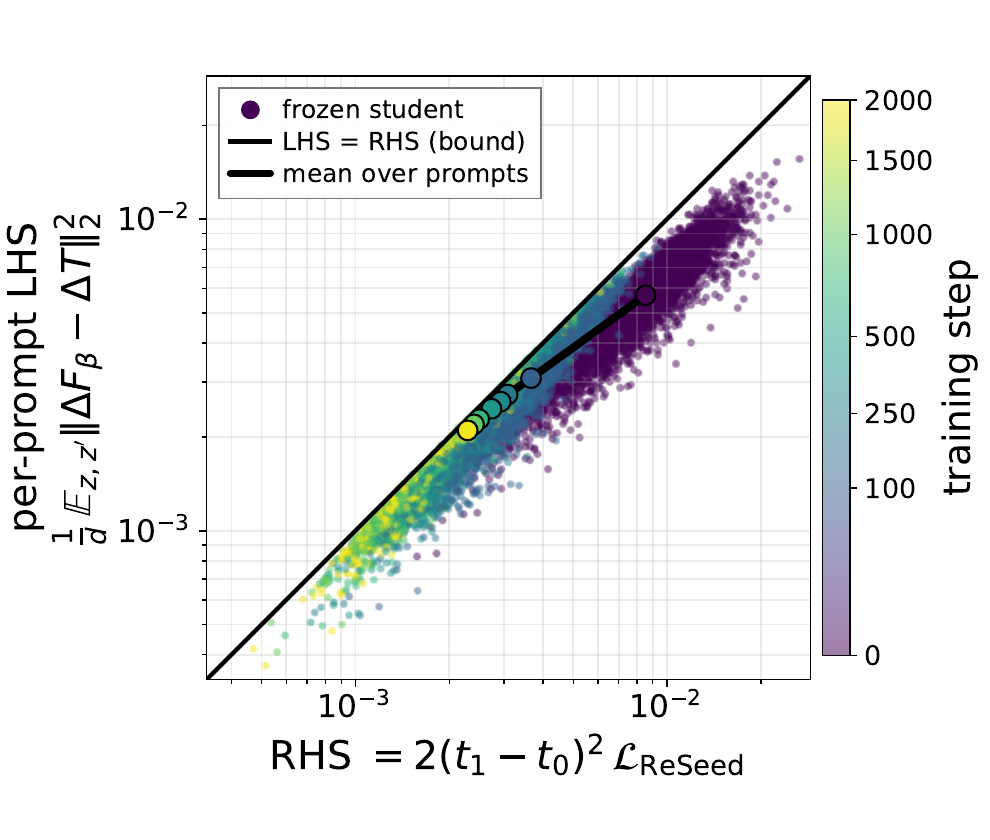}
\caption{Empirical analysis of Proposition~\ref{prop:training-pairwise-bound}
on COCO5K. Each point is one prompt at one checkpoint, coloured by training
step; step 0 is the frozen student. The black curve shows per-step means over
prompts. The left-hand side of Equation~\ref{eq:pairwise-bound} is plotted
against its right-hand side $2(t_1-t_0)^2\mathcal{L}_{\mathrm{ReSeed}}$.
The diagonal marks $\widehat P=\widehat B$; the exact finite-sample
bound is given in Equation~\ref{eq:pairwise-finite-sample}.
The median LHS/RHS rises from $0.68$ to $0.92$ during training
(Table~\ref{tab:prop1-coco5k}).}
\label{fig:prop1-bound-analysis}
\end{figure}

\subsection{Why a shared bias can change seed-dependent responses}

A constant added directly to every endpoint cancels from pairwise differences.
ReSeed instead adds its biases inside the network. The following local result
explains why the same parameter change can then produce different endpoint
changes for different seeds. It describes a possible mechanism, not a guarantee
that an arbitrary bias improves teacher alignment or diversity.

\begin{proposition}[An input-dependent bias Jacobian changes seed responses]
\label{prop:bias-jacobian}
Fix a prompt $c$ and write $G(z;\beta)=F_\beta(z,c)\in\mathbb{R}^d$.
Let $\beta\in\mathbb{R}^m$ collect the $m$ bias parameters, with $\beta=0$
denoting the original frozen student. Assume that
$G(z;\beta)$ is twice continuously differentiable in $\beta$ near zero and that
its second derivative with respect to $\beta$ is uniformly bounded over the
inputs considered. Define
$J_\beta(z)=\partial G(z;\beta)/\partial\beta\rvert_{\beta=0}
\in\mathbb{R}^{d\times m}$, the sensitivity of the endpoint to the bias
parameters at zero bias. Then, for any
two seeds $z,z'$ and sufficiently small $\beta$,
\begin{equation}
\begin{aligned}
&\bigl[G(z;\beta)-G(z';\beta)\bigr]
-\bigl[G(z;0)-G(z';0)\bigr] \\
&\qquad =\bigl[J_\beta(z)-J_\beta(z')\bigr]\beta
+O(\lVert\beta\rVert_2^2).
\end{aligned}
\label{eq:bias-jacobian}
\end{equation}
For a given bias direction, the first-order change vanishes when
$[J_\beta(z)-J_\beta(z')]\beta=0$; this can occur even if the two Jacobians
differ. It vanishes for every direction if and only if the Jacobians are equal.
The term $O(\|\beta\|_2^2)$ denotes a remainder whose norm is bounded by a
constant times $\|\beta\|_2^2$ near zero.
\end{proposition}

\paragraph{Proof.} Taylor expansion in $\beta$ at each fixed input gives
$G(z;\beta)=G(z;0)+J_\beta(z)\beta+O(\lVert\beta\rVert_2^2)$ and the analogous
expression for $z'$. Subtracting the two expansions gives
Equation~\ref{eq:bias-jacobian}. The uniform Hessian bound makes the remainder
uniform over the inputs considered. 

\paragraph{Interpretation and limits.}
The same bias is used for
every seed, but its output correction need not be the same because
$J_\beta(z)$ depends on the prompt, noise and activation state. In a jointly
affine map this Jacobian is constant and the bias cancels exactly from every
seed difference. In the nonlinear student, the leading correction is instead
potentially first order in $\beta$ and is controlled by the difference of two input-specific
Jacobians. If two inputs are close, that Jacobian difference can be described
locally by a mixed derivative, but independent Gaussian seeds need not be close;
the result instead expands in the small parameter change $\beta$, not in the
distance between seeds. It does not establish that the local approximation
remains accurate at the final trained bias, or that its direction will reduce
response error. Those are empirical questions.

In summary, Proposition~\ref{prop:training-pairwise-bound} identifies what the
regression objective controls, and Proposition~\ref{prop:bias-jacobian}
explains how shared internal biases can change seed differences.
The observed link to final-image diversity comes from the matched evaluations
above and the teacher-prefix intervention, not from either proposition alone.

\newgeometry{textheight=9.4in,textwidth=6.5in,top=0.6in,
  headheight=14pt,headsep=25pt,footskip=30pt}
\section{Qualitative comparisons}
\label{app:grids}

\begin{figure}[!htbp]
\centering
\includegraphics[width=0.97\textwidth]{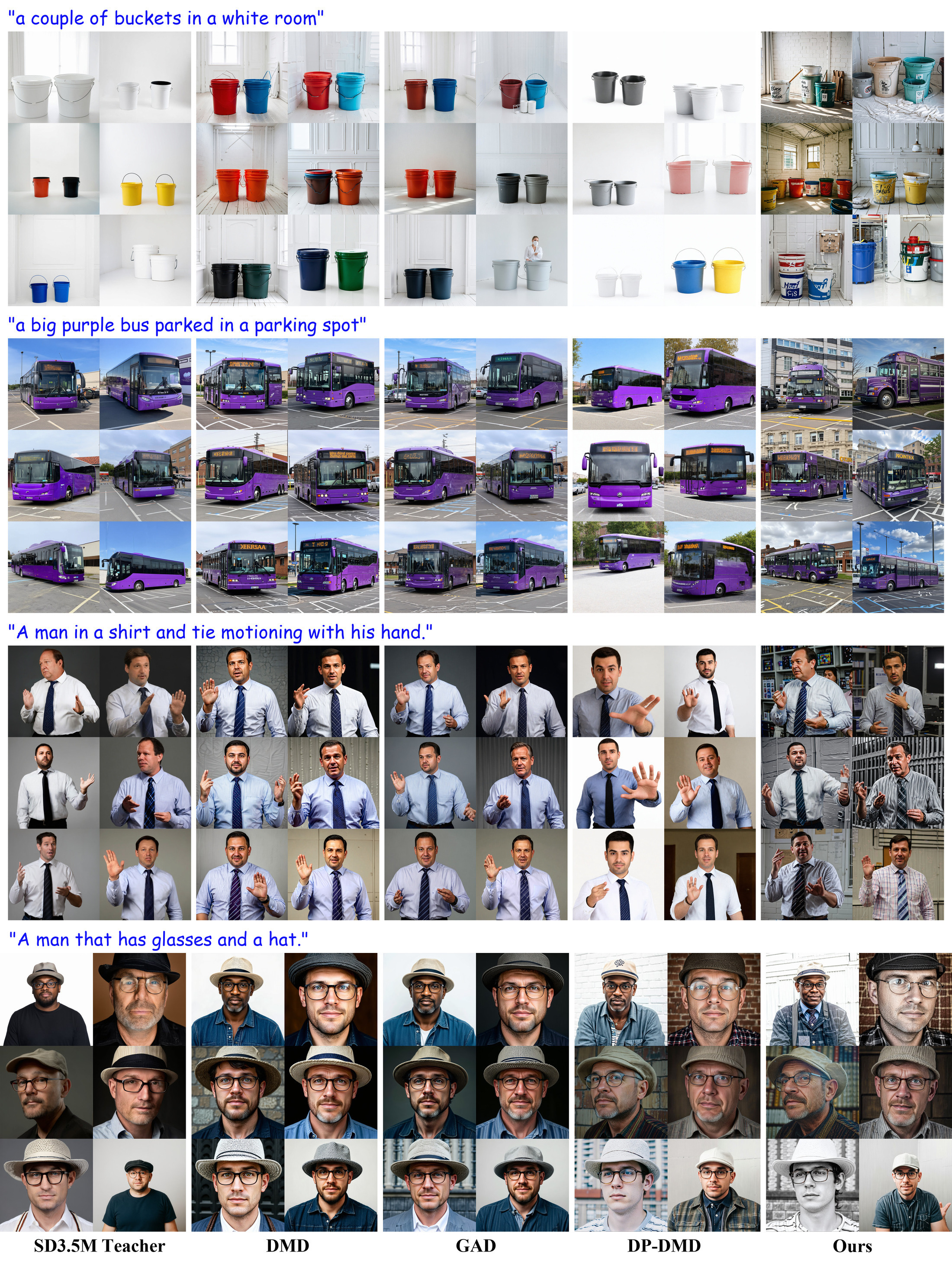}
\caption{Matched-noise SD3.5-Medium comparisons. Each method uses
the same six seeds. The figure continues on the next page.}
\label{fig:sd35m-additional-diversity}
\end{figure}
\clearpage

\begin{figure}[p]
\ContinuedFloat
\centering
\includegraphics[width=0.97\textwidth]{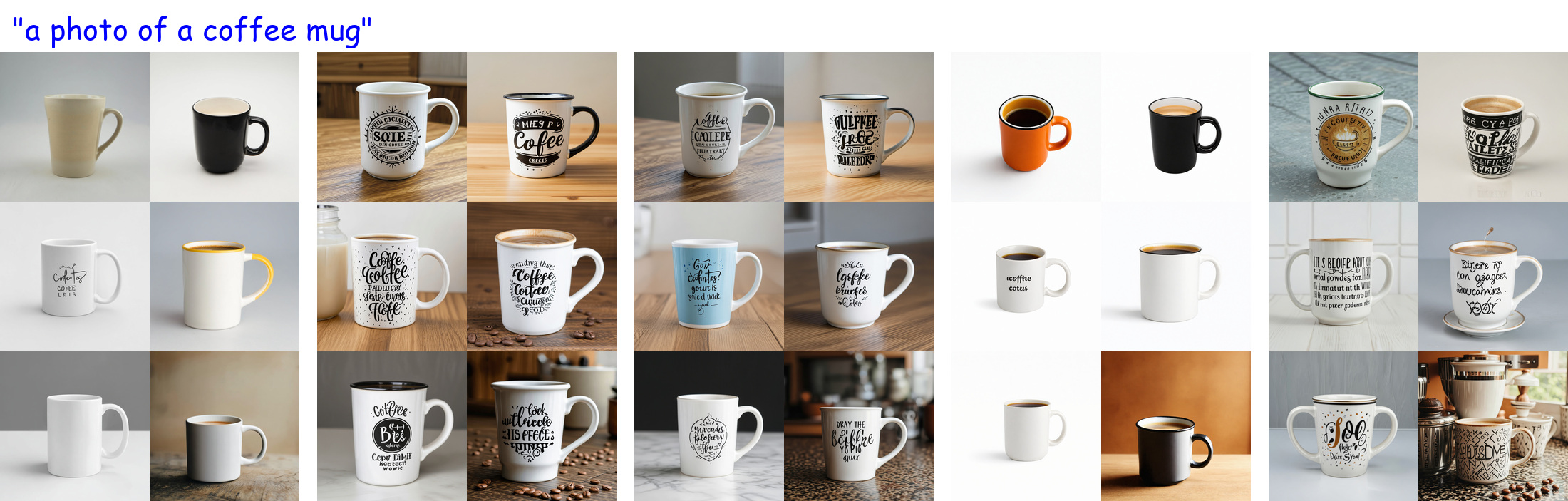}\\[0.12em]
\includegraphics[width=0.97\textwidth]{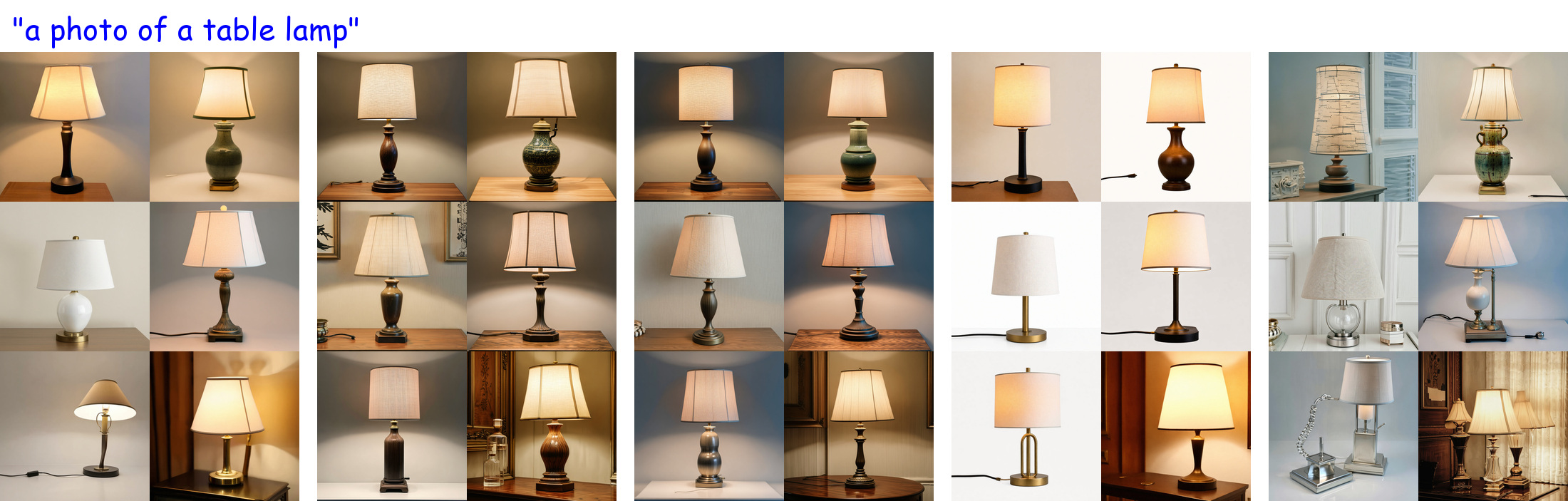}\\[0.12em]
\includegraphics[width=0.97\textwidth]{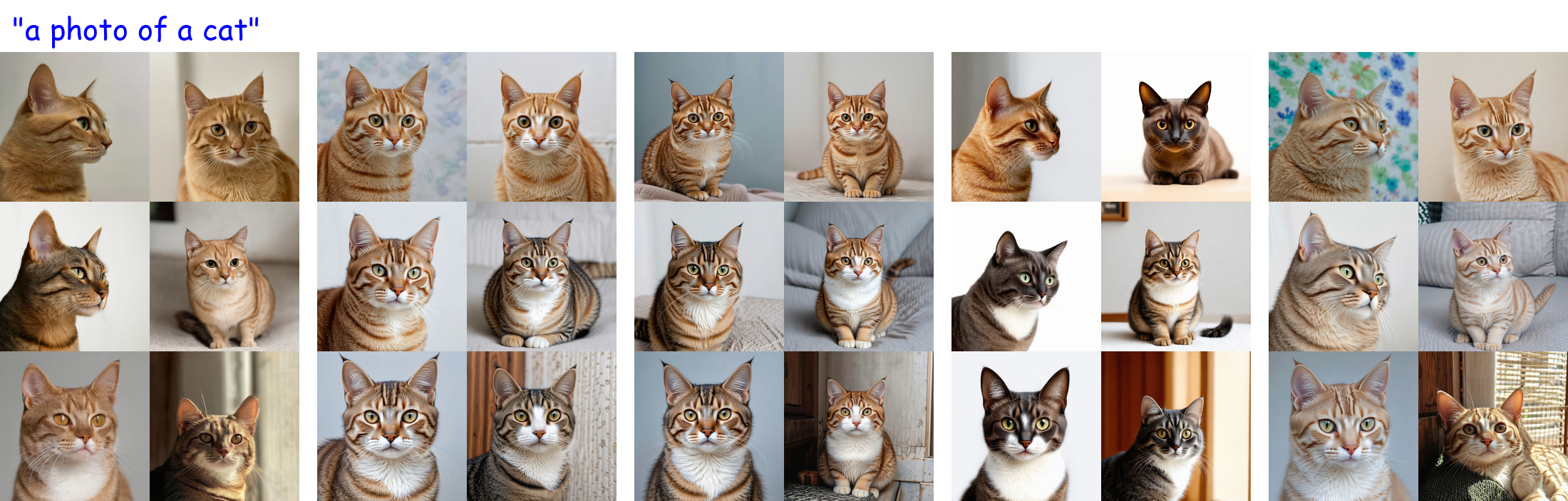}\\[0.12em]
\includegraphics[width=0.97\textwidth]{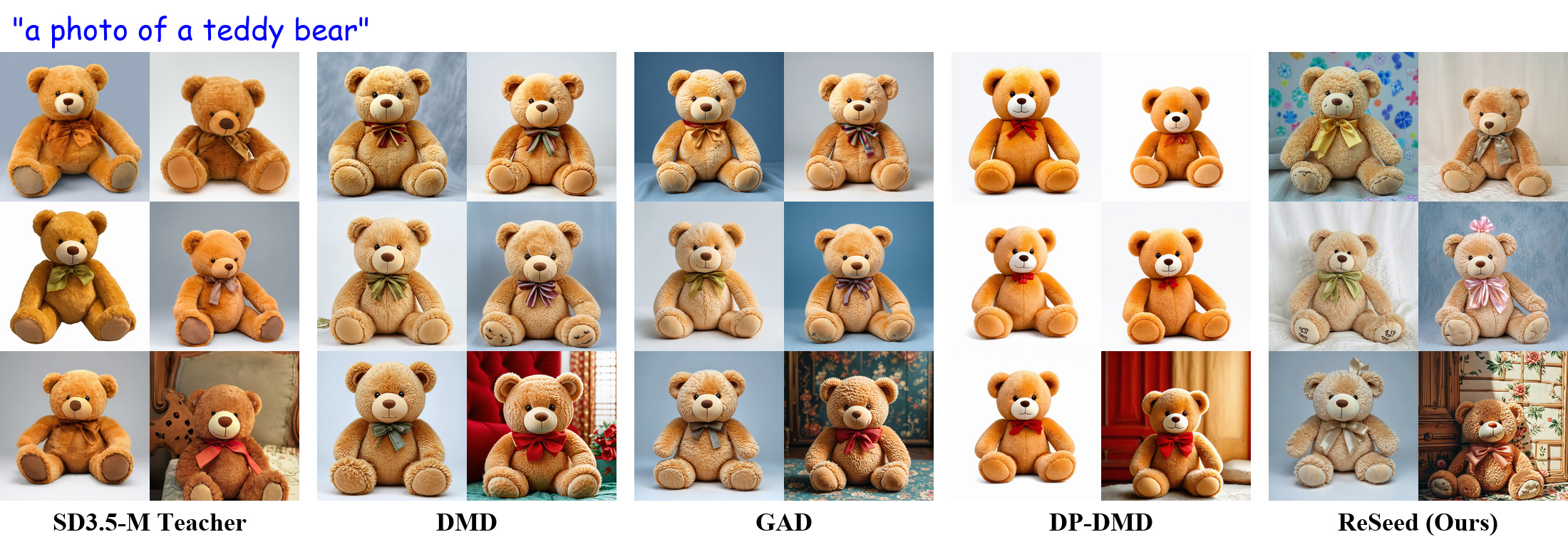}
\end{figure}
\newgeometry{textheight=9in,textwidth=6.5in,top=1in,
  headheight=14pt,headsep=25pt,footskip=30pt}

\begin{figure}[p]
\centering
\includegraphics[width=0.97\textwidth]{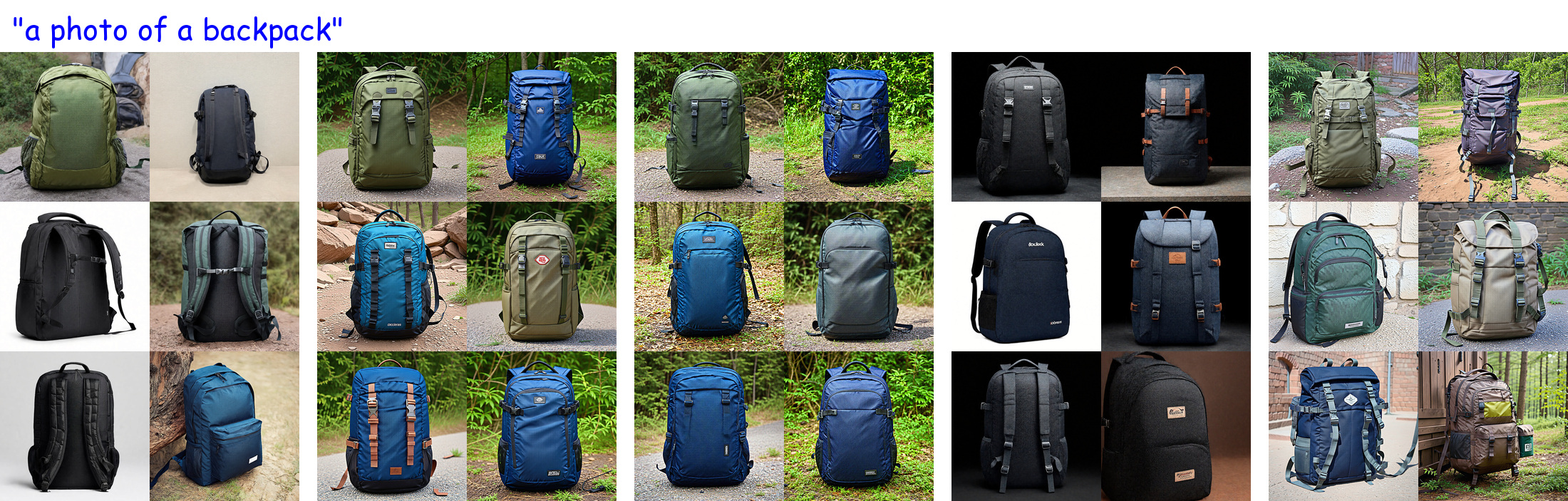}\\[0.12em]
\includegraphics[width=0.97\textwidth]{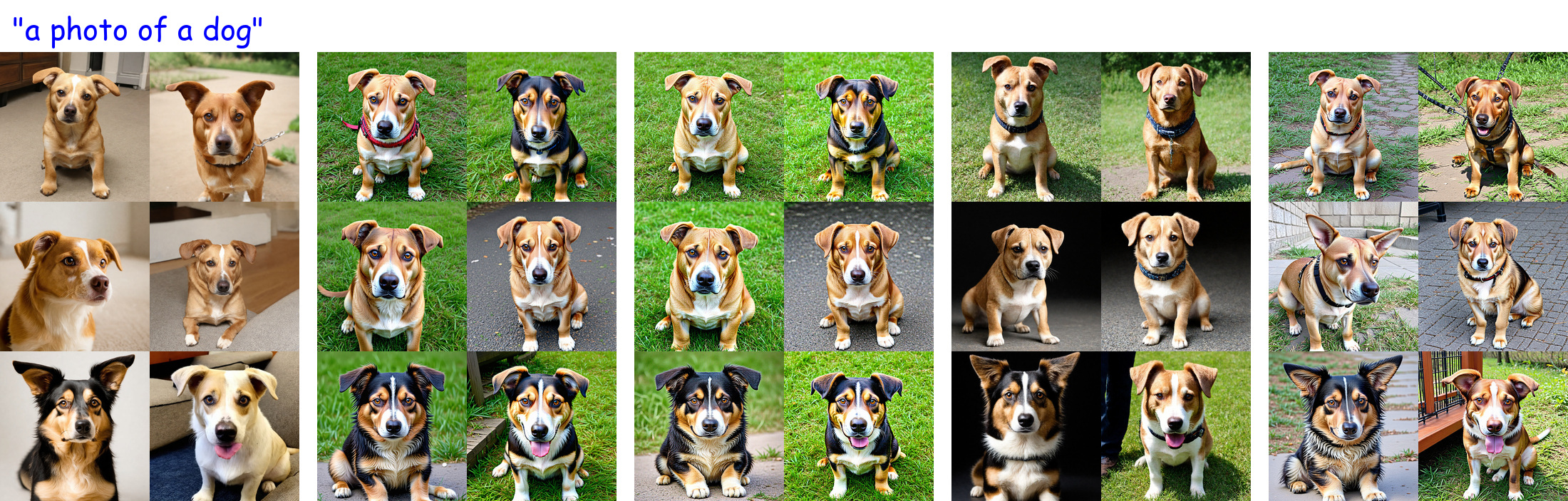}\\[0.12em]
\includegraphics[width=0.97\textwidth]{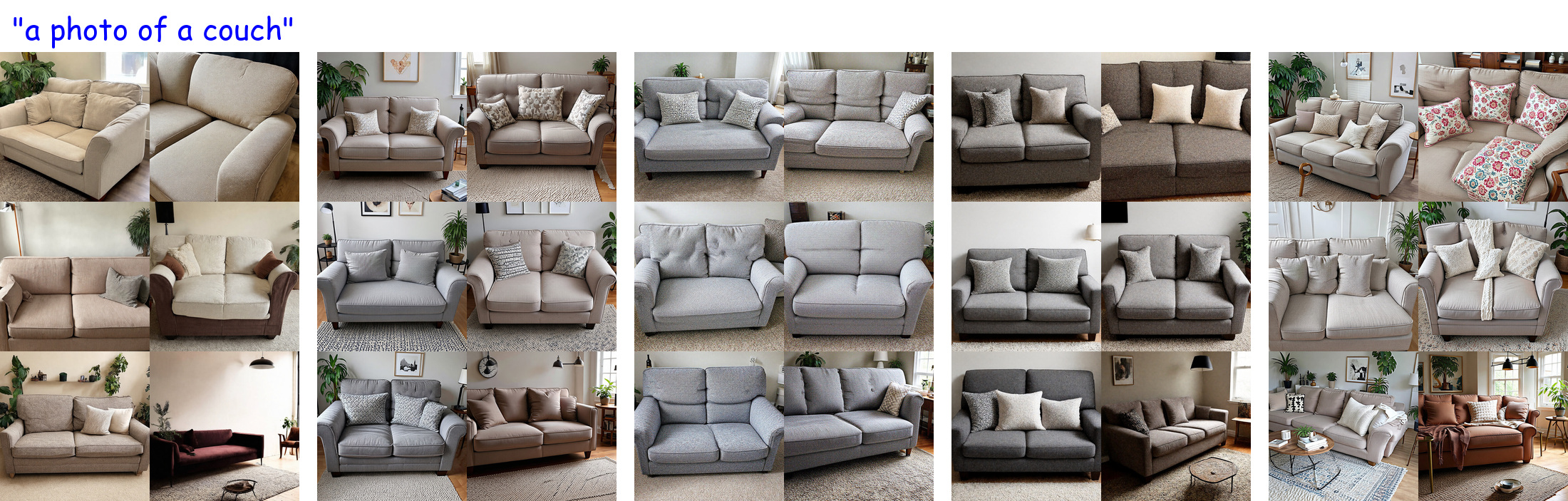}\\[0.12em]
\includegraphics[width=0.97\textwidth]{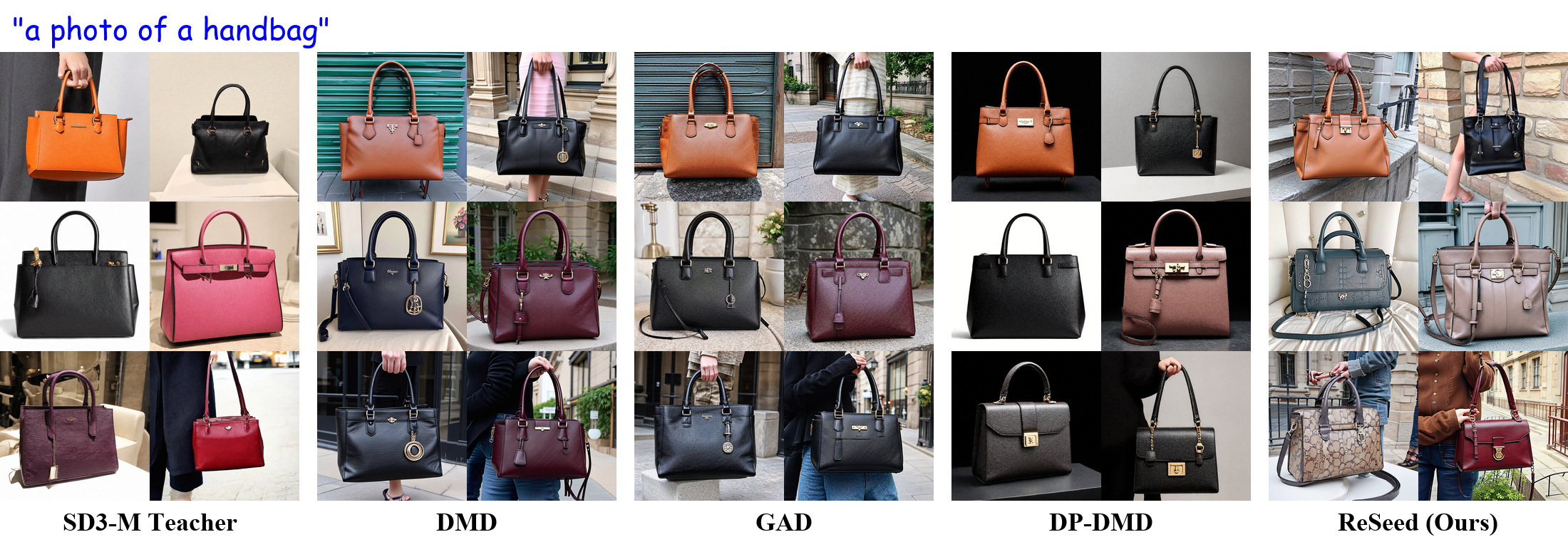}
\caption{Matched-noise SD3-Medium comparisons. Each method
uses the same six noises. }
\label{fig:sd3m-scenes-a}
\end{figure}
\clearpage

\newcommand{\fluxgridmethodlabels}{%
\par\vspace{2pt}
\noindent\begin{minipage}{0.97\textwidth}
\rmfamily\bfseries\fontsize{6.5}{7.5}\selectfont
\parbox[t]{0.32941176\linewidth}{\centering (a) FLUX.1-dev teacher}\hfill%
\parbox[t]{0.32941176\linewidth}{\centering (b) SenseFlow-FLUX}\hfill%
\parbox[t]{0.32941176\linewidth}{\centering (c) ReSeed (Ours)}
\end{minipage}\par
}

\begin{figure}[p]
\centering
\setlength{\abovecaptionskip}{4pt}
\includegraphics[width=0.97\textwidth]{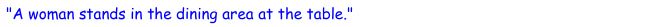}\par\nointerlineskip
\includegraphics[width=0.97\textwidth]{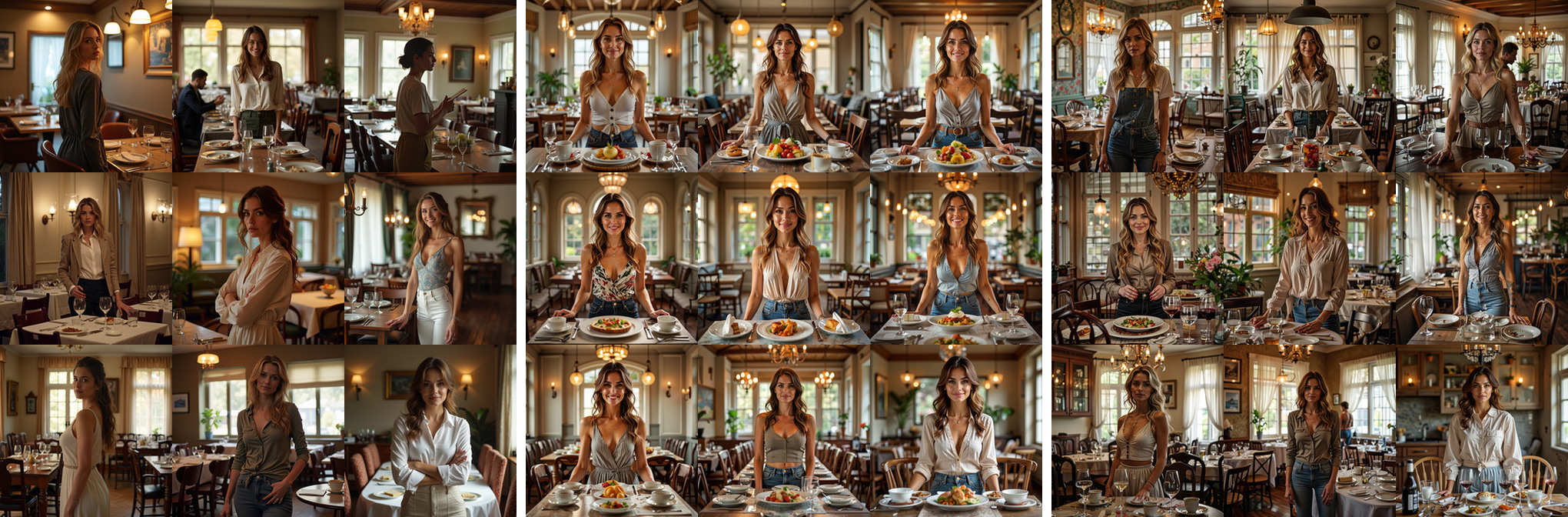}\\[0.12em]
\includegraphics[width=0.97\textwidth]{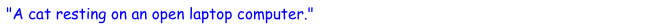}\par\nointerlineskip
\includegraphics[width=0.97\textwidth]{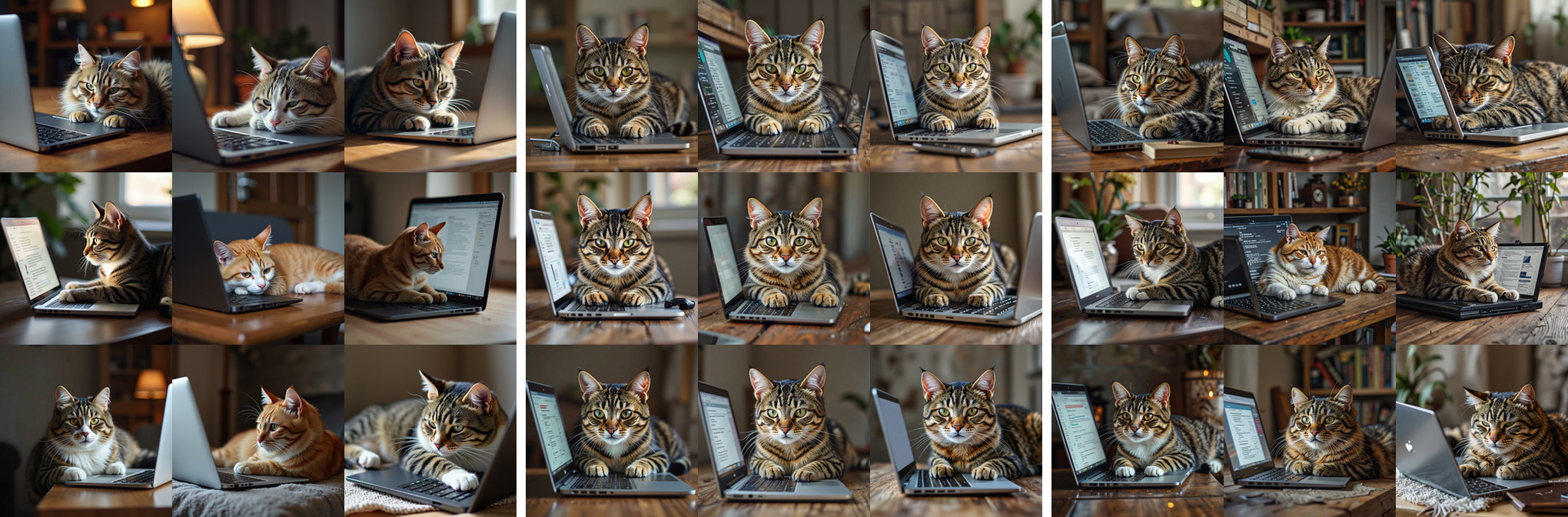}\\[0.12em]
\includegraphics[width=0.97\textwidth]{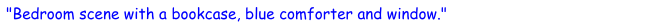}\par\nointerlineskip
\includegraphics[width=0.97\textwidth]{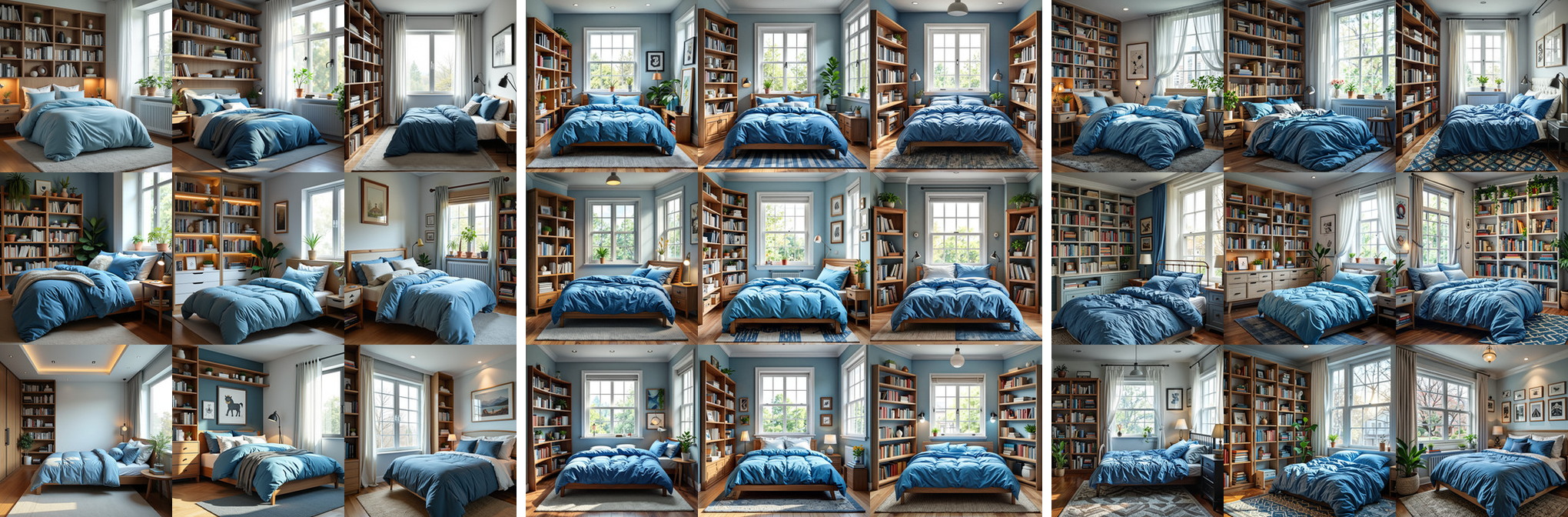}\\[0.12em]
\fluxgridmethodlabels
\caption{Matched-seed FLUX.1-dev comparisons. Each method uses the same nine
noises for each prompt.}
\label{fig:flux-multiprompt-a}
\end{figure}

\begin{figure}[p]
\ContinuedFloat
\centering
\setlength{\abovecaptionskip}{4pt}
\includegraphics[width=0.97\textwidth]{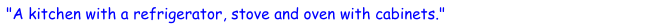}\par\nointerlineskip
\includegraphics[width=0.97\textwidth]{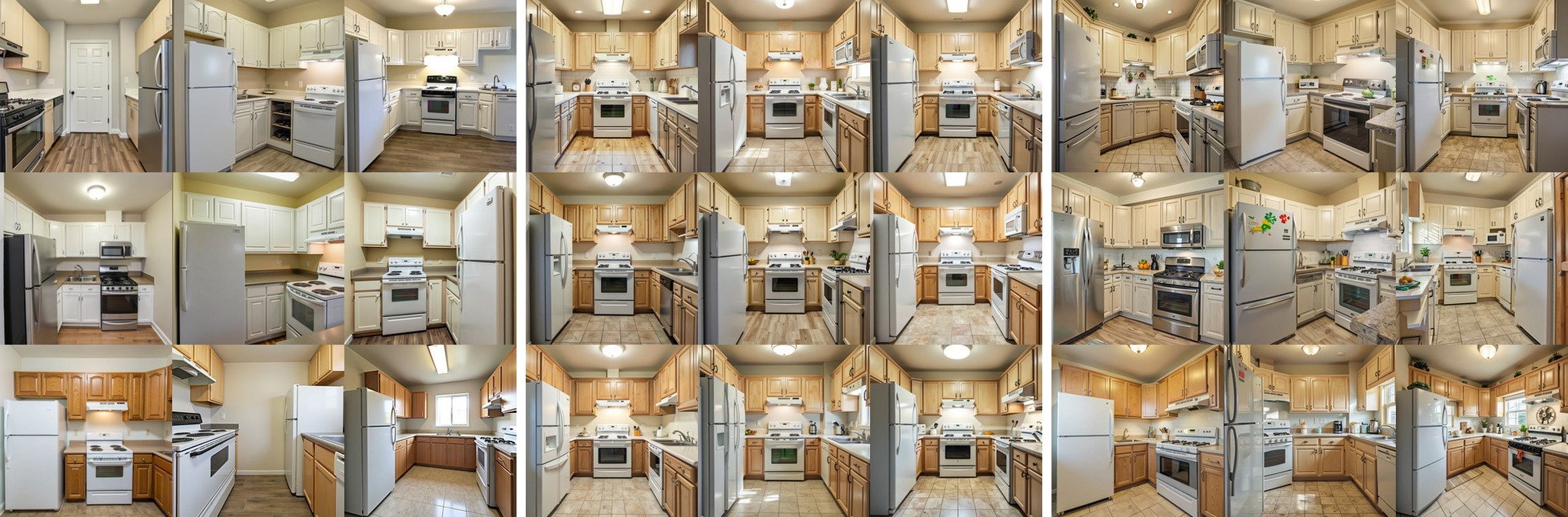}\\[0.12em]
\includegraphics[width=0.97\textwidth]{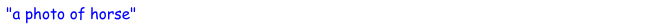}\par\nointerlineskip
\includegraphics[width=0.97\textwidth]{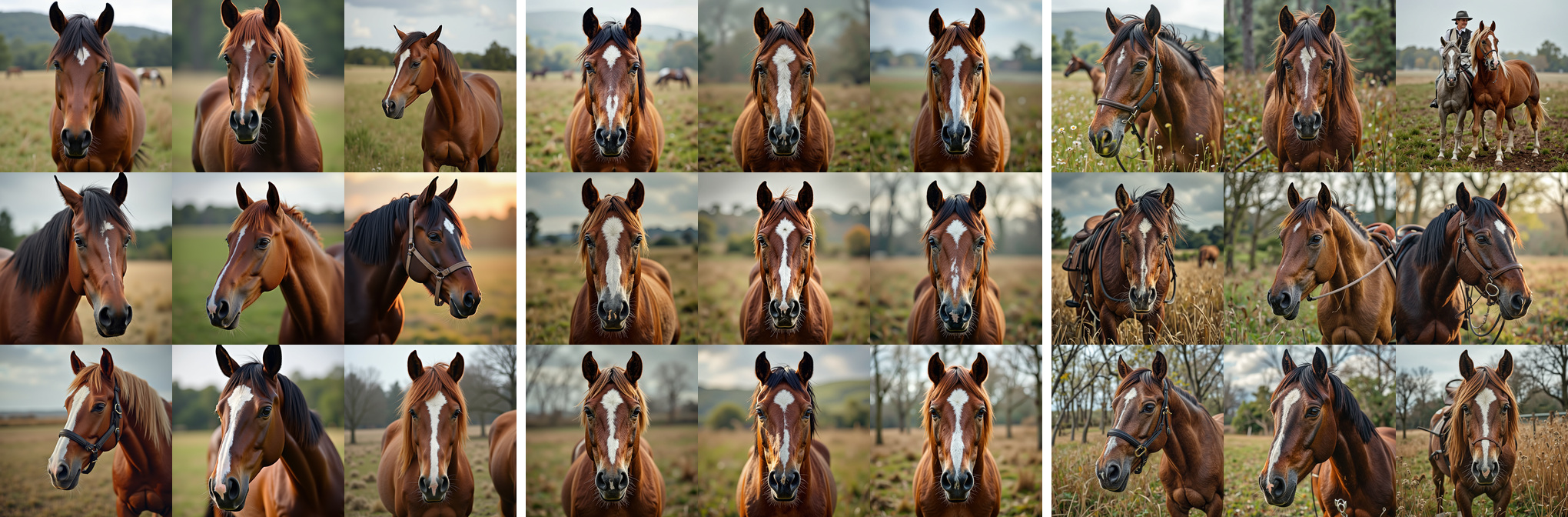}\\[0.12em]
\includegraphics[width=0.97\textwidth]{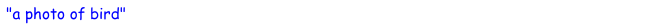}\par\nointerlineskip
\includegraphics[width=0.97\textwidth]{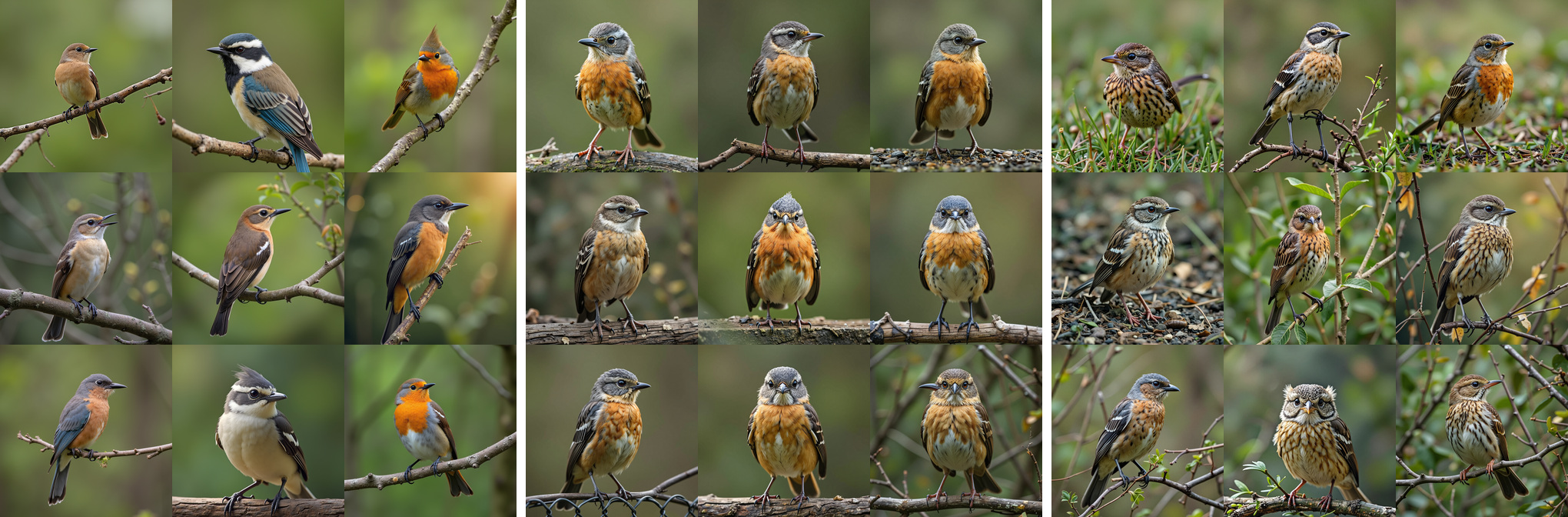}
\fluxgridmethodlabels
\caption{Matched-seed FLUX.1-dev comparisons (continued).}
\end{figure}

\clearpage
\begin{figure}[p]
\centering
\captionsetup[subfigure]{skip=1pt}
\begin{subfigure}[t]{0.495\textwidth}
\centering
{
\ifdefined\pdfimageresolution\pdfimageresolution=72\relax\fi
\resizebox*{!}{0.362\textheight}{%
\begin{minipage}{1948bp}
\offinterlineskip
\includegraphics[width=\linewidth,trim=0 1532bp 0 0,clip]{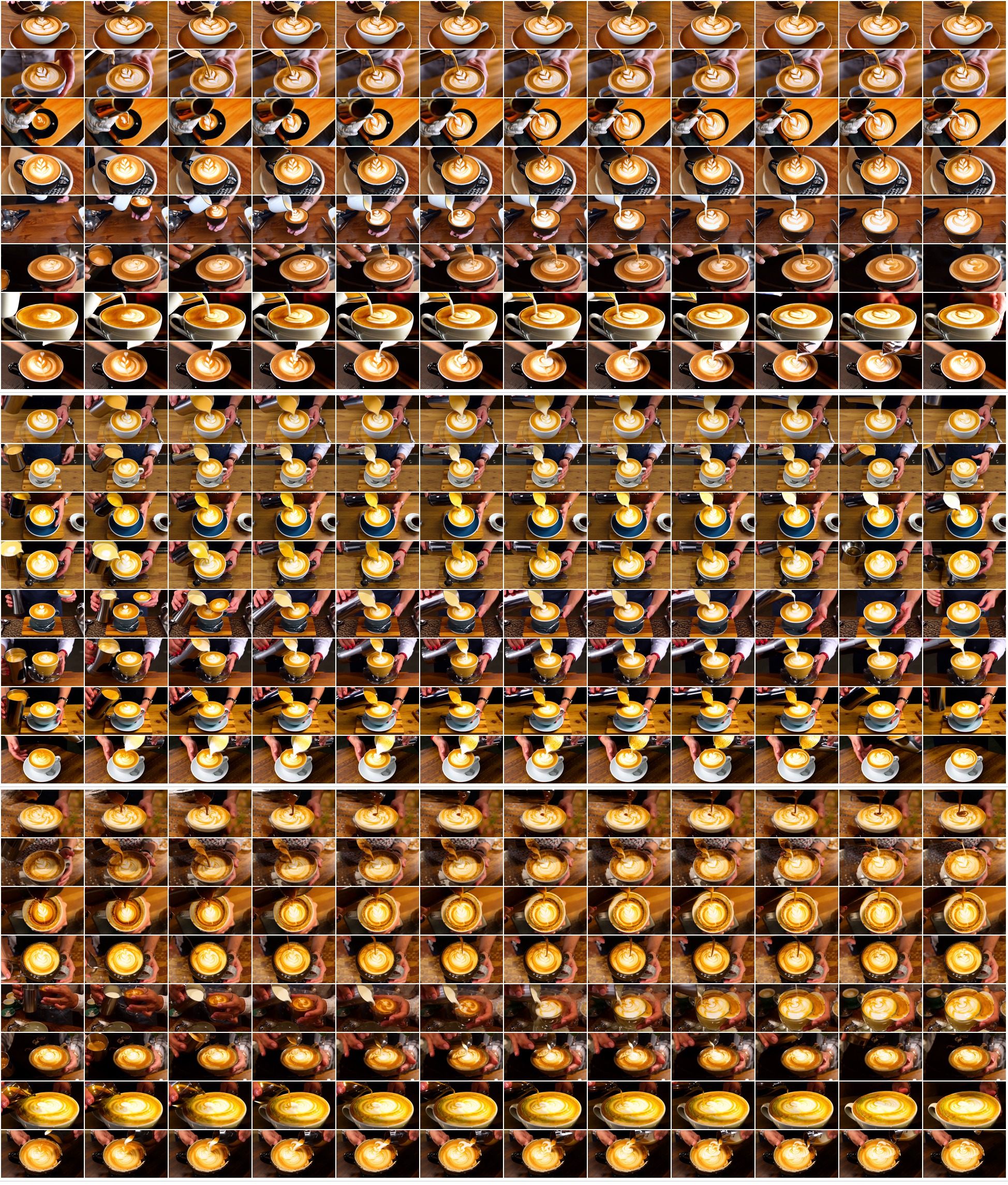}\par
\vspace{35bp}
\includegraphics[width=\linewidth,trim=0 770bp 0 758bp,clip]{fig_wan_filmstrips__wan_latte_art.jpg}\par
\vspace{35bp}
\includegraphics[width=\linewidth,trim=0 0 0 1520bp,clip]{fig_wan_filmstrips__wan_latte_art.jpg}\par
\end{minipage}}
}
\caption{}
\label{fig:wan-filmstrips-latte}
\end{subfigure}\hfill
\begin{subfigure}[t]{0.495\textwidth}
\centering
{
\ifdefined\pdfimageresolution\pdfimageresolution=72\relax\fi
\resizebox*{!}{0.362\textheight}{%
\begin{minipage}{1948bp}
\offinterlineskip
\includegraphics[width=\linewidth,trim=0 1532bp 0 0,clip]{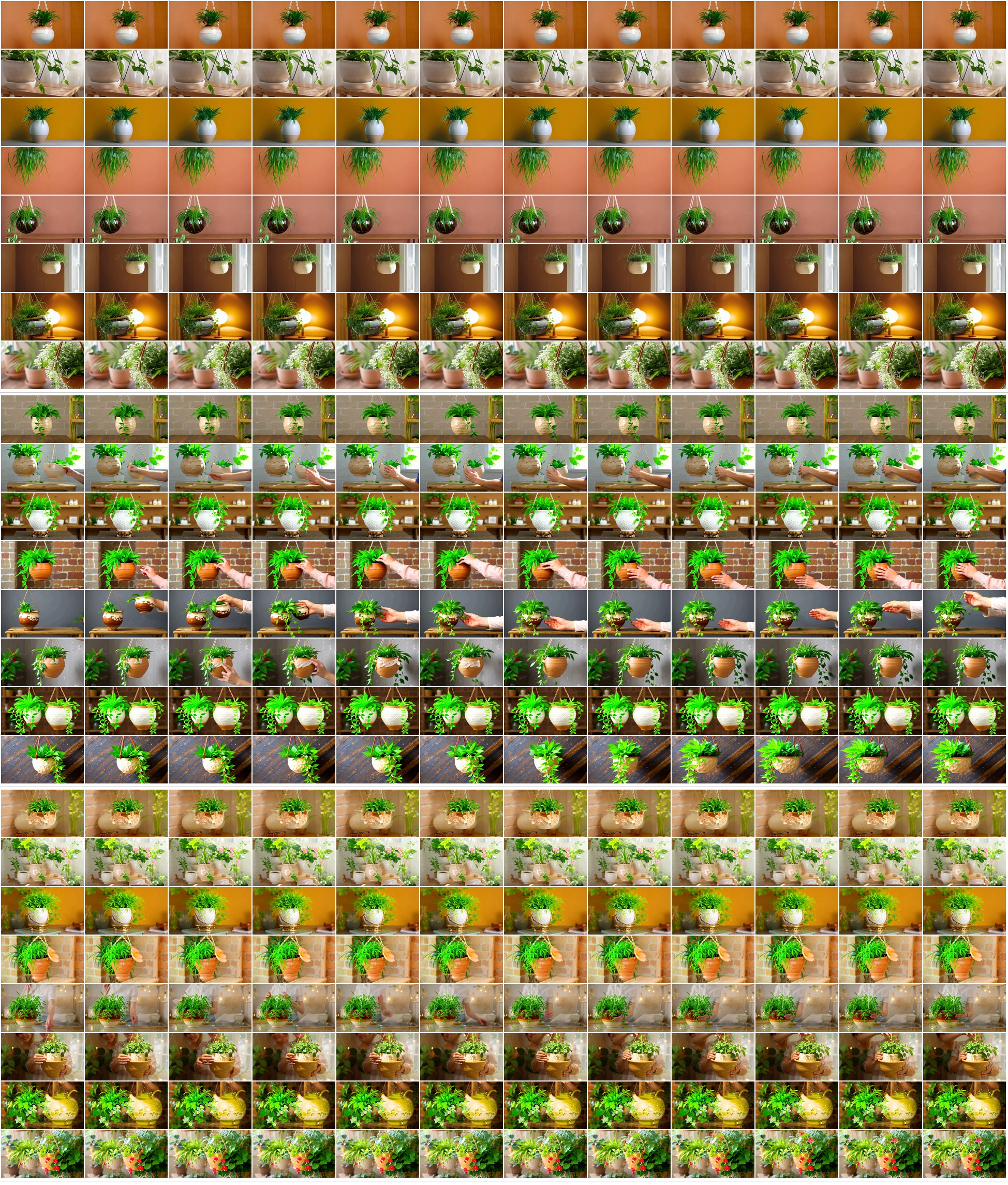}\par
\vspace{35bp}
\includegraphics[width=\linewidth,trim=0 770bp 0 758bp,clip]{fig_wan_filmstrips__wan_ceramic_plant_pot.jpg}\par
\vspace{35bp}
\includegraphics[width=\linewidth,trim=0 0 0 1520bp,clip]{fig_wan_filmstrips__wan_ceramic_plant_pot.jpg}\par
\end{minipage}}
}
\caption{}
\label{fig:wan-filmstrips-plant}
\end{subfigure}\\[0.6ex]
\begin{subfigure}[t]{0.495\textwidth}
\centering
{
\ifdefined\pdfimageresolution\pdfimageresolution=72\relax\fi
\resizebox*{!}{0.362\textheight}{%
\begin{minipage}{1948bp}
\offinterlineskip
\includegraphics[width=\linewidth,trim=0 1532bp 0 0,clip]{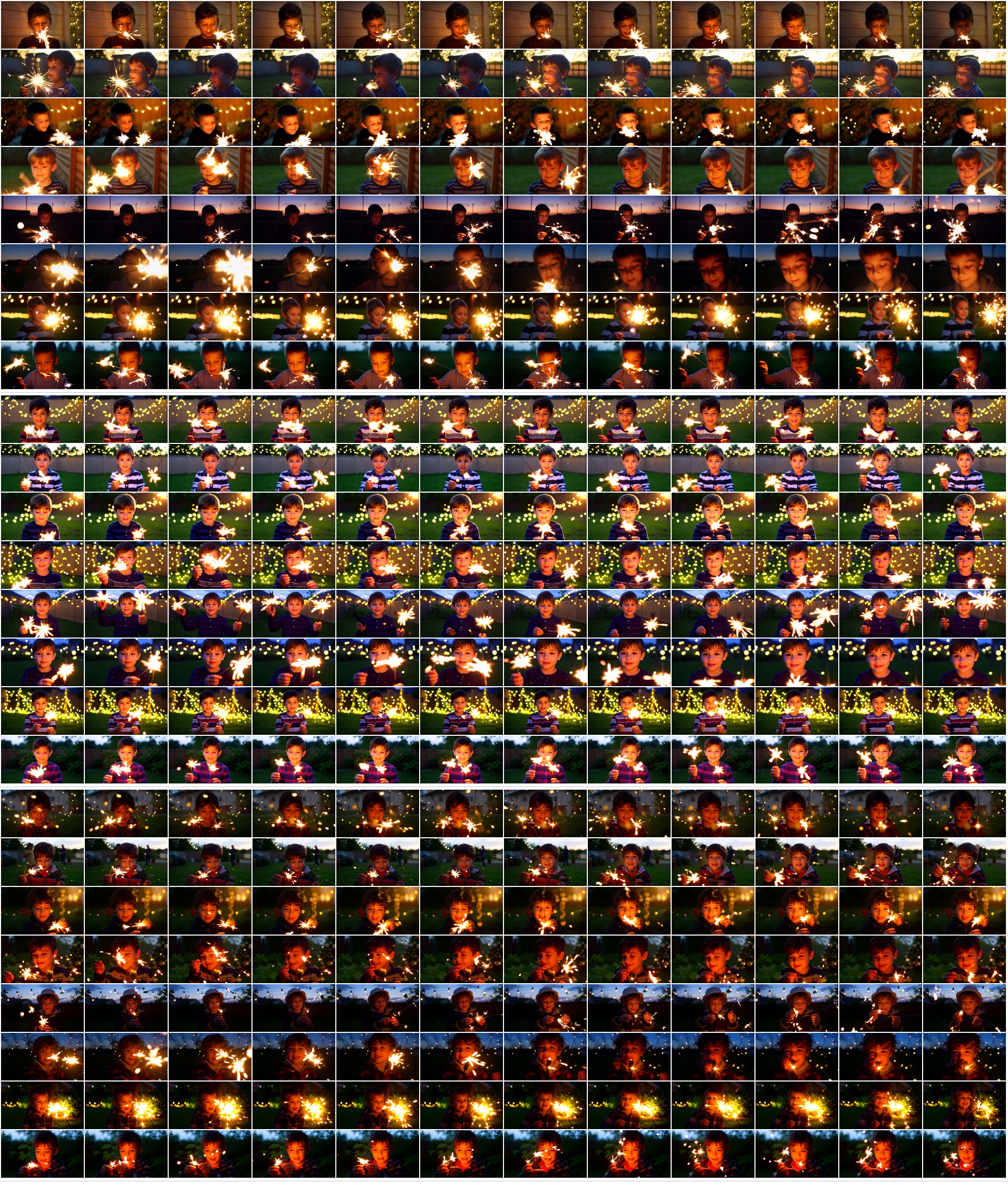}\par
\vspace{35bp}
\includegraphics[width=\linewidth,trim=0 770bp 0 758bp,clip]{fig_wan_filmstrips__wan_sparklers.jpg}\par
\vspace{35bp}
\includegraphics[width=\linewidth,trim=0 0 0 1520bp,clip]{fig_wan_filmstrips__wan_sparklers.jpg}\par
\end{minipage}}
}
\caption{}
\label{fig:wan-filmstrips-sparklers}
\end{subfigure}\hfill
\begin{subfigure}[t]{0.495\textwidth}
\centering
{
\ifdefined\pdfimageresolution\pdfimageresolution=72\relax\fi
\resizebox*{!}{0.362\textheight}{%
\begin{minipage}{1948bp}
\offinterlineskip
\includegraphics[width=\linewidth,trim=0 1532bp 0 0,clip]{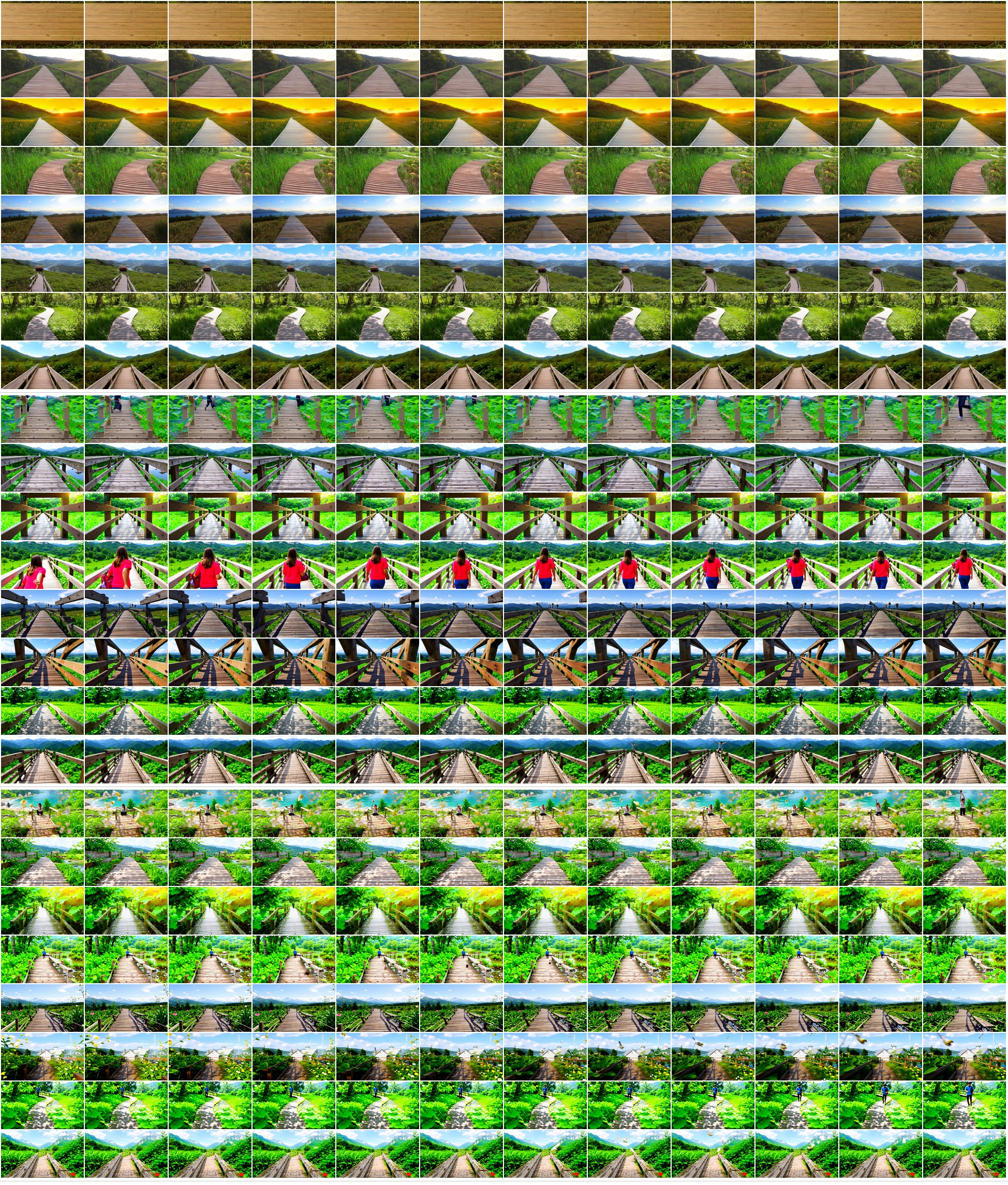}\par
\vspace{35bp}
\includegraphics[width=\linewidth,trim=0 770bp 0 758bp,clip]{fig_wan_filmstrips__wan_wooden_pathway.jpg}\par
\vspace{35bp}
\includegraphics[width=\linewidth,trim=0 0 0 1520bp,clip]{fig_wan_filmstrips__wan_wooden_pathway.jpg}\par
\end{minipage}}
}
\caption{}
\label{fig:wan-filmstrips-pathway}
\end{subfigure}
\caption{Matched-noise frame strips for Wan2.1-T2V-1.3B. In each panel, the
top, middle, and bottom blocks show the 50-step teacher, the 4-step DMD2
student, and the same student with ReSeed (300 updates); within a block, the
eight rows are seeds 0--7 with the same initial noise across blocks, and the
twelve columns are twelve evenly spaced frames from 0 to 80. Prompts and video-level DINO
diversity (teacher / DMD2 / ReSeed, eight seeds):
(a)~\emph{``A barista pours latte art into a cappuccino, viewed from directly
above as a delicate rosetta forms''}, $0.122/0.058/0.245$;
(b)~\emph{``an elegant ceramic plant pot and hanging plant on indoor''},
$0.607/0.247/0.588$;
(c)~\emph{``A young boy lights sparklers in his backyard as twilight settles,
his face illuminated in flickering gold''}, $0.113/0.082/0.132$;
(d)~\emph{``scenic view from wooden pathway''}, $0.599/0.339/0.475$.
}
\label{fig:wan-filmstrips}
\end{figure}

\clearpage

\begin{figure}[p]
\centering
\includegraphics[height=0.92\textheight]{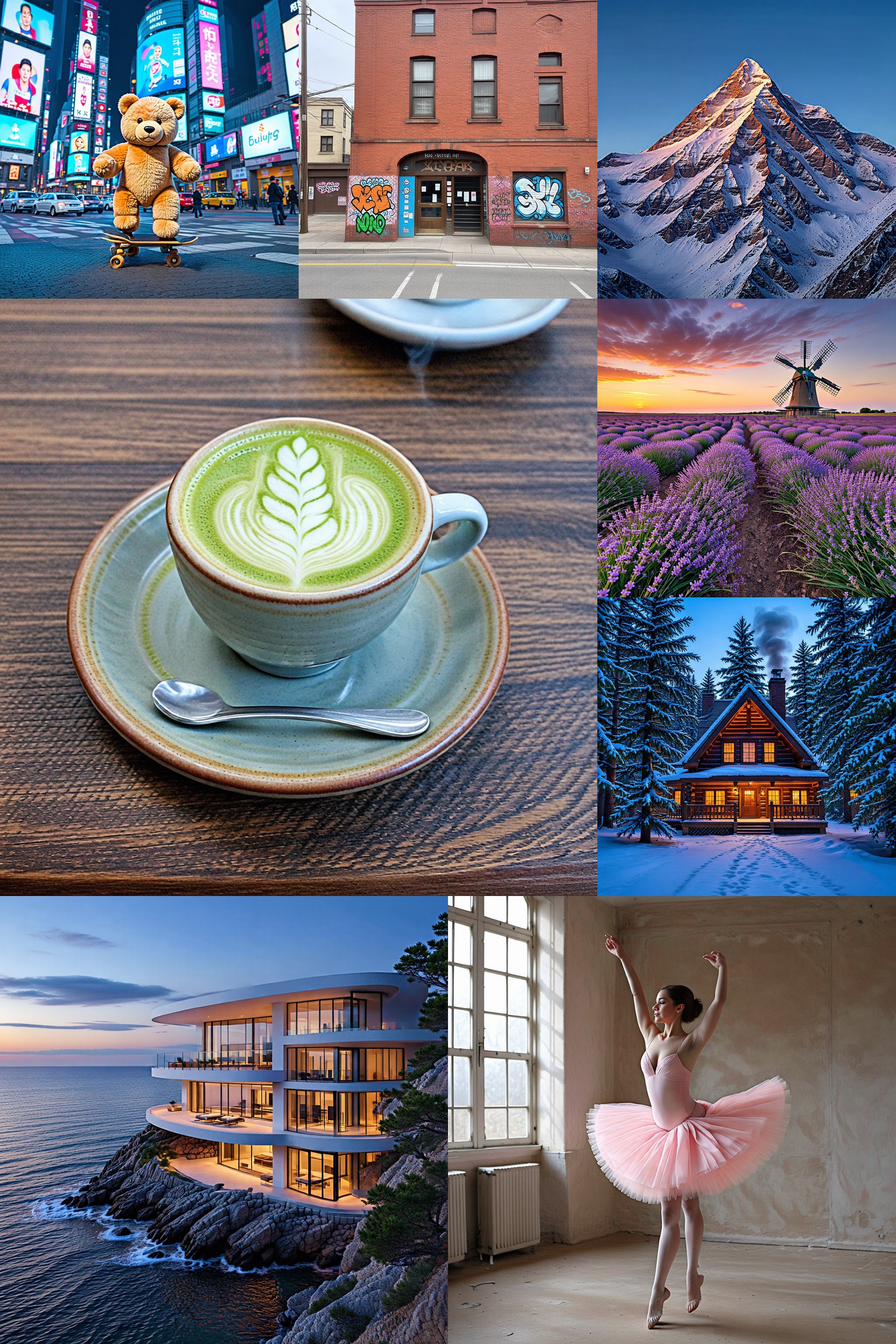}
\caption{SD3-Medium: a gallery of 4-step ReSeed samples at $1024\times1024$.}
\label{fig:showcase-sd3m}
\end{figure}

\begin{figure}[p]
\centering
\includegraphics[height=0.92\textheight]{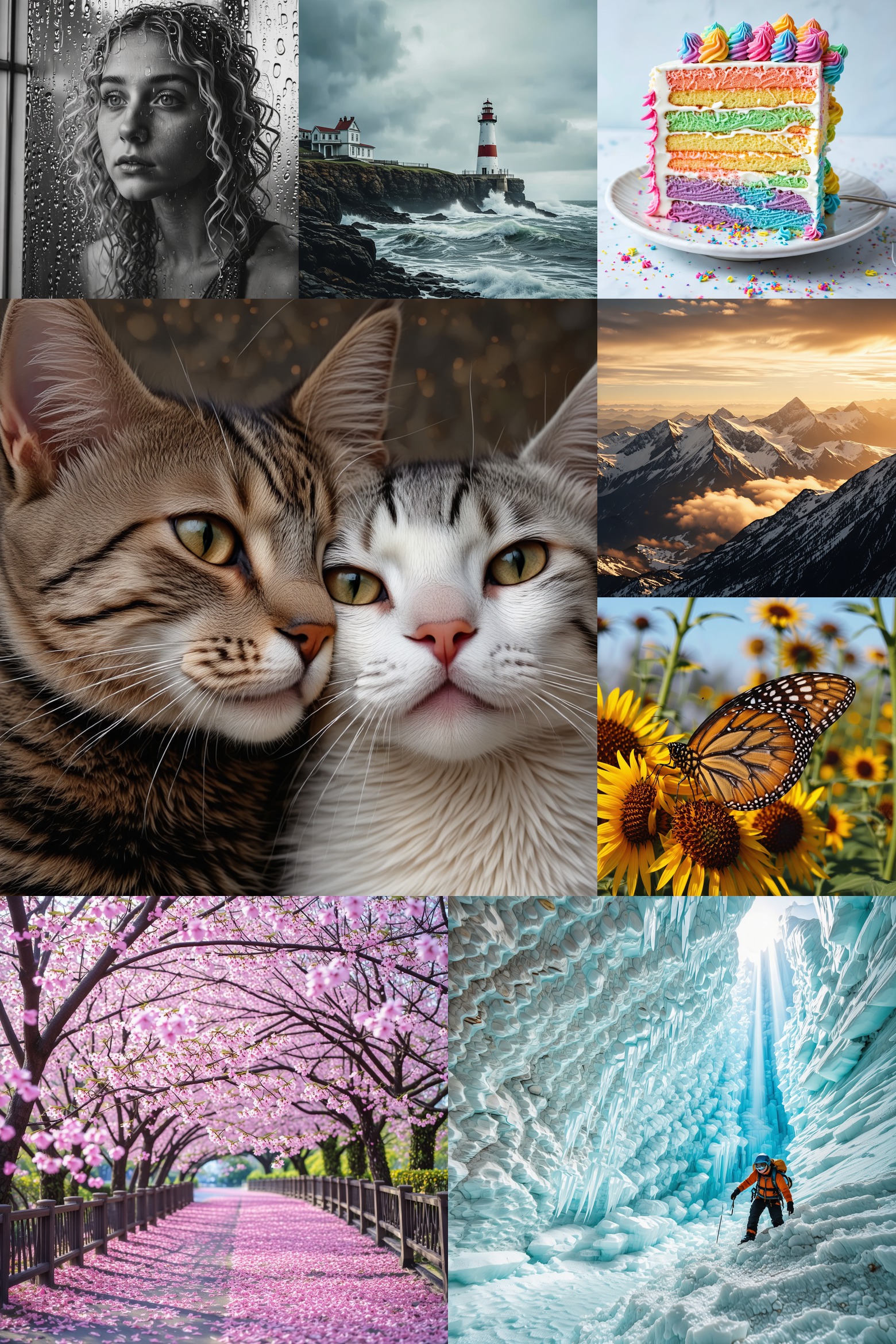}
\caption{SD3.5-Medium: a gallery of 4-step ReSeed samples at $1024\times1024$.}
\label{fig:showcase-sd35m}
\end{figure}

\begin{figure}[p]
\centering
\includegraphics[height=0.92\textheight]{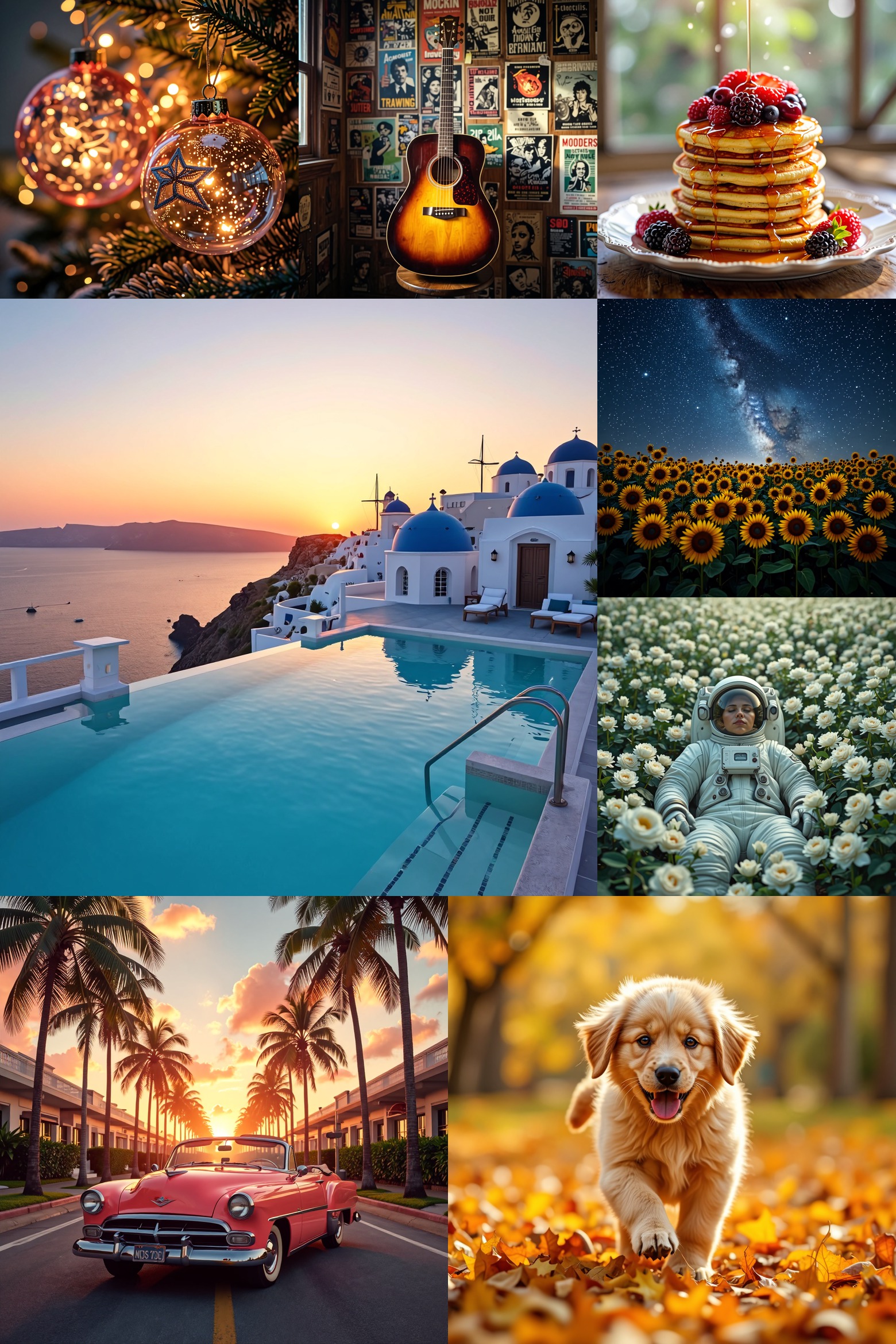}
\caption{FLUX.1-dev: a gallery of 4-step ReSeed samples at $1024\times1024$.}
\label{fig:showcase-senseflow}
\end{figure}

\clearpage

\end{document}